\documentclass{egpubl}
\usepackage{eurovis2026}

\SpecialIssueSubmission    %
\CGFccby
\usepackage{paralist}

\usepackage[T1]{fontenc}
\usepackage{dfadobe}  

\usepackage{cite}  %
\BibtexOrBiblatex
\electronicVersion
\PrintedOrElectronic
\ifpdf \usepackage[pdftex]{graphicx} \pdfcompresslevel=9
\else \usepackage[dvips]{graphicx} \fi

\usepackage{egweblnk}
\usepackage{lipsum}
\usepackage{amsmath, amssymb, amsfonts}
\usepackage{graphicx}
\usepackage{hyperref}
\usepackage{algorithm}
\usepackage{algorithmic}
\usepackage{bm}

\usepackage{tabularx}
\usepackage{booktabs}
\usepackage{graphicx}
\usepackage{xcolor,colortbl}

\usepackage{subcaption} %
\usepackage[colorinlistoftodos]{todonotes} %

\usepackage[table]{xcolor}

\definecolor{myblue}{RGB}{35,76,209}

\newcommand{\ourmetricFull}{Class Angular Distortion Index}
\newcommand{\ourmetricAbbrev}{CADI}

\title[CADI]
      {Class Angular Distortion Index for Dimensionality Reduction}

\author[Gunaratne et al.]
{\parbox{\textwidth}{\centering 
Kaviru Gunaratne\orcid{0009-0004-8091-9080}, Stephen Kobourov\orcid{0000-0002-0477-2724}, and Jacob Miller\orcid{0000-0002-0567-785X}
        }
        \\
{\parbox{\textwidth}{\centering 
Technical University of Munich\
       }
}
}
\begin{document}

\maketitle
\begin{abstract}
  Dimensionality reduction (DR) techniques are often characterized by whether they preserve global, high-level structures in the data or local, neighborhood structures. 
  This distinction matters in visualization: global methods can obscure clusters while local methods can over-emphasize them. Yet, even when clusters appear distinct, their relative arrangement in the projection may be arbitrary or misleading, a common issue in techniques such as t-SNE and UMAP. Existing cluster quality metrics either only measure cluster separability or assume spherical, globular clusters in the original space. We introduce the \ourmetricFull\ (\ourmetricAbbrev), a metric that uses internal angles among point triples to determine the faithfulness of cluster organization in a projection. We show cases on both real and synthetic data where existing cluster metrics fail, but \ourmetricAbbrev\ provides an interpretable result. Since it relies on computing angles, \ourmetricAbbrev\ is also differentiable, enabling optimization. We demonstrate this with a \ourmetricAbbrev-based DR technique.


\end{abstract}

\maketitle

\begin{figure*}[t]
     \centering
     \includegraphics[width=\linewidth]{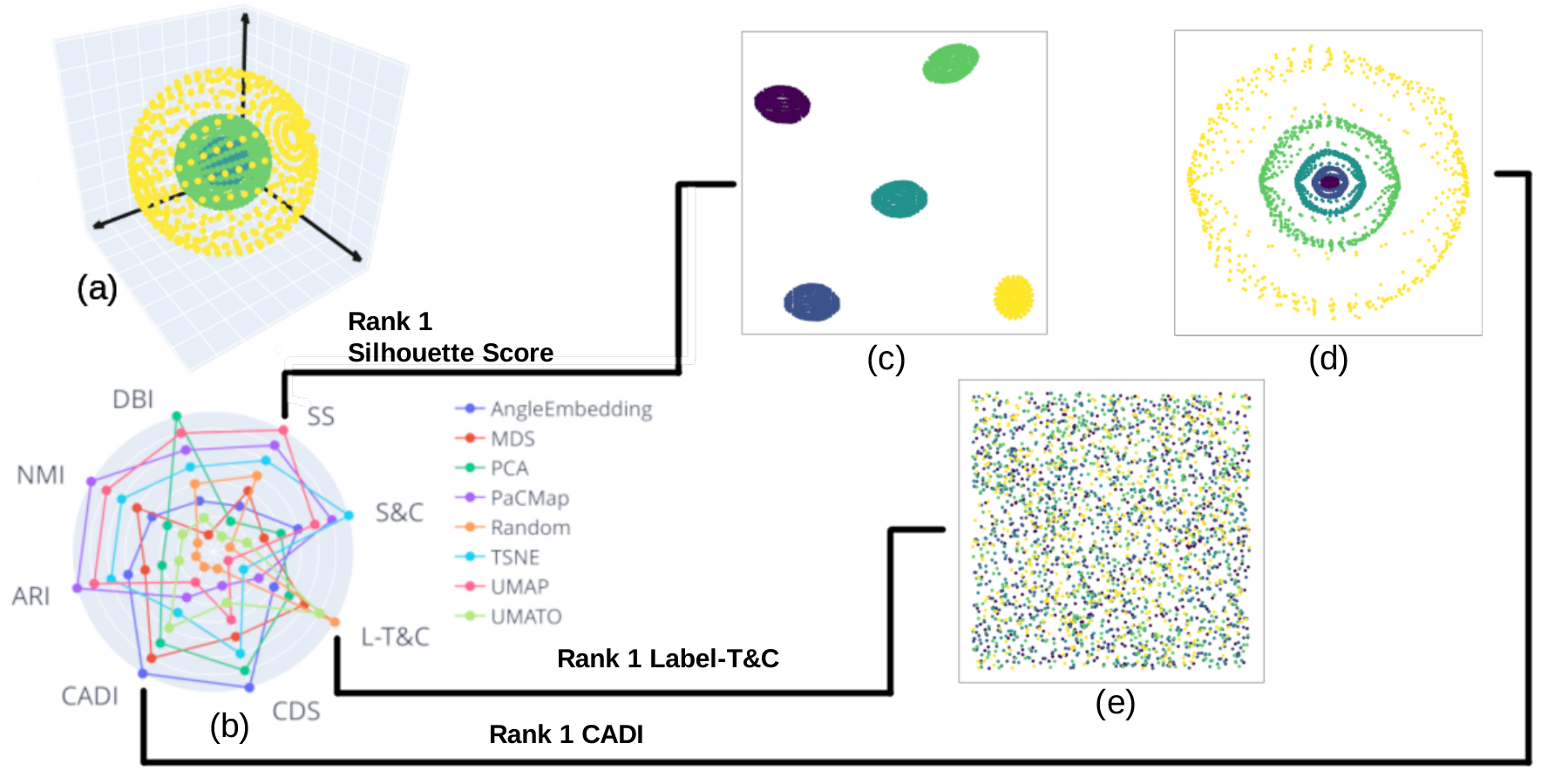}
    \caption{
    An example of a synthetic dataset where the globularity assumption for clusters fails to hold. (a) shows the input 3D dataset, which consists of points sampled from the surfaces of five nested spheres. (b) shows how various DR techniques rank under different metrics. (c) is the UMAP projection of the data, which is ranked best under the Silhouette score. (d) shows that under \ourmetricAbbrev, AngleEmbedding is ranked best as it best captures the inter-cluster relationships.
    Finally, (e) shows that Label-T\&C assumes that, as the cluster centroids overlap, a random projection is the most suitable.  
    }
    \label{fig:concentricTeaser}
    \vspace{-0.4cm}
\end{figure*}

\section{Introduction}
\label{sec:intro}
High-dimensional data arise whenever the number of features exceeds the limits of direct human visualization. Working in these spaces is useful, as linear algebra and geometry provide rich tools for analysis, but our intuition quickly breaks down when we cannot ``see'' the structure of the data. 

This inability to visualize the data directly creates a fundamental challenge. Analysts often do not know which patterns exist in the dataset or which analytical methods are appropriate until they can first obtain a meaningful visual representation. Dimensionality reduction (DR) addresses this by projecting high-dimensional data into two or three dimensions to enable visual exploration.

One of the most frequent tasks for DR plots is cluster identification, for which techniques such as t-SNE~\cite{maaten2008tsne} and UMAP~\cite{mcinnes2018umap} reliably reveal distinct groupings when they are present. However, more complex analyses frequently performed on these plots are often less reliable. DR techniques often position clusters in ways that do not reflect their true relationships in the original space. Cluster density, hierarchy, and shape are also regularly distorted. Despite this, analysts routinely draw inferences about relative positions, nested structure, or overall cluster geometry from DR visualizations, leading to incorrect conclusions about the data~\cite{jeon2025stop,jeon2025unveiling}. 

Quality metrics attempt to quantify such distortions by comparing structures in the high- and low-dimensional spaces. Yet despite the prevalence of cluster-level reasoning tasks, only a few metrics directly evaluate cluster-level preservation. Classical measures such as the Silhouette score~\cite{SilhouetteScore1987} and related indices~\cite{DaviesBouldinIndex1979} make the assumption that clusters are both well-separated in the dataset and globular in shape. The recently proposed metric, Label Trustworthiness and Continuity~\cite{jeon2023classesclusters}, addresses this first issue, accounting for how separated clusters are in the data. However, it still relies on centroid-based comparisons and therefore retains the globularity assumption.

In this work, we introduce the \ourmetricFull\ (\ourmetricAbbrev) which makes use of internal angles to measure how well shapes and geometric relationships of clusters are preserved in a projection. By focusing on between-class angular relationships, \ourmetricAbbrev\ moves beyond both separation and globularity assumptions, capturing complex nested, curved, or intertwined clusters.

This advantage is illustrated in \autoref{fig:concentricTeaser}. Here, the data are synthetically generated to come from the surface of five 3D spheres of different radii, each nested inside another. Despite this regular structure, the Silhouette score prefers projections that depict each shell as a ``blob'' completely losing the cluster relationships. Meanwhile, since Label T\&C makes use of the cluster centroids, it assumes all five clusters overlap and prefers projections that depict this. \ourmetricAbbrev\ avoids these pitfalls and accurately identifies the projection that best preserves the cluster-level structure. 

Our contributions are:
\begin{itemize}
    \item We introduce a new DR quality metric, \ourmetricAbbrev, to measure cluster-level structure preservation through internal angles.
    \item We evaluate \ourmetricAbbrev\ on synthetic and real-world datasets, demonstrating the limitations of existing cluster-level metrics and how our approach overcomes them.
    \item We provide an initial exploration of how \ourmetricAbbrev\ might be used as an optimization objective to produce angle-preserving projections that are qualitatively distinct from existing methods.
    \item We provide an open-source implementation of \ourmetricAbbrev\, along with all experimental scripts used to generate evaluation data, at 
    {\url{https://github.com/KaviruGunaratne/cadi}}. 
\end{itemize}

\section{Background and Related Work}
Dimensionality reduction (DR) is a broad class of unsupervised techniques that embed high-dimensional data into a low-dimensional space, typically for visualization. More formally, given a dataset $X \in \mathbb{R}^{n \times d}$, the dimensionality reduction problem is to find a map
$f: X \rightarrow Y$
where $Y \in \mathbb{R}^{n \times t}$ ($t = 2,3$). 
This map should preserve the structure of $X$ as faithfully as possible in the projection, $Y$. Comprehensive surveys on DR for visualization are found in Nonato and Aupetit~\cite{nonato2018multidimensional}, and Espadoto et al.~\cite{espadoto2019toward}. 

\subsection{Background}
Classical linear techniques such as principal component analysis (PCA)~\cite{pearson1901principal} or Laplacian Eigenmaps~\cite{belkin2003laplacian} provide interpretable axes but are limited in the structures they can express. Non-linear methods offer greater expressive power, though at the cost of interpretability and reliability: the projection axes have no inherent meaning. Despite this, the non-linear techniques of multi-dimensional scaling (MDS)~\cite{borg2005modern}, t-distributed Stochastic Neighbor Embedding (t-SNE)~\cite{maaten2008tsne}, and Uniform Manifold Approximation and Mapping (UMAP)~\cite{mcinnes2018umap} are among the most popular visualization techniques for high-dimensional data today. 

Nonlinear DR techniques can be viewed through the lens of manifold learning, which assumes that the data lie on a lower-dimensional manifold embedded in the original space. These techniques typically construct a (dis)similarity graph, implicitly in MDS and t-SNE, or explicitly in UMAP, and then optimize a low-dimensional layout that best preserves these relationships.

When selecting among non-linear DR methods, the key choice is whether to emphasize global or local structure. Global structure reflects broad pairwise relationships that determine the overall layout of the data, while local structure focuses on preserving neighborhoods. These goals are largely orthogonal: in high dimensions, neither can be perfectly maintained in a low-dimensional projection, and optimizing one often comes at the expense of the other. MDS is the canonical example of a global method, while t-SNE and UMAP are popular local methods. 

However, DR methods introduce well-documented distortions. Cluster visibility and separation depend strongly on hyperparameters, and the relative positions of clusters in a projection are largely arbitrary for t-SNE and UMAP~\cite{wattenberg2016use}. As a result, clusters that are close in the projection may be far apart in the original space, and vice versa. More complex relationships, such as elongated, nested, or intertwined clusters, are often lost. Quantifying these distortions requires quality metrics that capture something other than just global or local accuracy, but cluster-level structure.

\subsection{Quality Metrics for Dimensionality Reduction}

\begin{table*}[ht]
\centering
\caption{Cluster-related metrics used in dimensionality reduction evaluation, with their original references and representative DR papers that employ them.}
\label{tab:cluster-metrics}
\begin{tabular}{ l l l }
\hline
\textbf{Metric} & \textbf{Original Reference} & \textbf{ Usage in DR} \\
\hline

Silhouette Score 
& Rousseeuw (1987)~\cite{SilhouetteScore1987} 
& ~\cite{Kwon2025clusterEval}
~\cite{Cardarelli2022clusterEval}
~\cite{Yang2021UMAPtranscriptomic}
~\cite{Zahed2025Silhouette}
~\cite{Becker2019ReNDA}
\\[6pt]

Davies--Bouldin Index (DBI)
& Davies and Bouldin (1979)~\cite{DaviesBouldinIndex1979}
& ~\cite{Kwon2025clusterEval}
~\cite{Cardarelli2022clusterEval}
~\cite{Becker2019ReNDA}
\\[6pt]

Normalized Mutual Information (NMI)
& Danon et al. (2005)~\cite{normalizedMutualInformation2005}
& ~\cite{Kwon2025clusterEval}
~\cite{Sun2019scRNAclusterEval}
~\cite{Yang2021UMAPtranscriptomic}
 \\[6pt]

Adjusted Rand Index (ARI)
& Hubert and Arabie (1985)~\cite{adjustedrandindex1985}
& ~\cite{Kwon2025clusterEval}
~\cite{Sun2019scRNAclusterEval}
~\cite{Yang2021UMAPtranscriptomic}
\\[6pt]

Cluster Distance Score 
& Miller et al. (2023)~\cite{Jacob2023LGS}
& \cite{Jacob2023LGS}
\\ [6pt]

Label Trustworthiness and Continuity
& Jeon et al. (2023)~\cite{jeon2023classesclusters} 
& ~\cite{bae2025metric-design-ltnc}
~\cite{jeon2025datasetadaptivedimensionalityreduction}
\\ [6pt]

Steadiness and Cohesiveness
& Jeon et al. (2021)~\cite{jeon2021measuring} 
& ~\cite{bae2025metric-design-ltnc}
~\cite{UMATO}

\\

\hline
\end{tabular}\vspace{-0.5cm}
\end{table*}

In DR, a \textit{quality metric} is some computed value based on both the original and projected data that quantifies how ``well'' the projection represents the high-dimensional data. Such metrics are needed to quantify the error present in a projection, which is unavoidable for all but the most trivial datasets~\cite{nonato2018multidimensional}. Formally, a quality metric is a function $M: (X,Y) \rightarrow \mathbb{R}$, that evaluates how faithfully projection $Y$ represents $X$. Machado et al.~\cite{machado2025necessary} recently highlighted the limitations of current quality metrics, motivating further the design of new metrics that capture different information.

Most evaluation metrics in dimensionality reduction generally measure the preservation of structure along local and global axes. Local structures such as proximity of nearest neighbors are measured by Trustworthiness \& Continuity~\cite{kaski2003trustworthiness, venna2010nerv}, Neighborhood hit~\cite{espadoto2019toward}, or Scale-normalized KL divergence~\cite{sns_snkl}. Meanwhile, global structures such as pairwise distances are measured by the
Shepard goodness score~\cite{Shepard_1962_SGS}, random triplet loss~\cite{pacmap_randomtripletloss}, or Scale-normalized stress~\cite{sns_snkl}. These metrics are widely used but capture only pointwise or pairwise structure.

However, dimensionality reduction techniques are also evaluated at the cluster level. That is, by partitioning the data into clusters using a clustering algorithm or class labels, the preservation of these group-level structures is measured. Various metrics exist for this purpose. Halkidi et al.~\cite{Halkidi2001InternalExternal} consider two categories of metrics: Internal Validation Measures (IVMs) and External Validation Measures (EVMs). EVMS such as Normalized Mutual Information (NMI)~\cite{normalizedMutualInformation2005} and Adjusted Rand Index~\cite{adjustedrandindex1985} compare cluster assignments in the projection (produced by a clustering algorithm) to ground truth labels. Meanwhile, IVMs directly analyze the structure of the projected data without requiring clustering. The Silhouette Score~\cite{SilhouetteScore1987}, Davies-Bouldin Index~\cite{DaviesBouldinIndex1979}, and Distance Consistency~\cite{Sips2009DistanceConsistency} quantify cluster separability, assuming clusters are compact and well separated. 

Beyond classical separability-based indices, several recent measures assess cluster-level structure more directly. The Cluster Distance Score~\cite{Jacob2023LGS} evaluates whether the relative configuration of cluster centroids in the projection reflects their arrangement in the high-dimensional space, thus focusing on the global organization of clusters instead of separability. Originally defined for network layout, its application in DR aligns with the growing recognition~\cite{paulovich2025dimensionality} of shared principles between network layout and dimensionality reduction. 

Label Trustworthiness and Continuity (Label-T\&C)~\cite{jeon2023classesclusters} adapts Trustworthiness and Continuity to labeled data, rewarding separation in the projection only when groups are separated in the original space and thus accounting for potential cluster overlap. Steadiness and Cohesiveness (S\&C)~\cite{jeon2021measuring} also accounts for the structure of the dataset by clustering in both the dataset and the projection, and measuring how they correspond. However, S\&C determines clusters internally as part of the metric computation, limiting the evaluation of a user-specified partition.

Motta et al.~\cite{motta2015graph} also propose another metric that compares partitions derived from graphs constructed in both the high-dimensional and the projected spaces. Across all these metrics, we argue that they suffer from a common limitation: they implicitly favor compact, convex, and well-separated clusters, and only partially capture more complex geometric arrangements such as nested, curved, or intertwined structures.

\subsection{Measuring internal angles between points}

Our proposed cluster-level metric, \ourmetricFull, measures internal angles between triplets of points. Geometric relationships based on internal angles have been explored in other areas of machine learning. In metric learning, Wang et al.~\cite{wang2017angular} introduce an \emph{angular loss} as a loss function. Here, given a triplet of points \textit{i}, \textit{j}, and \textit{k}, where \textit{j} and \textit{k} are neighbors, and \textit{i} is a far-away point (a negative sample), the goal is to minimize the angle $\measuredangle{jik}$.  Wang et al. report that this \textit{angular loss} converges to a more optimal minimum and achieves better results on benchmarks. They provide theoretical justification based on the gradients of the loss functions that angular losses are easier to optimize than the distance-based ``triplet loss'' used in metric learning. However, metric learning and manifold learning represent distinctly different problems. Metric learning is a supervised problem that aims to learn an appropriate distance function based on data, while in our setting of manifold learning, it is unsupervised, and we learn a low-dimensional representation.

Angle-based ideas also appear in the manifold learning literature. Fischer et al.~\cite{fischer2024mercat} introduce a new dimensionality reduction algorithm, MERCAT, that attempts to preserve internal angles between triplets in the dataset in a projection on a 2-sphere. However, while MERCAT optimizes a function that aims to preserve all internal angles, we focus on preserving internal angles between pairs of clusters, focusing on cluster-level structure. Additionally, instead of projecting on a 2-sphere, we show that this more generally works on traditional flat, low-dimensional projections.

\section{\ourmetricFull\ (\ourmetricAbbrev)}
Here we introduce the \ourmetricFull\ (\ourmetricAbbrev). Inspired by how a human observer moves around an object to understand its shape from multiple viewpoints, our metric uses internal angles between triplets to measure class relationships. Specifically, we look at between-class angles so as to capture this class-level structure. 
This accounts for non-globular class shapes in a way existing metrics do not.

First, we establish the notation: we are given a $d$-dimensional dataset $X = \{x_1, \dots, x_n\}$ and a corresponding low-dimensional projection $Y = \{y_1, \dots, y_n\}$. We are also provided with a partition $C = \{C_1,\dots,C_m\}$ of $X$ that assigns each data point to a class/cluster (e.g., from ground-truth labels).

\noindent
For any 3 points $x_i, x_j, x_k \in X$ let 
$\theta_X(j,i,k)$
denote the internal angle at $x_i$ between vectors $(x_j - x_i), (x_k - x_i)$. The corresponding angle in the projection is written $\theta_Y(j,i,k)$.

\noindent
Then, \ourmetricAbbrev\ is defined as:
\begin{equation*}
{\text{\ourmetricAbbrev}}(X,Y, C) = 
\frac{1}{T}
\sum_{\substack{i \in C_a \\ j,k \in C_b \\ j \neq k}}
\left(
\cos\bigl(\theta_X(j,i,k)\bigr) - \cos\bigl(\theta_Y(j,i,k)\bigr)
\right)^2
\end{equation*}
where the sum ranges over all triplets in which the reference point $i$ lies in class $C_a$ and the (unordered) pair $(j,k)$ lies in a different class $C_b$. {$T$ is the total number of triplets considered}, which may be as large as $n \cdot \binom{n/2}{2}$ when the data consists of only two classes of equal size. Therefore, to compute \ourmetricAbbrev\ in full is at least $\mathcal{O}(n^3)$.

In summary, \ourmetricAbbrev\ quantifies how classes are geometrically arranged relative to one another by measuring the preservation of inter-class angle distributions.

We show that restricting the triplets sampled using $C$ instead of sampling from all possible triplets has a considerable effect on this metric in the supplemental material.

\subsection{Design rationale}

At first glance, this expression resembles a cosine-similarity comparison between points in class $C_b$. The crucial difference is that the cosine is taken with respect to a reference point $i$ in a different class $C_a$. For each pair of classes, $(C_a, C_b)$, the set of angles $\theta_X(j,i,k)$ captures how $C_b$ ``looks'' from the perspective of $C_a$. By comparing these angles before and after projection, \ourmetricAbbrev\ measures how faithfully that inter-class geometry is preserved.

This perspective-based formulation has several important consequences. 
If class $C_a$ lies far from class $C_b$ in the high-dimensional space, the angles $\theta_X(j,i,k)$ will be small; \ourmetricAbbrev\ rewards projections that maintain this configuration either by placing $C_b$ farther from $C_a$, by preserving the internal spread of $C_b$, or both. Note that if a class is non-globular (e.g., elongated, curved, nested, intertwined, etc.), the angles seen from neighboring classes encode that shape. In this case, \ourmetricAbbrev\ penalizes projections which collapse these more complex structures into a ``blob'' and rewards projections which more closely preserve the higher-dimensional geometry.

This behavior is illustrated in \autoref{fig:concentricTeaser}. Established class-level metrics like the Silhouette Score favor projections in which all classes appear as tightly packed, well-separated groups, even when the ground-truth structure is nested.
Label T\&C improves on this, but for datasets with nested or concentric structure, it incorrectly rewards scattered or uninformative projections. In contrast, \ourmetricAbbrev\ correctly identifies the projection that best reflects the true relationship among classes.

\subsection{Practical considerations}
We discuss several practical considerations of \ourmetricAbbrev.

\paragraph*{Sampling.}

As noted, the computational complexity to compute the full sum of \ourmetricAbbrev\ is impractical for even medium-sized datasets. 
We observe that it is not necessary to consider all triplets to get a sufficiently accurate result for the metric. 
Instead, we sample $k$ triplets, and would like to choose $k = \mathcal{O}(n)$ (here $n$ is defined as the number of points in the dataset).
Experimentally, we see that this already yields reliable results. 
We analyze the variance in CADI values when sampling different multiples of $n$ across all datasets in our benchmark (see Section~\ref{sec:eval}).
Even for small multiples of $n$, we see consistent results, often only varying after 3 decimal places. An example is shown in 
\autoref{fig:usps_stability}, and figures for all datasets are included 
in the supplemental material. We then choose $k=40n$ for all reported values of \ourmetricAbbrev\ for consistent accuracy in a reasonable time.

\begin{figure*}[t!]
    \centering
    \includegraphics[width=0.52\linewidth]{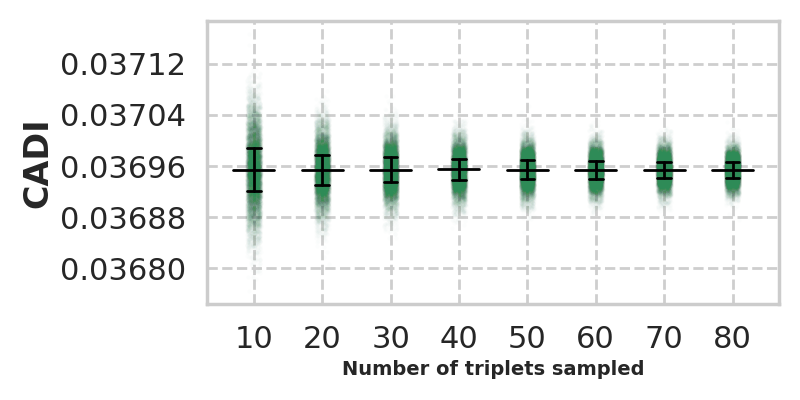}
    \includegraphics[width=0.35\linewidth]{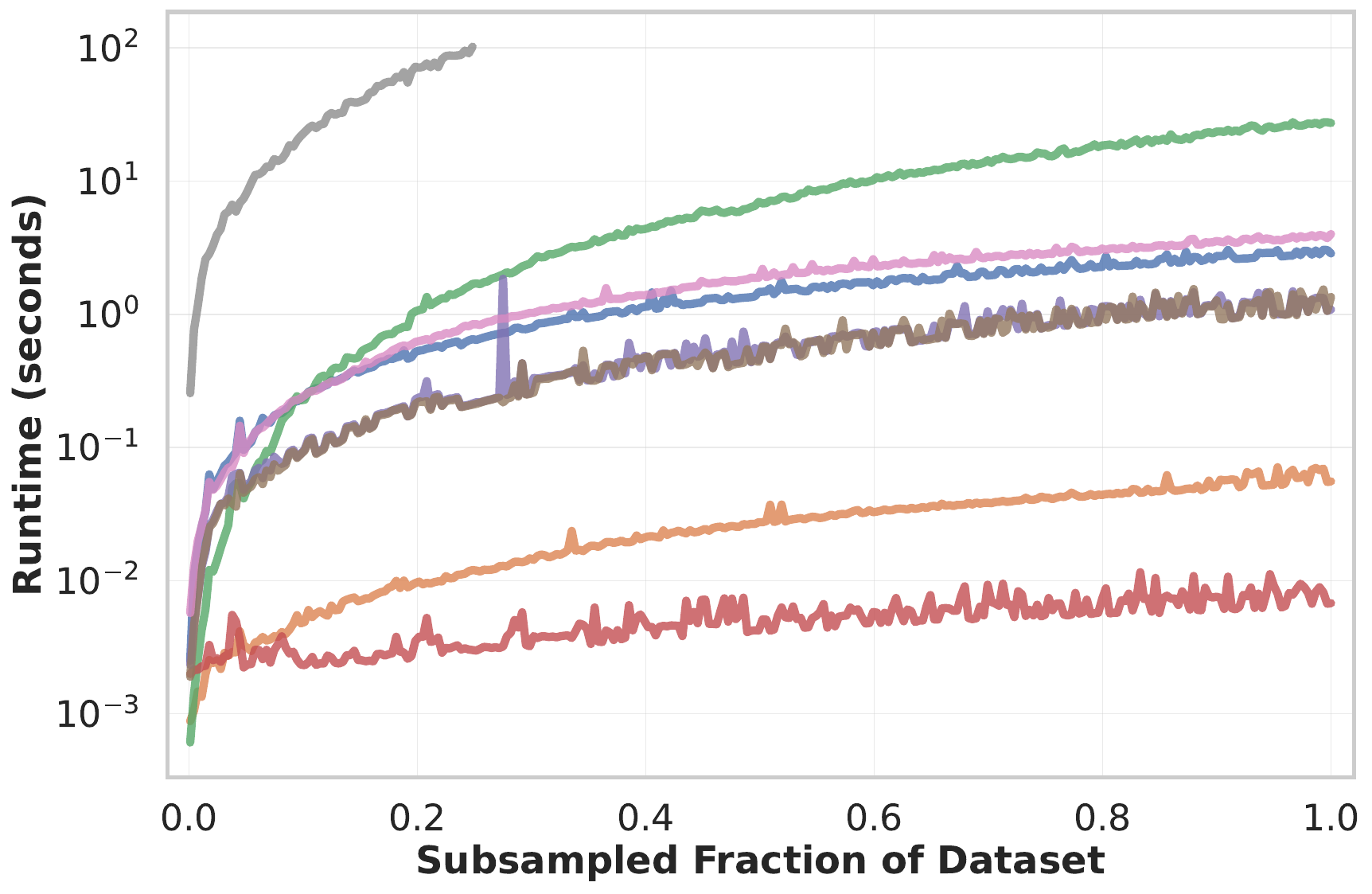}    
    \includegraphics[width=0.12\linewidth]{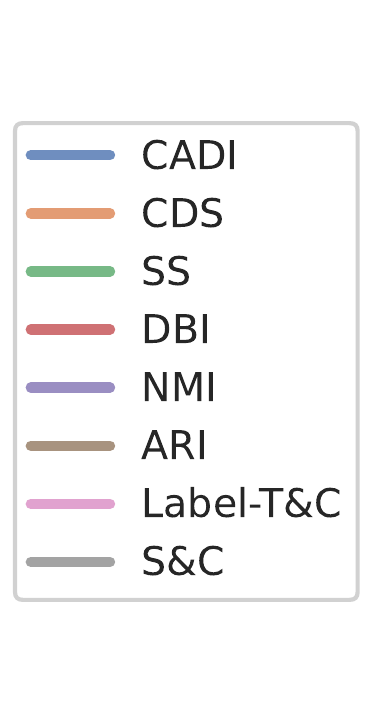}
    \parbox[c]{0.49\linewidth}{\centering (a)}
    \parbox[c]{0.49\linewidth}{\centering (b)}
    \caption{
    (a) For different numbers of triplets sampled, we calculate \ourmetricAbbrev\ for the t-SNE projection of the MNIST dataset on 10,000 different runs and plot their distributions.  
    The black lines mark the interquartile ranges of the score distributions. We notice that sampling $10n$ triplets is already sufficient to achieve a score accurate to 3 decimal digits. Note the small range of the y-axis.
    (b) The times taken on a consumer-grade laptop to evaluate different metrics on differently sized subsamples of the t-SNE projection of the MNIST handwritten digits dataset. 
    }
    \label{fig:usps_stability}
\end{figure*}

Therefore, \ourmetricAbbrev\ can be modified to scale linearly with the number of items in the dataset. We show in \autoref{fig:usps_stability} (b) that \ourmetricAbbrev\ has competitive runtimes compared to existing quality metrics, completing in a little over a second on a consumer-grade laptop for the largest of our datasets. We plot the times taken for each metric on the rest of the datasets in the supplemental material.

\paragraph*{Invariances.}
A good-quality metric should be robust to affine transformations of the data in both high- and low-dimensional spaces, ensuring that conclusions do not depend on arbitrary choices of coordinate systems or positions. Because \ourmetricAbbrev\ relies solely on internal angles, it is invariant under translation, rotation, and uniform scaling. 

{It should be noted that CADI is sensitive to non-linear transformations. This includes feature scaling, a common practice to normalize the dimensions of a dataset as a preprocessing step}. However, this also affects any distance-based metric, and the question of how feature scaling influences projection quality is an active area of research~\cite{dierkes2025towards}.

\paragraph*{Edge Cases.}
A few edge cases must be addressed to ensure the metric is well-defined. We assume the partition contains at least two clusters and that at least one cluster has size greater than two; otherwise, no valid between-cluster triplets exist, and \ourmetricAbbrev\ is undefined. This can be verified with a simple preliminary check.
We also assume that points are in general position, but this may not be true of real data. For triplet configurations in which an angle is undefined, i.e., overlapping points, we define the corresponding angle to be $0$, ensuring the metric remains computationally stable.  

We assume the cluster partition is given as part of the input, either through class labels or through cluster algorithm pre-processing. In practice, this is a nontrivial choice for unlabeled data or for data in which the class labels are not meaningful. A particularly bad partition, i.e., one which is not correlated at all to the data, makes \ourmetricAbbrev\ uninterpretable.

\subsection{Interpretation}

\ourmetricAbbrev\ uses the well-studied residual sum of squares function to quantify the error in a projection. This means the metric is non-negative, and that values closer to $0$ are better. A high \ourmetricAbbrev\ value indicates that for any class $C_a$, there are many points within it for which a second class $C_b$ appears incorrectly shaped. For a 3-dimensional example, a torus-shaped class that has been ``squashed'' into a circular shape in a projection will result in a high \ourmetricAbbrev\ value, since other classes notice it is not torus-like.

The cardinality of each class has an impact on the resulting value of \ourmetricAbbrev. Larger clusters influence the value more than smaller clusters, and this is reinforced further by our sampling approach. While we could normalize the impact by class size, we intentionally do not, as the largest classes offer the most amount of insight into the data.

We note that the full \ourmetricAbbrev\ sum can be further decomposed into class pairs. That is, each pair of clusters $(C_a, C_b)$ contributes to the total sum. Analyzing this carefully provides further insight into the quality of a projection. One can see, for instance, which class is viewing another class incorrectly.

\subsection{AngleEmbedding}
\label{sec:optimization}

Since \ourmetricAbbrev\ is differentiable, we can use it as a loss function to be minimized with gradient descent. This results in a novel supervised DR technique, which we refer to as AngleEmbedding. This serves as an additional understanding of what a low \ourmetricAbbrev\ score means, especially when compared to existing algorithms. We present a method to optimize the function using a parametric optimization scheme.

We employ a simple feedforward neural network as a parametric approach to minimize this loss. Parametric methods for dimensionality reduction have been proposed as early as 2009, with parametric counterparts proposed for both t-SNE~\cite{parametric-tsne} and UMAP~\cite{parametric-umap}. Here, the idea is to use the coordinates of the vectors in the high-dimensional dataset as input to a neural network, which has an output layer corresponding to the target projection dimension (typically 2 or 3). The network is trained individually for every dataset, and after training, the entire dataset is passed through a final time to produce a projection. We chose this framework over a non-parametric method, as we saw better results overall.

We use a simple network architecture consisting of only two hidden layers with 128 units each, which proved to produce projections that sufficiently optimize our metric. 
Developing more sophisticated techniques that better optimize \ourmetricAbbrev\ remains an open direction for further study. We trained the model using the Adam optimizer~\cite{adam_optimization} with a learning rate of $10^{-3}$, which provided the most stable performance in preliminary experiments. No weight decay is applied, as our goal is to fully fit the network to the dataset rather than generalize. Our implementation is available open source; see the link in \autoref{sec:intro}.

\section{Evaluation Design}
\label{sec:eval}
Here we detail the design of our evaluation setup, including the datasets, algorithms, and metrics we use.

\paragraph*{Datasets.}

\begin{table}[t!]
    \small
    \centering
    \caption{Statistics of datasets used in experiments. A ${}^*$ indicates the original data was reduced to that dimensionality through PCA before other algorithms are applied.}
    \label{tab:datasets}

    \begin{tabular}{l l r r r}
        \hline
        \textbf{Name} & \textbf{Type} & \textbf{Samples} & \textbf{Dims} & \textbf{Clusters} \\
        \hline
        concentric3 & synthetic & 2760 & 3 & 5 \\
        concentric4 & synthetic & 3240 & 4 & 5 \\
        donuts & synthetic & 3750 & 3 & 3 \\
        matryoshka & synthetic & 6400 & 3 & 7 \\
        rings & synthetic & 4000 & 100 & 20 \\
        \hline 
        {acl\_imdb}~\cite{aclimdb-dataset} & text & 50000 & 384 & 2 \\
        coil20~\cite{coil20-dataset} & image & 1440 & 16384 & 20 \\
        coil100~\cite{coil100-dataset} & image & 7200 & 16384 & 100 \\
        emotion~\cite{emotion-dataset} & text & 16000 & 384 & 6 \\
        \shortstack[l]{\small F-MNIST~\cite{fashionMNIST-dataset}} & image & 60000 & 784 & 10 \\
        liver~\cite{liver-dataset} & gene & 357 & 50${}^*$ & 2 \\
        MNIST~\cite{mnist-dataset} & image & 60000 & 784 & 10 \\
        olivetti~\cite{olivetti-dataset} & image & 400 & 4096 & 40 \\
        pbmc3k~\cite{pbmc3k-dataset} & gene & 2700 & 2000 & 7 \\
        pendigits~\cite{pendigits-dataset} & image & 10992 & 16 & 10 \\
        penguins~\cite{penguins-dataset} & tabular & 342 & 4 & 3 \\
        sentiment~\cite{sentiment-dataset} & text & 3000 & 384 & 2 \\
        {trec}~\cite{trec-dataset} & text & 15452 & 384 & 6 \\
        usps~\cite{pendigits-dataset} & image & 9298 & 256 & 10 \\
        \hline
    \end{tabular}\vspace{-0.35cm}
\end{table}

We collect and curate a wide variety of synthetic and real datasets to produce an informative benchmark. The datasets span several domains, sizes, and dimensionalities to provide an overview of how cluster-level measures perform. The datasets used are summarized in \autoref{tab:datasets}. We describe the synthetically generated datasets: 

\begin{compactitem}
    \item \textbf{rings}: Consists of 20 two-dimensional rings that are linked together to form a chain. The rings are rotated so that all rings are mutually orthogonal, resulting in an intrinsic dimensionality of 21. Additionally, Gaussian noise is added in 100-dimensional space.
    \item \textbf{concentric3} and \textbf{concentric4}: Consists of 5 three- and four-dimensional hollow (hyper)spheres, with each sphere containing 552 and 648 points for concentric3 and concentric4, respectively.
    \item \textbf{matryoshka}: Emulating nested Matryoshka dolls, this dataset consists of 3 dumbbell-shaped surfaces that are nested one inside the other. Inside the innermost dumbbell, we place 2 hollow concentric spheres at either end of the dumbbell. Each dumbbell and sphere form a separate class.
    \item \textbf{donuts}: Consists of 3 nested tori in 3D. Each torus consists of 1250 points.
\end{compactitem}
Each synthetic dataset is meant to emphasize a failure point of common cluster algorithms, which do not account for non-globular cluster shapes. 

\paragraph*{Algorithms.}

We project each of the above datasets to 2D with 6 different dimensionality reduction techniques:

\begin{compactitem}
    \item \textbf{AngleEmbedding}, which optimizes \ourmetricAbbrev\ as described in Section~\ref{sec:optimization}. As it is directly optimized, we expect AngleEmbedding to have the lowest \ourmetricAbbrev\ score.
    \item \textbf{MDS}~\cite{borg2005modern} is included as a representative global non-linear method. It often struggles to maintain cluster separation and is susceptible to failure due to distance concentration in very high-dimensional spaces. Still, for low-dimensional and simple datasets, we expect MDS to maintain a low \ourmetricAbbrev\ score. 
    \item \textbf{t-SNE}~\cite{maaten2008tsne} and \textbf{UMAP}~\cite{mcinnes2018umap} are included as two local non-linear methods. They are known to suffer from cluster relationship distortions, and we expect them to have relatively high \ourmetricAbbrev\ scores.
    \item \textbf{PaCMAP}~\cite{pacmap_randomtripletloss} and \textbf{UMATO}~\cite{UMATO} are methods that aim to preserve both local and global relationships. 
    \item \textbf{PCA}~\cite{pearson1901principal}, the classical linear method
    \item {\textbf{Random} is simply found by assigning positions uniformly at random in a unit square. We use this as a baseline.}
\end{compactitem}
This forms a representative collection of DR techniques and allows us to ground our metric with well-understood DR projections.

\paragraph*{Cluster-level metrics used.}

To assess the proposed \ourmetricFull\ as a cluster-level metric, we analyze its results on projections of the above datasets alongside the scores provided by established cluster-level metrics. These metrics capture different aspects of a projection, including cluster separability, preservation of class structure, and global inter-cluster relationships. The metrics used are summarized on \autoref{tab:metric-summary} and are further detailed below.

\begin{table}[t!]
\centering
\caption{Summary of cluster-level metrics used. Measures where higher is better are indicated by an up arrow $\uparrow$ while metrics where lower is better are indicated by a down arrow $\downarrow$.}
\label{tab:metric-summary}
\begin{tabular}{l r r}
\hline
\textbf{Name of Metric} & \textbf{Range} & \\
\hline
Cluster Distance Score (CDS)       & $0 \le$ & $\downarrow$ \\
Silhouette Score (SS)              & $[-1, 1]$ & $\uparrow$ \\
Davies--Bouldin Index (DBI)        & $0 \le$ & $\downarrow$ \\
Normalized Mutual Information (NMI)& $[0,1]$ & $\uparrow$ \\
Adjusted Rand Index (ARI)          & $[-0.5,1[$ & $\uparrow$ \\
Label-T\&C (L-T\&C)    & $[0,1]$ & $\uparrow$ \\
Steadiness \& Cohesiveness (S\&C)  & $[0,1]$ & $\uparrow$ \\
\hline
\end{tabular}
\end{table}
\begin{compactitem}
    
    \item \textbf{Cluster Distance Score (CDS)}~\cite{Jacob2023LGS} — Measures how well distances between cluster centroids in the high-dimensional space are preserved in the projection. This metric only measures global geometric consistency of cluster centroids. This metric is degenerate when there are fewer than 3 clusters. 
    We implement a scale-invariant version following Smelser et al.~\cite{sns_snkl}.

    \item \textbf{Silhouette Score (SS)}~\cite{SilhouetteScore1987} and \textbf{Davies--Bouldin Index (DBI)}~\cite{DaviesBouldinIndex1979} — Both metrics assess cluster separability by comparing intra-cluster distances with inter-cluster distances. While \textbf{SS} measures the similarity of a point to its own cluster versus all other clusters, the \textbf{DBI} only measures the average similarity between each cluster and its most similar one. While effective for evaluating separation, they inherently assume that clusters are globular and well-separated, and may penalize non-convex or interacting clusters.

    \item \textbf{Normalized Mutual Information (NMI)}~\cite{normalizedMutualInformation2005} and \textbf{Adjusted Rand Index (ARI)}~\cite{adjustedrandindex1985} — These metrics quantify the agreement between clustering assignments in the projection and known ground-truth labels. In our experiments, we use HDBSCAN for clustering. Unlike the previous pair, when paired with HDBSCAN, these metrics can identify non-convex clusters when they are well separated. However, these metrics cannot distinguish between geometrically distinct but label-equivalent clusterings, and additionally may get confused by overlapping or interacting clusters.

    \item \textbf{Label Trustworthiness and Continuity (Label T\&C) or L-T\&C}~\cite{jeon2023classesclusters} — 
    This metric is the harmonic mean of two different metrics: Label Trustworthiness and Label Continuity. Label Trustworthiness checks whether clusters that should be well-separated in the projection are well-separated, while Label Continuity penalizes well-separated clusters in the projection if they aren't as well-separated in the dataset. Therefore, this metric accounts for overlapping clusters in the dataset, but is still blind to cluster structure due to its reliance on cluster centroids.

    \item {
    \textbf{Steadiness and Cohesiveness (S\&C)}~\cite{jeon2021measuring} — This is also the harmonic mean of two different metrics. Steadiness measures whether clusters formed in the projection are truly coherent in the original space, while Cohesiveness measures whether original clusters remain intact in the projection. A nearest-neighbors-based clustering algorithm is used to capture possibly non-convex clusters, but this metric can also get confused by complicated cluster structures. We note that it is not possible to inform S\&C using class labels in its current implementation.
    }

\end{compactitem}

\begin{figure*}[t!]
    \centering

    \begin{minipage}[c]{0.32\linewidth}
        \centering
        \fbox{\includegraphics[width=\linewidth,trim=50pt 50pt 20pt 70pt, clip]{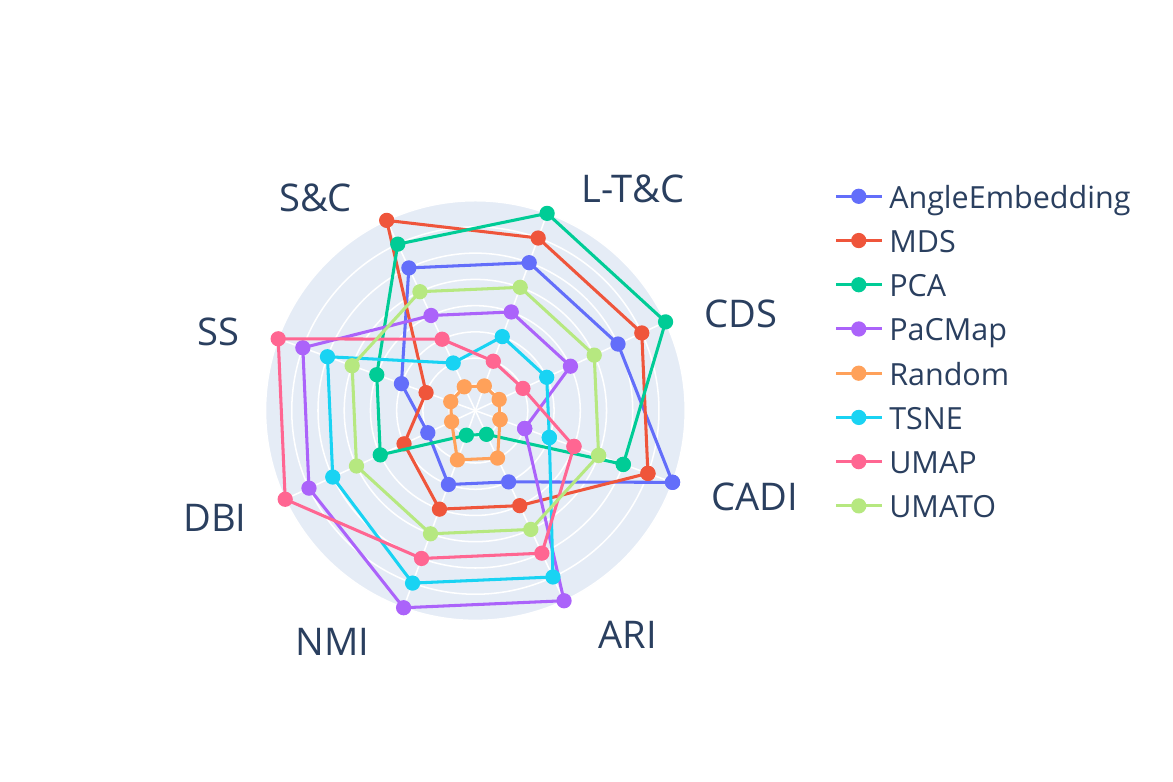}} \vspace{-0.5cm}
        \caption*{rings}
    \end{minipage}
    \hfill
    \begin{minipage}[c]{0.67\linewidth}
        \centering
        \includegraphics[width=0.24\linewidth]{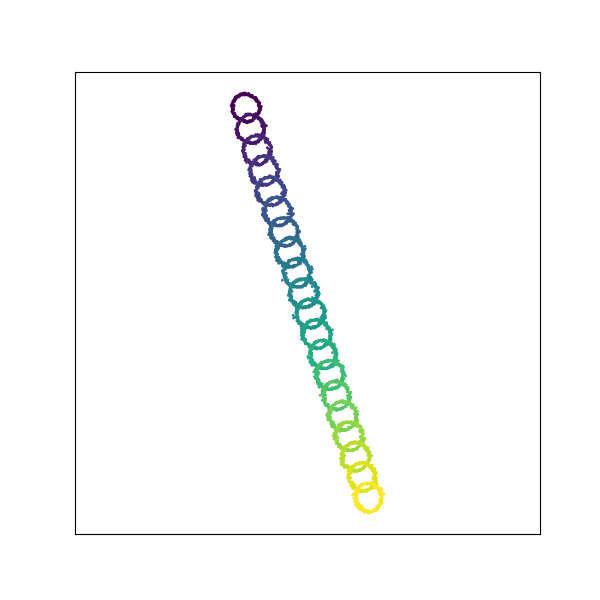}
        \includegraphics[width=0.24\linewidth]{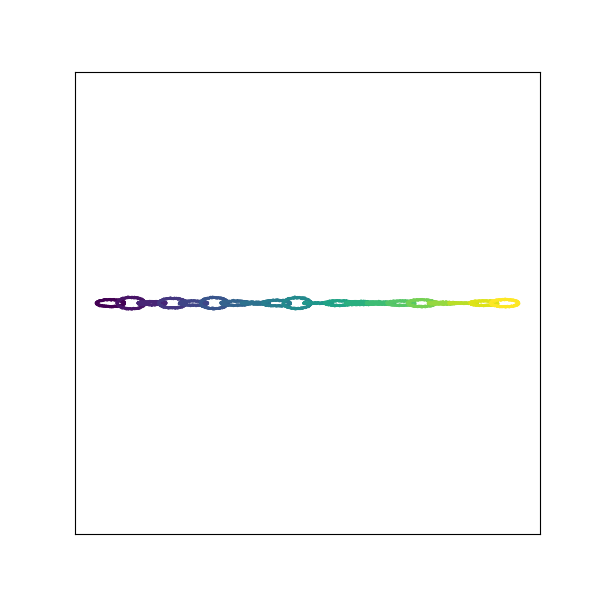}  
        \includegraphics[width=0.24\linewidth]{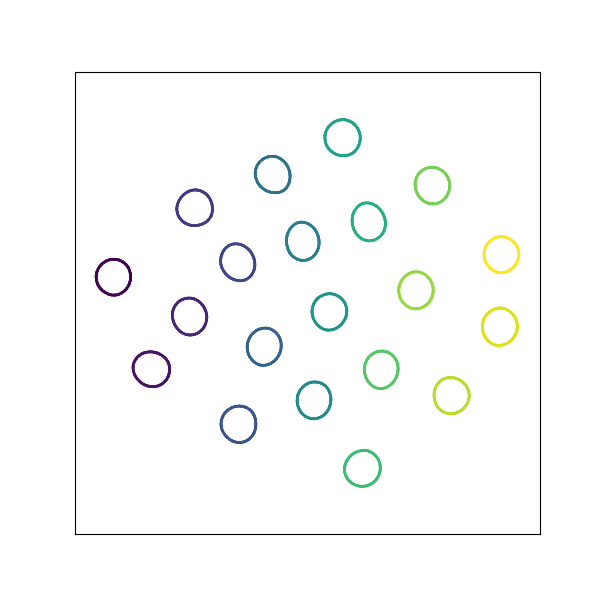}
        \includegraphics[width=0.24\linewidth]{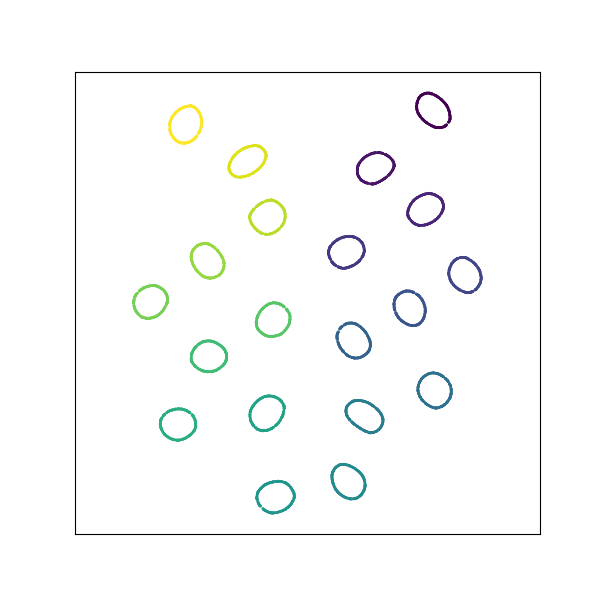}
    
        \parbox[c]{0.24\linewidth}{\centering AngleEmbedding}
        \parbox[c]{0.24\linewidth}{\centering PCA}
        \parbox[c]{0.24\linewidth}{\centering t-SNE}
        \parbox[c]{0.24\linewidth}{\centering UMAP}         
    \end{minipage}

    \begin{minipage}[c]{0.32\linewidth}
        \centering
        \fbox{\includegraphics[width=\linewidth,trim=50pt 50pt 20pt 70pt, clip]{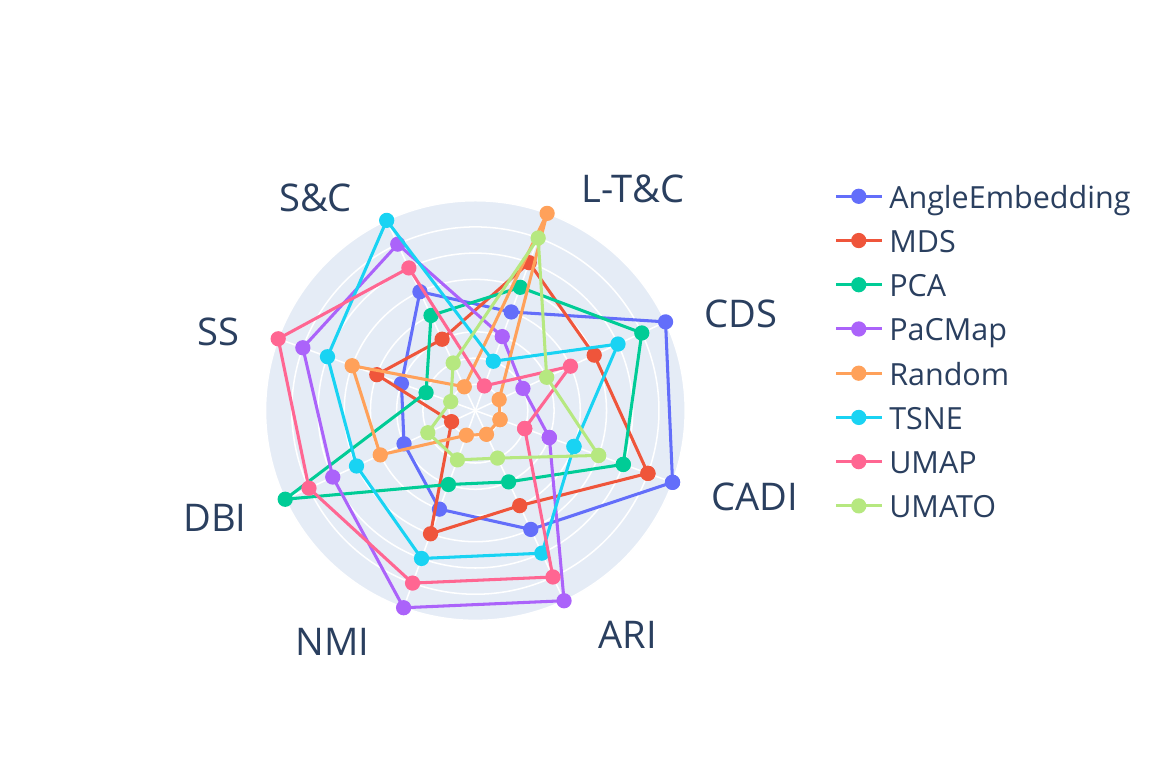}} \vspace{-0.5cm}
        \caption*{concentric3}
    \end{minipage}
    \hfill
    \begin{minipage}[c]{0.67\linewidth}
    \centering
    \includegraphics[width=0.24\linewidth]{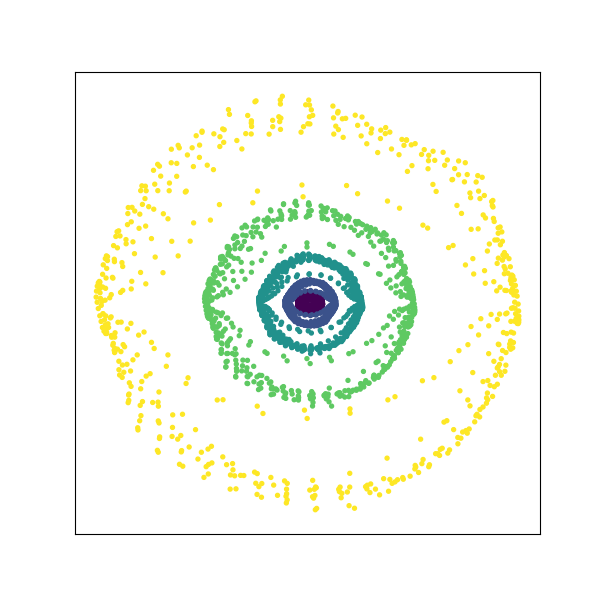}
    \includegraphics[width=0.24\linewidth]{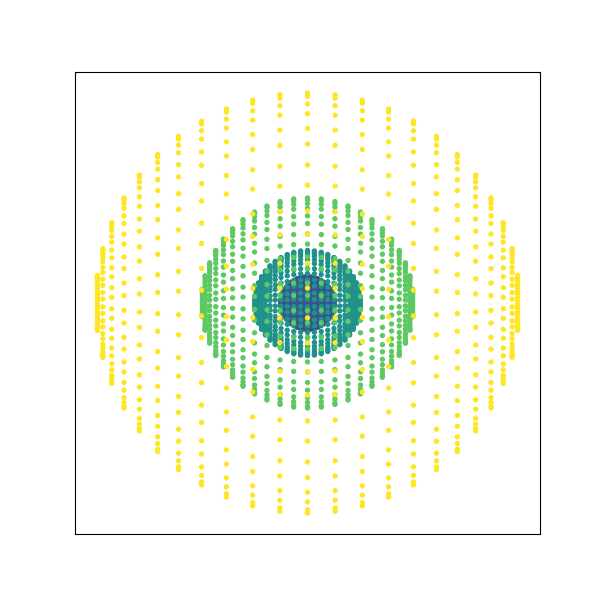}  
    \includegraphics[width=0.24\linewidth]{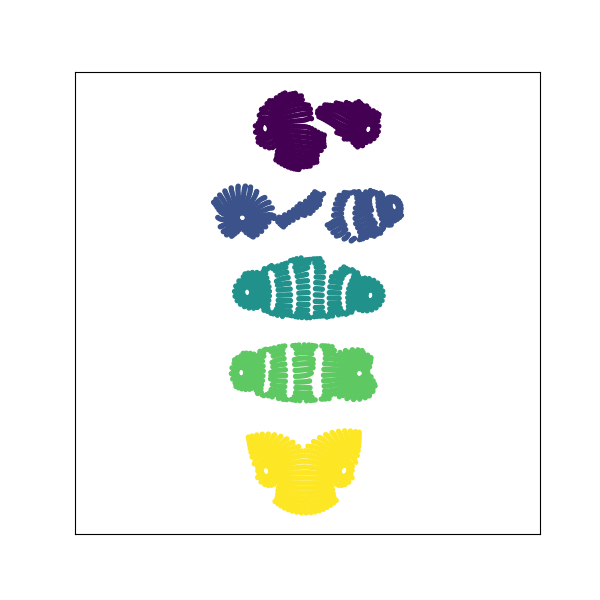}
    \includegraphics[width=0.24\linewidth]{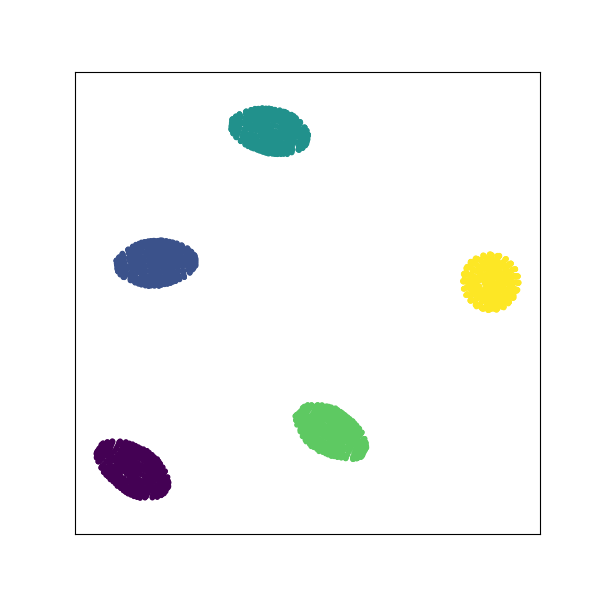}

    \parbox[c]{0.24\linewidth}{\centering AngleEmbedding}
    \parbox[c]{0.24\linewidth}{\centering PCA}
    \parbox[c]{0.24\linewidth}{\centering t-SNE}
    \parbox[c]{0.24\linewidth}{\centering UMAP}         
    \end{minipage}

    \caption{Projections of the \textbf{rings} (top) and \textbf{concentric3} (bottom) datasets created by different DR algorithms.}
    \label{fig:concentric3-eval}
    \vspace{-0.4cm}
\end{figure*}

{
Collectively, these metrics measure either global cluster arrangement or cluster separability. Measures that do not involve cluster-level separation, such as neighborhood hit or stress, are not included due to differing objectives. 
}
Our \ourmetricFull\ is designed to capture aspects of the projection that are not fully assessed by existing metrics, particularly the preservation of cluster manifolds and interactions that may occur between clusters.

\paragraph*{Implementation details.}

For the projection techniques, we use Scikit-learn's implementations of t-SNE, MDS, and PCA, and use the author's implementation of UMAP. We use the ZADU library~\cite{zadu} for Label-T\&C, and our own implementation of CDS. All other metrics use the Scikit-learn implementation. We use AngleEmbedding, setting the number of sampled triplets to $10n$, where $n$ is the number of points in a given dataset.
All experiments were conducted on a machine with an AMD Ryzen 9 7900X3D 12-core processor (24 threads, 5.66 GHz max frequency) and 128 GB of RAM, running Rocky Linux 9.6.

\section{Evaluation Results}

\begin{table}[t!]
        \caption{Spearman rank correlation between cluster-level metrics.}
    \centering

\scriptsize
\begin{tabular}{lcccccccc}
\hline
 & {\tiny \ourmetricAbbrev} & {\tiny CDS} & {\tiny L-T\&C} & {\tiny S\&C} & {\tiny SS} & {\tiny DBI} & {\tiny NMI} & {\tiny ARI} \\ \hline
{\tiny \ourmetricAbbrev} & \cellcolor[rgb]{0.086,0.377,0.663}{\textcolor{white}{1.00}} & \cellcolor[rgb]{0.562,0.756,0.860}{\textcolor{black}{0.20}} & \cellcolor[rgb]{0.215,0.503,0.738}{\textcolor{white}{0.80}} & \cellcolor[rgb]{0.453,0.690,0.835}{\textcolor{black}{0.41}} & \cellcolor[rgb]{0.543,0.744,0.856}{\textcolor{black}{0.24}} & \cellcolor[rgb]{0.493,0.714,0.844}{\textcolor{black}{0.33}} & \cellcolor[rgb]{0.422,0.671,0.827}{\textcolor{black}{0.46}} & \cellcolor[rgb]{0.453,0.689,0.835}{\textcolor{black}{0.41}} \\
{\tiny CDS} & - & \cellcolor[rgb]{0.086,0.377,0.663}{\textcolor{white}{1.00}} & \cellcolor[rgb]{0.591,0.768,0.871}{\textcolor{black}{0.14}} & \cellcolor[rgb]{0.568,0.758,0.862}{\textcolor{black}{0.19}} & \cellcolor[rgb]{0.595,0.770,0.872}{\textcolor{black}{0.13}} & \cellcolor[rgb]{0.551,0.749,0.858}{\textcolor{black}{0.22}} & \cellcolor[rgb]{0.719,0.825,0.918}{\textcolor{black}{-0.16}} & \cellcolor[rgb]{0.708,0.820,0.914}{\textcolor{black}{-0.13}} \\
{\tiny L-T\&C} & - & - & \cellcolor[rgb]{0.086,0.377,0.663}{\textcolor{white}{1.00}} & \cellcolor[rgb]{0.460,0.694,0.836}{\textcolor{black}{0.39}} & \cellcolor[rgb]{0.490,0.712,0.843}{\textcolor{black}{0.34}} & \cellcolor[rgb]{0.489,0.712,0.843}{\textcolor{black}{0.34}} & \cellcolor[rgb]{0.430,0.676,0.829}{\textcolor{black}{0.45}} & \cellcolor[rgb]{0.457,0.692,0.836}{\textcolor{black}{0.40}} \\
{\tiny S\&C} & - & - & - & \cellcolor[rgb]{0.086,0.377,0.663}{\textcolor{white}{1.00}} & \cellcolor[rgb]{0.484,0.709,0.842}{\textcolor{black}{0.35}} & \cellcolor[rgb]{0.516,0.728,0.850}{\textcolor{black}{0.29}} & \cellcolor[rgb]{0.434,0.678,0.830}{\textcolor{black}{0.44}} & \cellcolor[rgb]{0.439,0.681,0.831}{\textcolor{black}{0.43}} \\
{\tiny SS} & - & - & - & - & \cellcolor[rgb]{0.086,0.377,0.663}{\textcolor{white}{1.00}} & \cellcolor[rgb]{0.268,0.551,0.766}{\textcolor{white}{0.72}} & \cellcolor[rgb]{0.381,0.646,0.817}{\textcolor{black}{0.54}} & \cellcolor[rgb]{0.328,0.606,0.798}{\textcolor{white}{0.63}} \\
{\tiny DBI} & - & - & - & - & - & \cellcolor[rgb]{0.086,0.377,0.663}{\textcolor{white}{1.00}} & \cellcolor[rgb]{0.422,0.670,0.827}{\textcolor{black}{0.47}} & \cellcolor[rgb]{0.397,0.655,0.821}{\textcolor{black}{0.51}} \\
{\tiny NMI} & - & - & - & - & - & - & \cellcolor[rgb]{0.086,0.377,0.663}{\textcolor{white}{1.00}} & \cellcolor[rgb]{0.139,0.430,0.695}{\textcolor{white}{0.91}} \\
{\tiny ARI} & - & - & - & - & - & - & - & \cellcolor[rgb]{0.086,0.377,0.663}{\textcolor{white}{1.00}} \\
\hline
\end{tabular}

    \vspace{-0.3cm}
    \label{tab:metric_corr}
\end{table}
{

We compute the projections with all the mentioned dimensionality reduction algorithms and evaluate scores on all metrics. \autoref{tab:metric_corr} shows the Spearman rank correlations of the scores between metrics, and it is clear that while there is generally positive correlation between metric scores, the relatively low correlations show that \ourmetricAbbrev\ is indeed measuring something different from the other cluster metrics. We would like to note the high correlation between \ourmetricAbbrev\ and Label-T\&C; we attribute this to the fact that Label-T\&C takes into account that points in a cluster are not necessarily tightly-packed around their centroid, and therefore gets fooled less often by non-globular clusters. However, since Label-T\&C still mainly only uses cluster centroids, it is still susceptible to fail on instances where \ourmetricAbbrev\ is more robust; see Figure~\ref{fig:concentricTeaser}.

While we report all relevant projections and scores in the supplementary material, we also analyze a few examples below to showcase these failure modes, and show that the \ourmetricFull\ is robust to such issues. For each dataset, we provide a radar chart that plots the ranking of each projection according to all metrics. 
}

\subsection{Synthetic data}
Here, we present a quantitative analysis of various synthetic datasets, in which the true underlying structure is known. Each dataset is carefully crafted to violate the globularity assumption while still maintaining a simple, easy-to-interpret cluster-level structure. A good cluster-level preserving projection should make these structures visually evident. 

\paragraph*{Rings.}
The rings dataset is composed of a 100-dimensional chain of 2D rings, arranged on a straight line in the ambient space; see \autoref{fig:concentric3-eval} (top).
global methods such as PCA correctly embed the rings along a straight line and therefore obtain strong Cluster Distance Scores. However, the individual rings are heavily distorted, and cluster sizes become uneven. Local methods such as t-SNE produce separated, ring-shaped clusters with a scattered left-to-right ordering, while UMAP arranges the rings in a U-shaped configuration. In all three cases, the interlocked chain structure is not preserved.

Metrics focusing on cluster separation (SS, DBI, NMI, ARI) favor t-SNE, UMAP, and PaCMAP, despite their unlinked rings. Label-T\&C ranks MDS, PCA, and AngleEmbedding relatively equally, as it relies on centroid relationships and is therefore insensitive to the distortion of the rings. S\&C similarly favors MDS and PCA, ranking them slightly above AngleEmbedding despite their distortion. CADI assigns the best score to AngleEmbedding, which preserves 20 equally sized, interlinked rings arranged linearly with minimal distortion, while projections that break the interlocked structure (t-SNE, UMAP, PaCMAP) receive the lowest scores after Random.

\paragraph*{Concentric data.}

On the concentric3 and concentric4 datasets, we again observe that local methods such as t-SNE and UMAP separate the clusters well, forming ``blobs'', but fail to communicate the nested structure of the clusters; see \autoref{fig:concentric3-eval} (bottom). The standard cluster separation metrics (SS, DBI, NMI, and ARI), as well as S\&C, again rank these projections the highest.

However, PCA and AngleEmbedding preserve the nested structure, producing concentric rings; AngleEmbedding additionally reveals the spherical geometry through ``latitude lines''. Only \ourmetricAbbrev\ and Label-T\&C recognize these projections with higher scores; however, as seen in \autoref{fig:concentricTeaser}, Label-T\&C also assigns the best scores to the Random projection, as it assumes that all clusters fully overlap due to the fact that the cluster centroids almost coincide.

\paragraph*{Other synthetic data.}
We note that this pattern continues in the remaining two synthetic datasets. 
UMAP, t-SNE, and PaCMAP
continue to separate clusters into uninformative ``blobs'', and yet receive high scores on SS, DBI, NMI, and ARI, while the more cluster-level faithful projections by MDS, PCA, and AngleEmbedding are scored badly. Again, Label-T\&C gives better scores to MDS, PCA, and AngleEmbedding, but also scores the random projection highly. Only \ourmetricAbbrev\ ranks the projections in a consistent order. See the supplemental material for more details.

\subsection{Real data}

{

\begin{figure*}[h]
    \centering
    \begin{minipage}[c]{0.32\linewidth}
        \centering
        \fbox{\includegraphics[width=\linewidth,trim=50pt 50pt 20pt 70pt, clip]{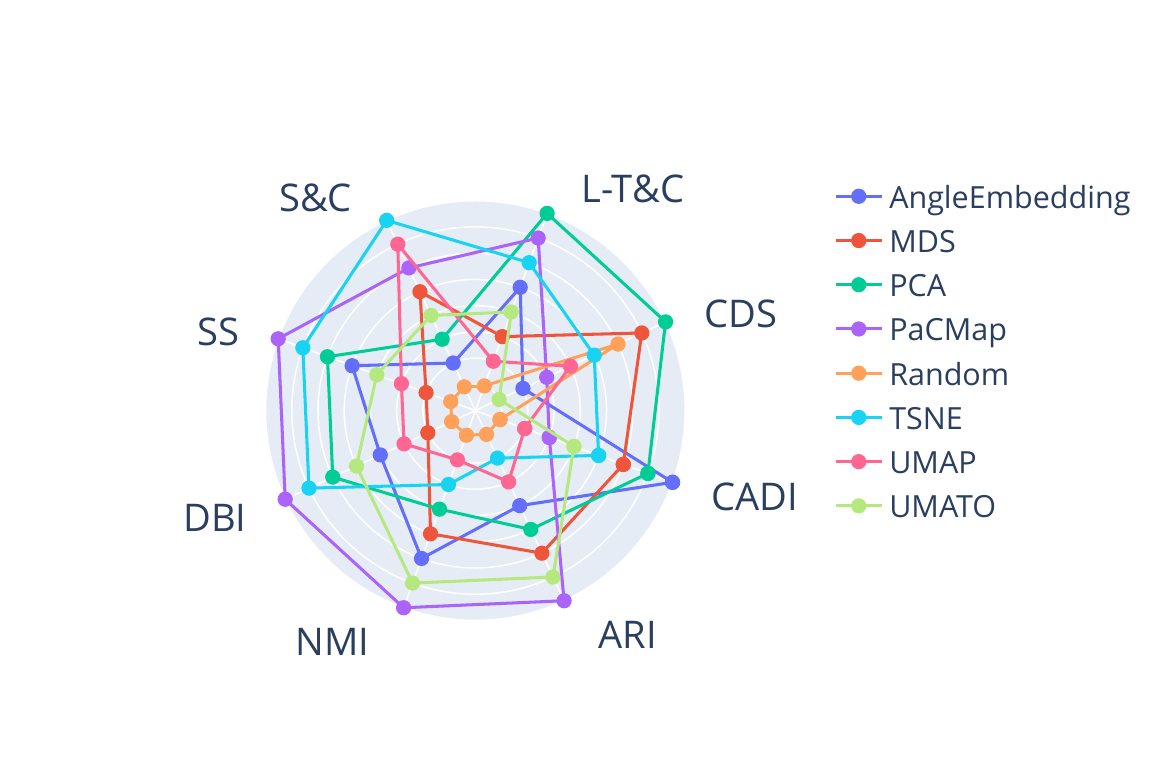}} \vspace{-0.5cm}
        \caption*{liver}
    \end{minipage}
    \hfill
    \begin{minipage}[c]{0.67\linewidth}
        \centering
        \includegraphics[width=0.24\linewidth]{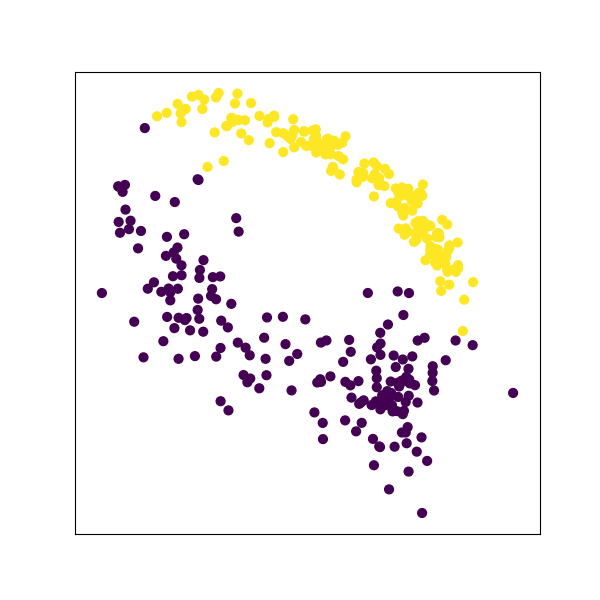}
        \includegraphics[width=0.24\linewidth]{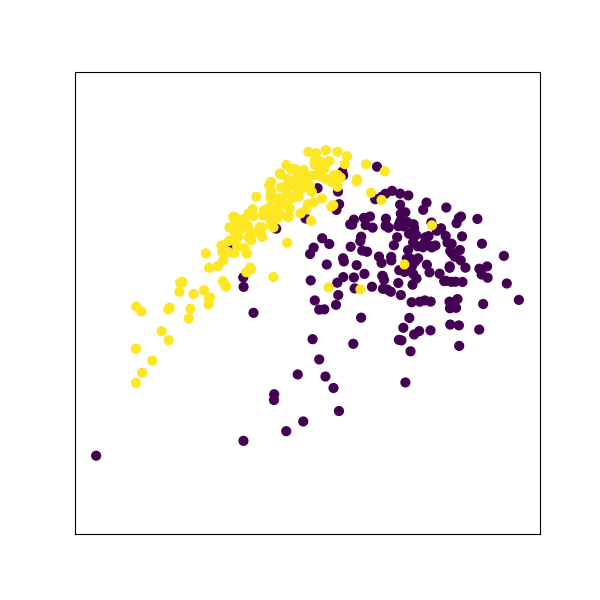}  
        \includegraphics[width=0.24\linewidth]{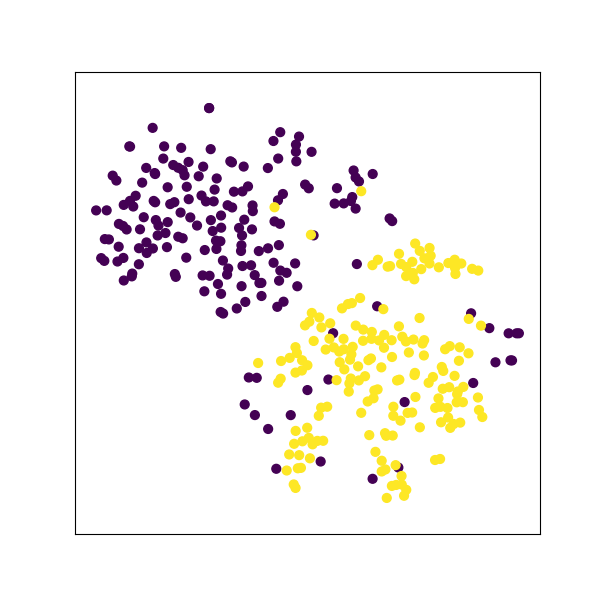}
        \includegraphics[width=0.24\linewidth]{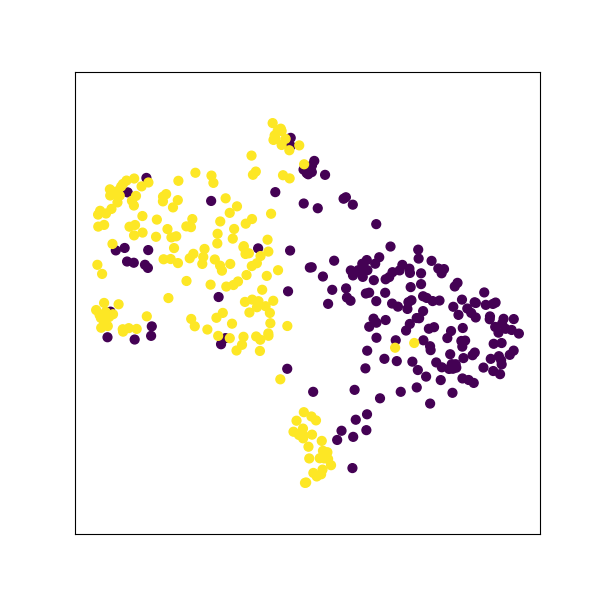}
    
        \parbox[c]{0.24\linewidth}{\centering  AngleEmbedding}
        \parbox[c]{0.24\linewidth}{\centering  PCA}
        \parbox[c]{0.24\linewidth}{\centering  t-SNE}
        \parbox[c]{0.24\linewidth}{\centering  UMAP}         
    \end{minipage}     

    \begin{minipage}[c]{0.32\linewidth}
        \centering
        \fbox{\includegraphics[width=\linewidth,trim=50pt 50pt 20pt 70pt, clip]{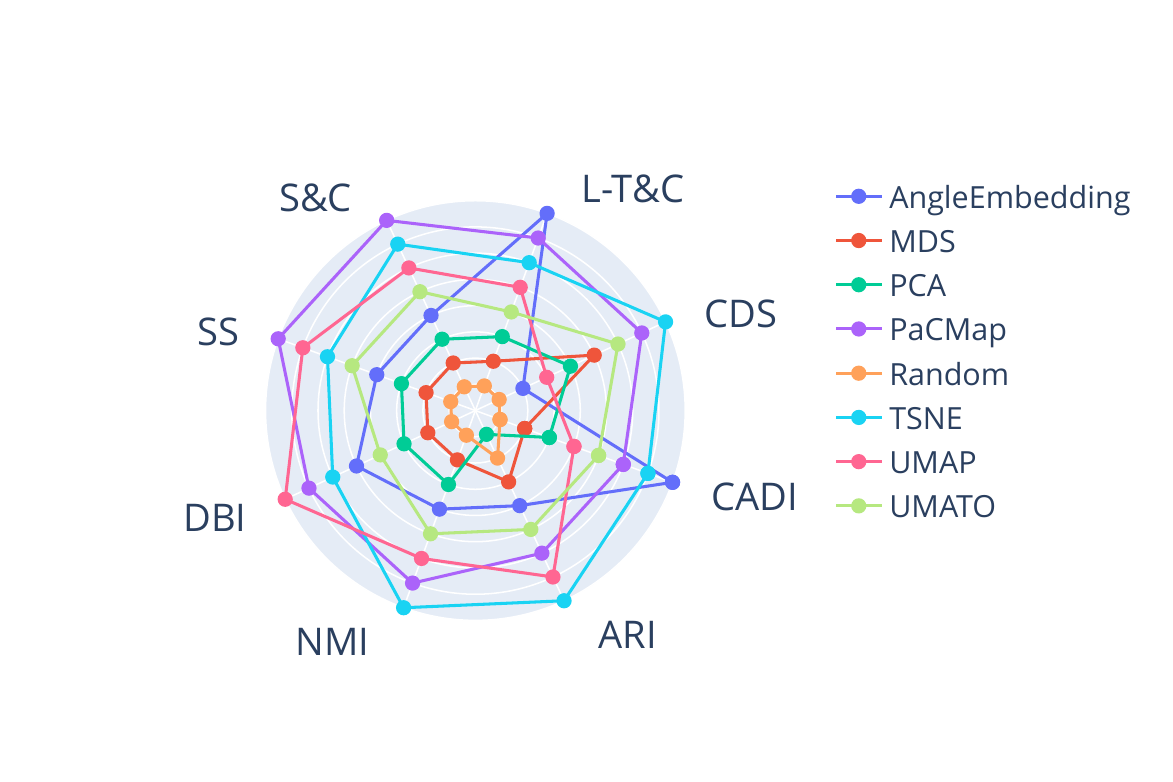}} \vspace{-0.5cm}
        \caption*{usps}
    \end{minipage}
    \hfill
    \begin{minipage}[c]{0.67\linewidth}
        \centering
        \includegraphics[width=0.24\linewidth]{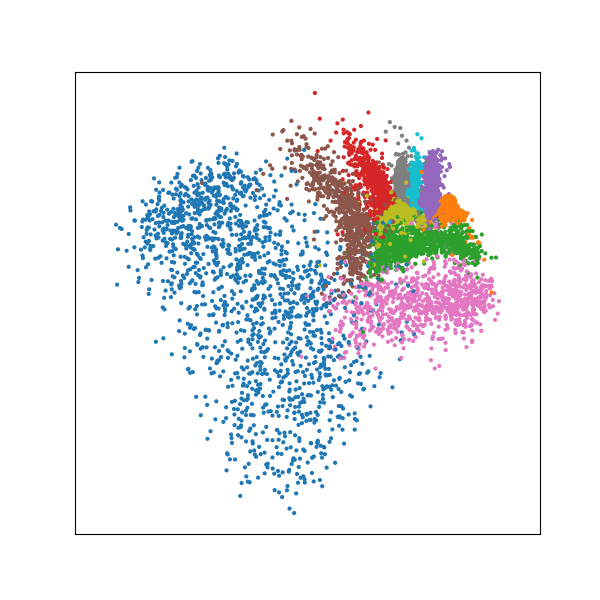}
        \includegraphics[width=0.24\linewidth]{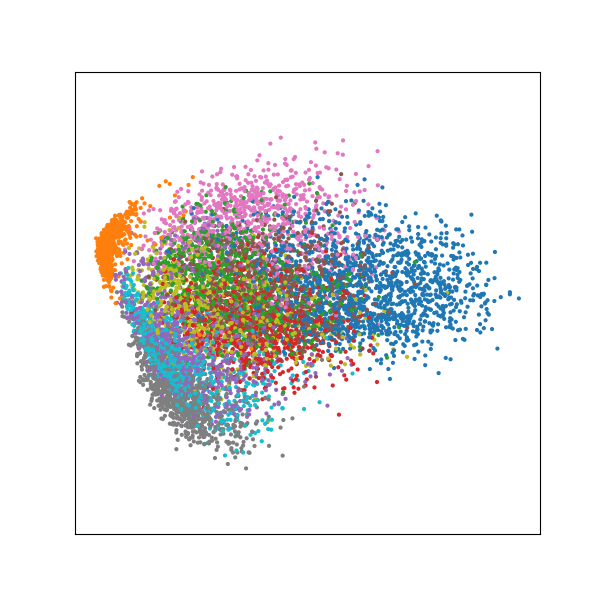}  
        \includegraphics[width=0.24\linewidth]{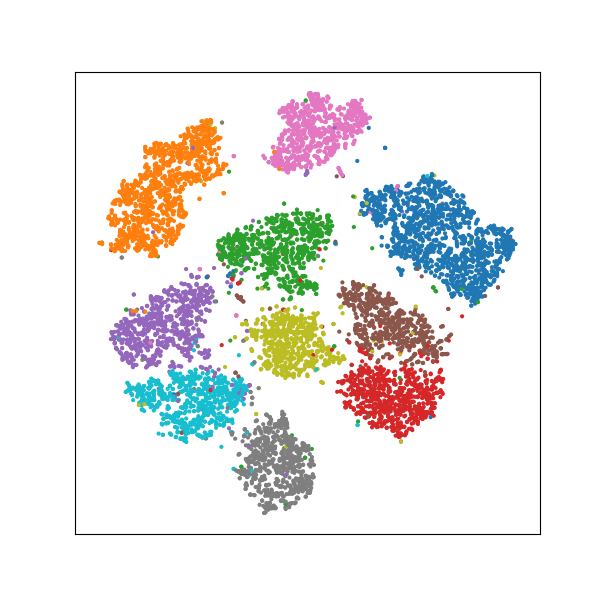}
        \includegraphics[width=0.24\linewidth]{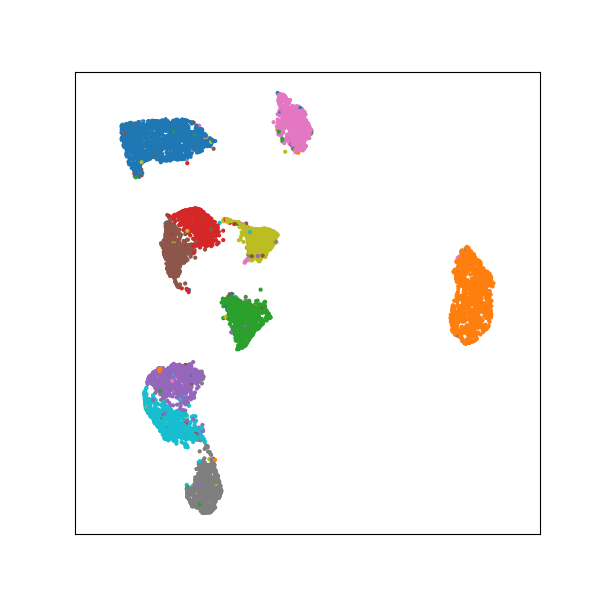}
    
        \parbox[c]{0.24\linewidth}{\centering  AngleEmbedding}
        \parbox[c]{0.24\linewidth}{\centering  PCA}
        \parbox[c]{0.24\linewidth}{\centering  t-SNE}
        \parbox[c]{0.24\linewidth}{\centering  UMAP}         
    \end{minipage}

    \begin{minipage}[c]{0.32\linewidth}
        \centering
        \fbox{\includegraphics[width=\linewidth,trim=50pt 50pt 20pt 70pt, clip]{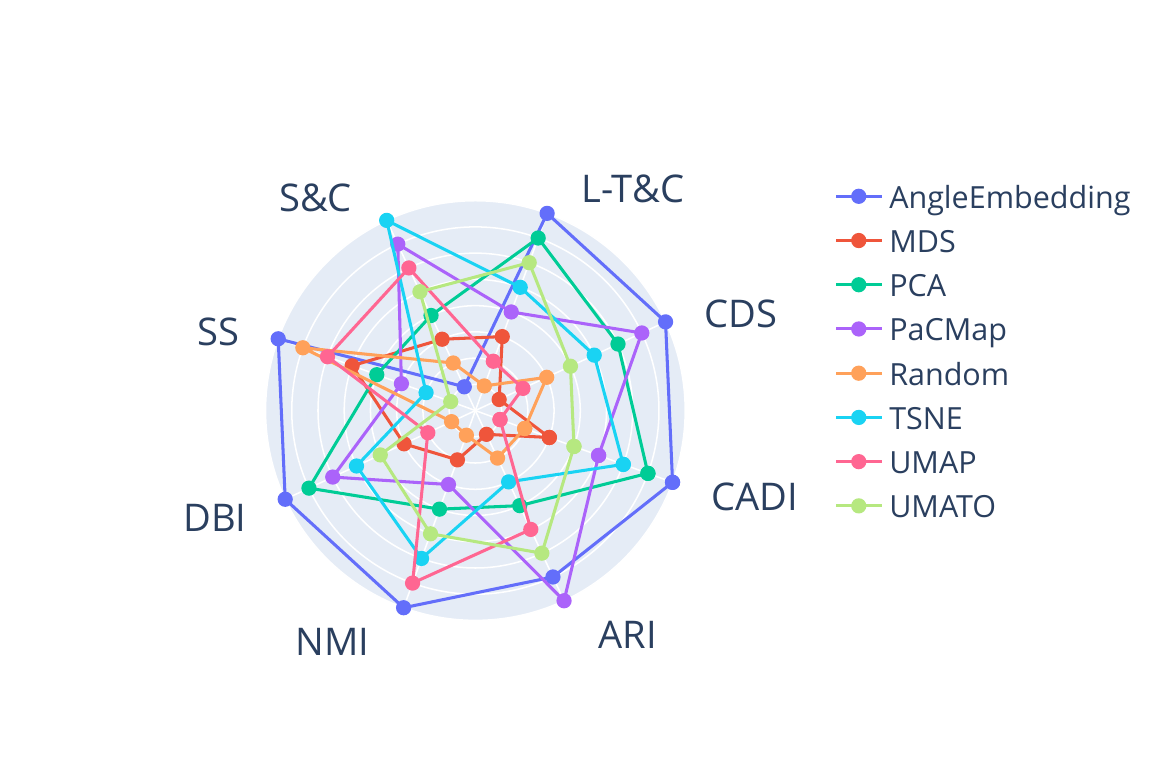}} \vspace{-0.5cm}
        \caption*{trec}
    \end{minipage}
    \hfill
    \begin{minipage}[c]{0.67\linewidth}
        \centering
        \includegraphics[width=0.24\linewidth]{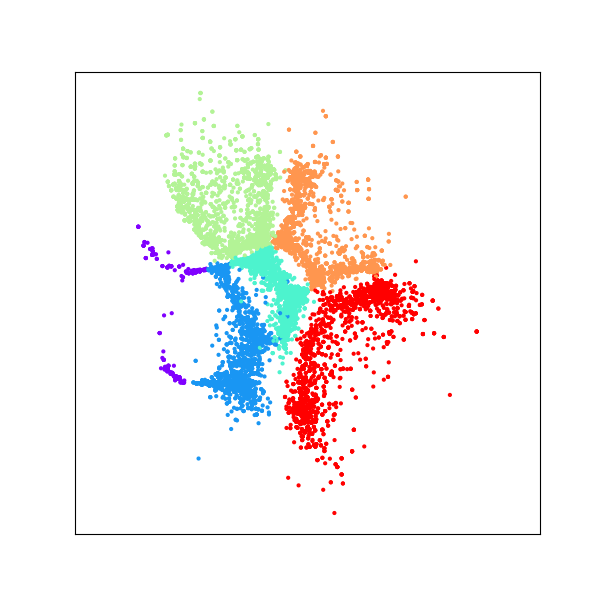}
        \includegraphics[width=0.24\linewidth]{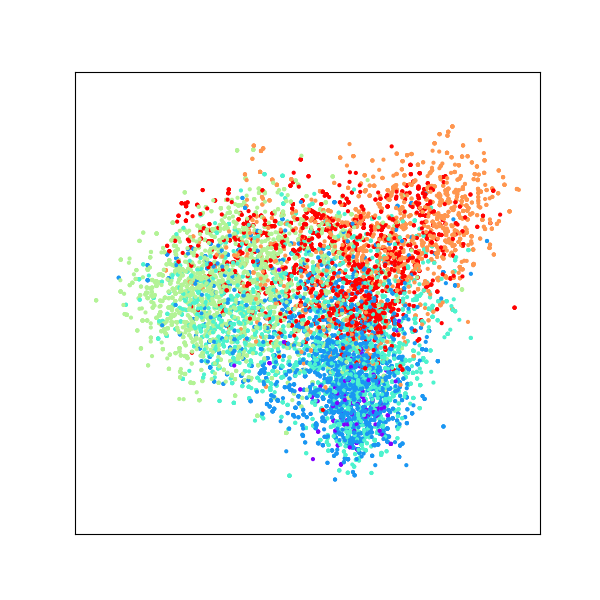}  
        \includegraphics[width=0.24\linewidth]{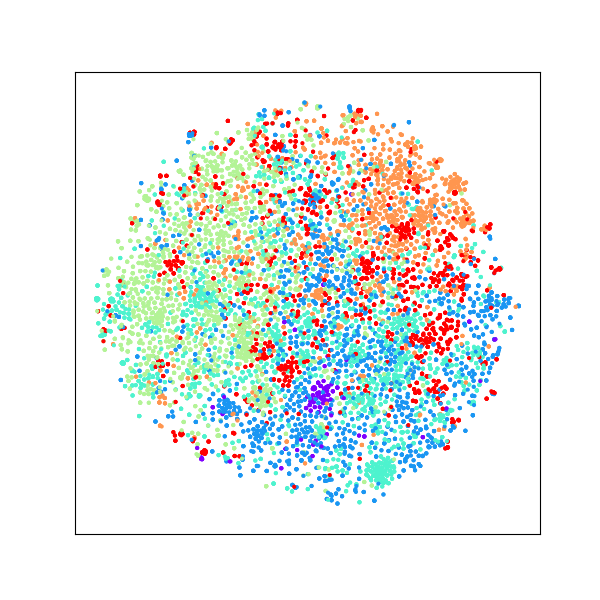}
        \includegraphics[width=0.24\linewidth]{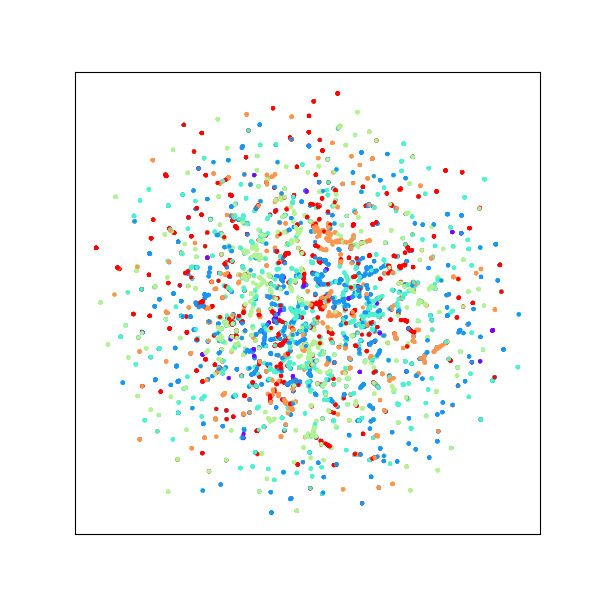}
    
        \parbox[c]{0.24\linewidth}{\centering  AngleEmbedding}
        \parbox[c]{0.24\linewidth}{\centering  PCA}
        \parbox[c]{0.24\linewidth}{\centering t-SNE}
        \parbox[c]{0.24\linewidth}{\centering  UMAP}         
    \end{minipage}    

    \caption{Projections of the \textbf{liver} (top), \textbf{usps} (middle), and \textbf{trec} (bottom) datasets created by different DR algorithms.
    }
    \label{fig:emotion-eval}
    \vspace{-0.4cm}
\end{figure*}

We also analyze several real datasets to show how \ourmetricAbbrev\ can be used to identify cluster-level structure.

\paragraph*{Liver data.}
The liver dataset contains gene expression profiles of both cancerous and healthy liver cells. The structural differences between the two classes (hepatocellular carcinoma (HCC) and normal liver tissue) are of clinical interest. Due to the large dimensionality (22,278), we first reduce the data  to the top 50 principal components with PCA. {The HCC cells are cancerous, arising from mutation. As such, they are expected to have higher within-class variance than the normally functioning liver cells. However, typical DR projections obscure this phenomenon; see Figure~\ref{fig:emotion-eval}.}

While most DR algorithms separate the two classes, the shapes of the clusters are different. t-SNE, for example, generates two clusters of equal size, and obtains good scores on SS, DBI, NMI, S\&C, and ARI. However, the HCC cluster is much more spread out in the ambient space (see supplemental material),
which is only revealed in the PCA and AngleEmbedding projections. \ourmetricAbbrev\ rewards these projections with the best scores. While Label-T\&C also ranks PCA well, it does not rank AngleEmbedding as highly.

\paragraph*{USPS.}

The USPS dataset is composed of $16 \times 16$ px images of handwritten digits from 0 to 9. We observe that local algorithms like t-SNE separate the 10 classes very well, and so obtain top scores in S\&C as well as the traditional cluster separation metrics (SS, DBI, NMI, ARI). PCA, which is unable to clearly separate the clusters, scores badly across most metrics.

However, while these algorithms generate clusters that are more or less equal in size, it is widely known that cluster sizes in neighbor embedding algorithms such as t-SNE are not interpretable. The AngleEmbedding projection has slightly more overlap between clusters, and cluster sizes vary in accordance with the data; further details are in the supplemental material. While Label-T\&C is not sensitive to cluster sizes, it does take into account meaningful overlap between clusters in the dataset, and therefore also ranks AngleEmbedding at the top alongside \ourmetricAbbrev.

It is worth noting that if the goal is strictly cluster identification, then t-SNE and UMAP remain effective tools. However, when the goal is to faithfully represent relationships between classes, \ourmetricAbbrev\ accurately quantifies this error.

{

\paragraph*{TREC.}
The TREC dataset is composed of sentence embeddings of questions classified into 6 coarse-grained classes and 50 fine-grained classes; we use the coarse-grained classes here. We observe that PCA forms clusters that overlap, though they are still nearly separable. t-SNE also seems to separate the classes somewhat, but we observe that it is more concerned with clustering smaller, nearest neighbor groups together. AngleEmbedding, on the other hand, clearly separates the 6 classes, in addition to splitting one of the classes into 2 clusters; further details are in the supplemental material. 

While most metrics rank AngleEmbedding at the top due to its class separation, S\&C ranks AngleEmbedding last, instead rewarding t-SNE the most. Due to the local nature of t-SNE, it clusters tiny groups of points of the same class together, but fails to capture the larger-scale structure. S\&C, which relies on neighbor clustering, identifies and rewards these clusters formed by t-SNE. However, it fails to recognize the class-level information revealed by AngleEmbedding and ranks it last as a result.
}

\paragraph*{Other data.}
Across all datasets, only \ourmetricAbbrev\ consistently assigns the worst score to the Random projection. All other metrics at least once assign higher scores to the Random projection when other, clearly more informative projections are given. We attribute this to the globularity assumption made by other metrics; their blindness to cluster structure makes them susceptible to errors that the \ourmetricFull\ successfully avoids.

Label-T\&C agrees with \ourmetricAbbrev\ on several datasets: USPS, emotion, and Fashion-MNIST. In each of these, AngleEmbedding does not show clearly separated clusters. As Label-T\&C measures the degree of separation of class labels in the original data, this would indicate that a cleanly separated plot, as produced by t-SNE or UMAP, does not preserve cluster-level relationships accurately. 
}

\subsection{Case study}

\begin{figure}[h]
    \centering
    \includegraphics[width=\linewidth]{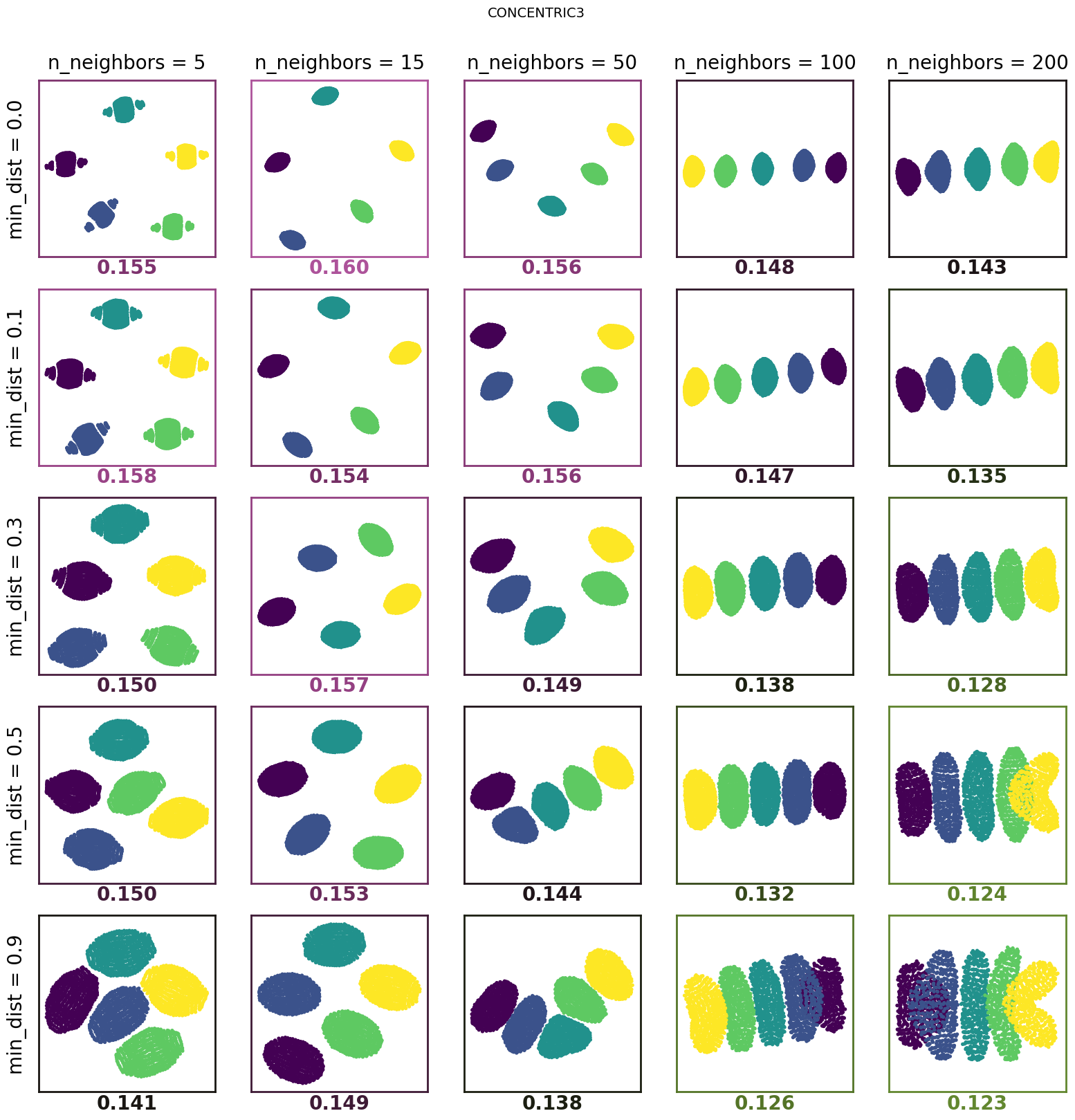}
    \caption{\ourmetricAbbrev\ scores for UMAP projections of the concentric3 dataset with different values for the hyperparameters \texttt{min\_dist} and \texttt{n\_neighbors}. Green borders indicate better scores, while red borders indicate worse scores. 
    }
    \vspace{-0.4cm}
    \label{fig:umap-hparam}
\end{figure}

We illustrate how \ourmetricFull\ can support hyperparameter tuning using UMAP on the concentric3 dataset; see \autoref{fig:umap-hparam}. We vary UMAP’s two primary hyperparameters, \texttt{n\_neighbors} and \texttt{min\_dist}, which respectively control the size of local neighborhoods and the compactness of clusters.

As expected, none of the UMAP configurations recovers the true nested structure of the spheres, since UMAP focuses on local relationships rather than global topology. However, \ourmetricAbbrev\ reveals clear trends in how hyperparameters affect cluster organization. Increasing \texttt{n\_neighbors} yields more coherent global structure: clusters become ordered along a line, with points from the innermost sphere appearing at one end and progressively outer spheres arranged sequentially. This organization leads to better \ourmetricAbbrev\ scores.

The worst scores occur in the upper-left region of the parameter grid, where clusters fragment into several pieces. Interestingly, the best scores arise in the lower-right region, where some adjacent clusters partially merge. Although this merging may appear undesirable at first glance, it reflects a more faithful representation of the high-dimensional relationships between neighboring spheres, which are closer to each other than to non-adjacent ones.

Beyond a certain threshold, larger values of the hyperparameters yield little qualitative change, so we omit them for clarity. Overall, this case study demonstrates that \ourmetricAbbrev\ can guide hyperparameter selection by identifying projections that best preserve cluster-level structure, even when the true relationships are not visually obvious.

\section{Discussion, Limitations and Conclusion}
We conclude with a synthesis of our results and limitations. 

\paragraph*{Discussion.}

Our evaluation leads to several key observations. First, measuring internal angles of points between clusters is an effective way to measure inter-cluster structure. Across all synthetic datasets, \ourmetricAbbrev\ consistently identifies projections that {preserve inter-cluster relationships}, even when those relationships depend on topology (Rings), hierarchy (Concentric), or non-globular shapes. This is indicated by how AngleEmbedding successfully recovers the embedded structure in our synthetic datasets. 

Second,  existing cluster-level measures assume that clusters are compact, well-separated, and approximately globular. This assumption produces systematic failure modes; all existing metrics assign higher scores to a random projection than to a more informative one in at least one example. Many metrics evaluate separability rather than structure, and therefore cannot reliably assess fidelity in non-globular or nested settings.

Third,  real-world data rarely exhibit clearly separated clusters. Instead, classes often overlap and intersect in complex ways that reflect real properties of the underlying data. 
In such cases, a faithful projection may mix classes more than aesthetically pleasing methods like t-SNE or UMAP. \ourmetricAbbrev\ captures this nuance by rewarding projections that preserve inter-class relationships rather than enforcing artificial separation.

Finally, our results help clarify when different tools should be used. If the analytical goal is simply to estimate the number of clusters or to visually separate groups for labeling or exploration, t-SNE and UMAP remain effective and convenient methods. However, when the goal is to understand how clusters relate to one another, our results show that \ourmetricAbbrev\ is a more reliable quantitative guide and AngleEmbedding produces projections that more faithfully reflect the underlying cluster-level geometry.

\paragraph*{Limitations.}
While \ourmetricFull\ captures the preservation of cluster-level topology better than the other metrics, it does not always capture cluster separability. We do not aim to replace cluster-level metrics like Silhouette Score, as they may be a better measure in instances where the dataset is indeed composed of tightly-packed, well-separated clusters with little or uninteresting topological structure. When relationships between clusters are of interest, \ourmetricAbbrev\ provides a measure of structural preservation not captured by other measures.

Similar to the other cluster-level metrics, \ourmetricAbbrev\ requires a clustering assignment as a hyperparameter. While we have used class labels to measure this in our experiments, as done in previous literature~\cite{Cardarelli2022clusterEval, Becker2019ReNDA}, a ground truth class labeling may not always be available. In this case, class labels could be assigned through clustering algorithms. As scores may vary with different clustering assignments, metric results should be evaluated
with respect to the clustering assignment chosen. Further, our evaluation has only a finite number of datasets, algorithms, and metrics. This represents a potential threat to the external validity of our experiments, and future work is needed to understand \ourmetricAbbrev 's robustness to labels.

We acknowledge that the aggregation step in \ourmetricAbbrev 's definition warrants additional clarification. While individual angular comparisons capture geometric relationships independently of the globularity assumption, summation across triplets introduces additional effects, possibly based on class size and internal sampling. Larger classes may exert greater influence. The precise qualities of the summed measure are not yet characterized fully theoretically, and a deeper analysis is an important direction for future work.

We further note that since the AngleEmbedding algorithm optimizes \ourmetricAbbrev, it is supervised and therefore it would be unfair to compare its projections with those of unsupervised algorithms like t-SNE and UMAP. We only present AngleEmbedding as a demonstration of projections that \ourmetricAbbrev\ promotes. 

While we note the lack of cluster-level structure preservation from t-SNE and UMAP, we acknowledge that this is not the intended result of these algorithms. Rather, they try to highlight local patterns and separate clusters when they exist. When analytic tasks only require identifying clusters rather than understanding their relationships, these algorithms remain effective tools.

Finally, while Label T\&C and \ourmetricAbbrev\ share similar motivations, our evaluation shows that they differ significantly in what they measure. A more thorough comparison between these measures is left for future work.

Potentially interesting avenues of future work include extending AngleEmbedding to non-Euclidean spaces such as the sphere or hyperboloid~\cite{miller2024state}, and optimizing several perspectives of AngleEmbedding simultaneously~\cite{miller2024state,hossain2020multi}. 

\paragraph*{Conclusion.}
We introduced the \ourmetricFull\ to measure cluster-level structure preservation through internal angles. The metric is conceptually simple, but effective in capturing cluster-level distortions in projections. We evaluated \ourmetricAbbrev\ on a collection of synthetic and real-world datasets, demonstrating that it avoids the pitfalls commonly present in other metrics. Finally, AngleEmbedding is a new DR algorithm which optimizes \ourmetricAbbrev, and provides qualitatively distinct and informative results. Together, these contributions offer a new perspective on evaluating DR projections at the cluster level.

\bibliographystyle{eg-alpha-doi} 
\bibliography{references}

\appendix

\section{The Importance of Cluster-Constrained Triplet Selection}
\label{sec:adi_vs_cadi}

A natural question is whether the class-constrained triplet selection used in 
\ourmetricAbbrev\ is actually necessary. One may instead consider a simpler strategy:
sample arbitrary triplets $(j,i,k)$ uniformly at random from the dataset,
without taking the cluster partition $C$ into account.
In this section, we investigate this alternative and show that it fundamentally
changes the nature of the metric.

We define the \emph{Angular Distortion Index (ADI)} as

\begin{equation*}
\mathrm{ADI}(X,Y)
=
\frac{1}{T_{\mathrm{all}}}
\sum_{\substack{i,j,k \\ \text{distinct}}}
\left(
\cos\bigl(\theta_X(j,i,k)\bigr)
-
\cos\bigl(\theta_Y(j,i,k)\bigr)
\right)^2,
\end{equation*}

where the sum ranges over all unordered triplets of distinct indices
(and $T_{\mathrm{all}}$ denotes their total number).
We sample $100n$ triplets in practice, where $n$ is the total number of points in the dataset.
Unlike \ourmetricAbbrev, ADI does \emph{not} use the partition $C$.

When measuring ADI on our datasets and projections, we observe that the results vary drastically from \ourmetricAbbrev. We mark the top 2 ranked algorithms for each dataset according to ADI in \autoref{tab:adi-ranks}. We notice that the algorithms that focus on preserving global structure, MDS and PCA, are always ranked at the top. MDS is ranked 1st in 13 out of 19 datasets, and is ranked 2nd in 5 out of the rest of the 6 datasets. More notably, either MDS, PCA, or both appear in the top 2 ranks on all datasets. This behavior stands in clear contrast to the fact that MDS and PCA clearly fail to reveal cluster-level structure on some datasets, as presented in earlier sections.

\begin{table}[h!]
    \centering
    \begin{tabular}{lcc}
    \toprule
     & \textbf{Rank 1} & \textbf{Rank 2} \\
    \midrule
    MNIST & MDS & TSNE \\
    acl\_imdb & MDS & AngleEmbedding \\
    coil100 & MDS & PCA \\
    coil20 & MDS & PCA \\
    concentric3 & AngleEmbedding & MDS \\
    concentric4 & AngleEmbedding & PCA \\
    donuts & MDS & PCA \\
    emotion & MDS & UMATO \\
    fashionMNIST & MDS & PCA \\
    liver & MDS & AngleEmbedding \\
    matryoshka & AngleEmbedding & MDS \\
    olivetti & MDS & PCA \\
    pbmc3k & MDS & Random \\
    pendigits & MDS & PCA \\
    penguins & PCA & MDS \\
    rings & AngleEmbedding & MDS \\
    sentiment & AngleEmbedding & MDS \\
    trec & MDS & TSNE \\
    usps & MDS & PCA \\
    \bottomrule
    \end{tabular}
    \caption{Top two dimensionality reduction algorithms per dataset ranked by Angular Distortion Index (ADI). Globally oriented methods (MDS and PCA) consistently occupy the highest ranks across datasets.}
    \label{tab:adi-ranks}
\end{table}

These findings confirm that unconstrained triplet sampling induces a
global angular distortion measure. Because it averages uniformly over all differences in internal angles, information about cluster-level geometries is drowned out by noise. This is evidence that the constrained triplet selection is a critical component in \ourmetricAbbrev\ to measure inter-cluster relationships.

\section{Time taken}
\label{sec:time-taken}

The times taken to compute each metric on the t-SNE projection for each dataset are displayed in \autoref{fig:times-taken}.

{
\noindent
\centering
\begin{minipage}{\linewidth}
    \centering
    \includegraphics[width=\linewidth]{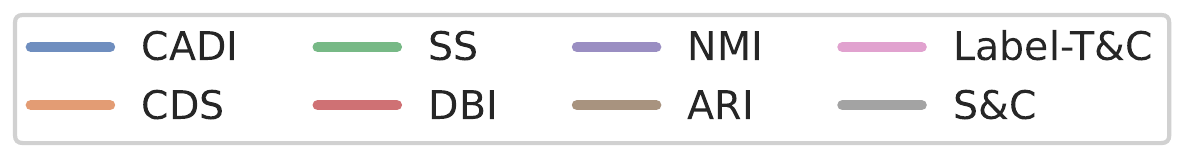}
    \\[2mm]
    (a) Legend
\end{minipage}
\vspace{6pt}
\begin{minipage}{\linewidth}
    \centering
    \includegraphics[width=\linewidth]{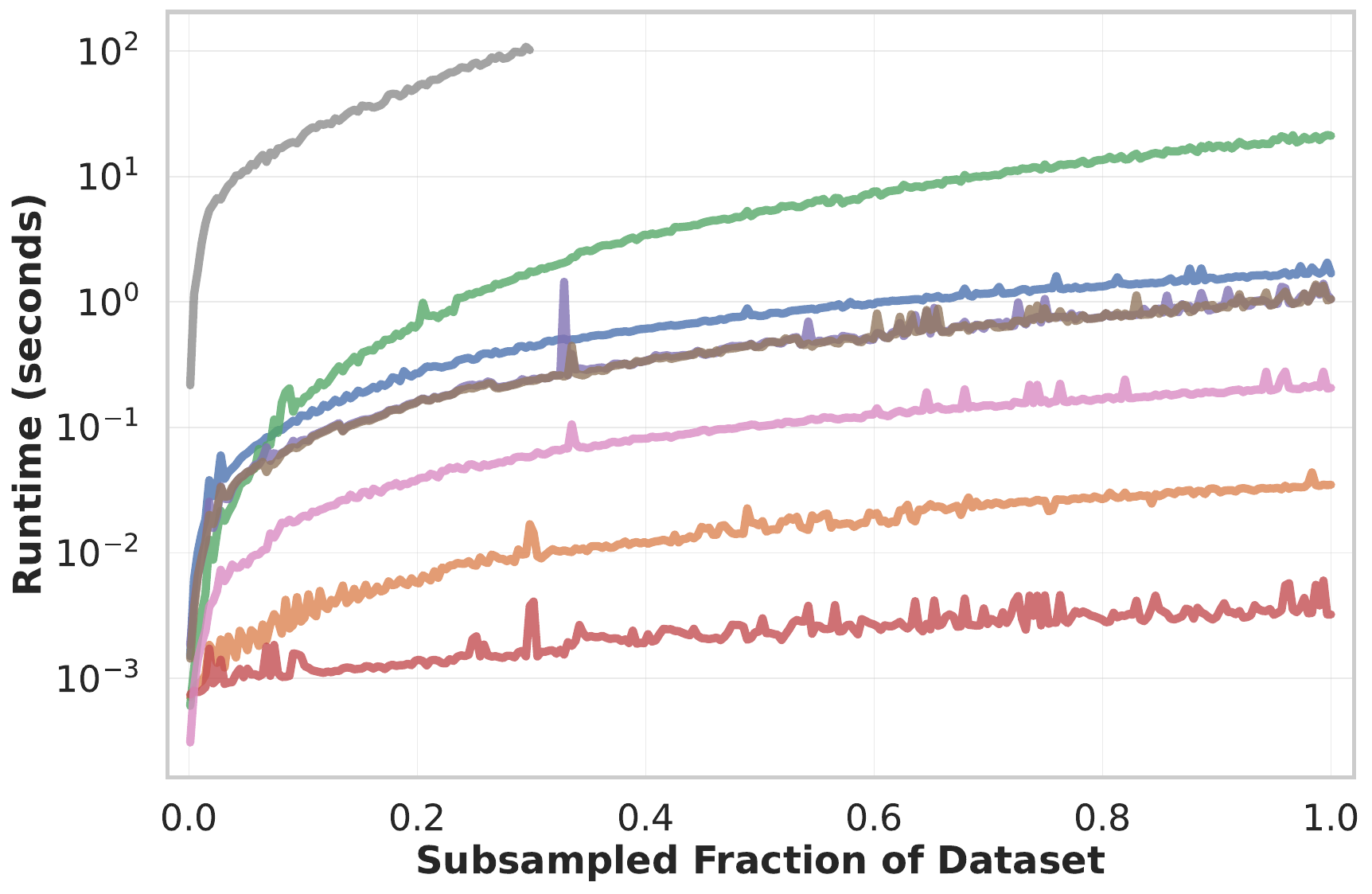}
    \\[2mm]
    (b) acl\_imdb
\end{minipage}
\vspace{6pt}
\begin{minipage}{\linewidth}
    \centering
    \includegraphics[width=\linewidth]{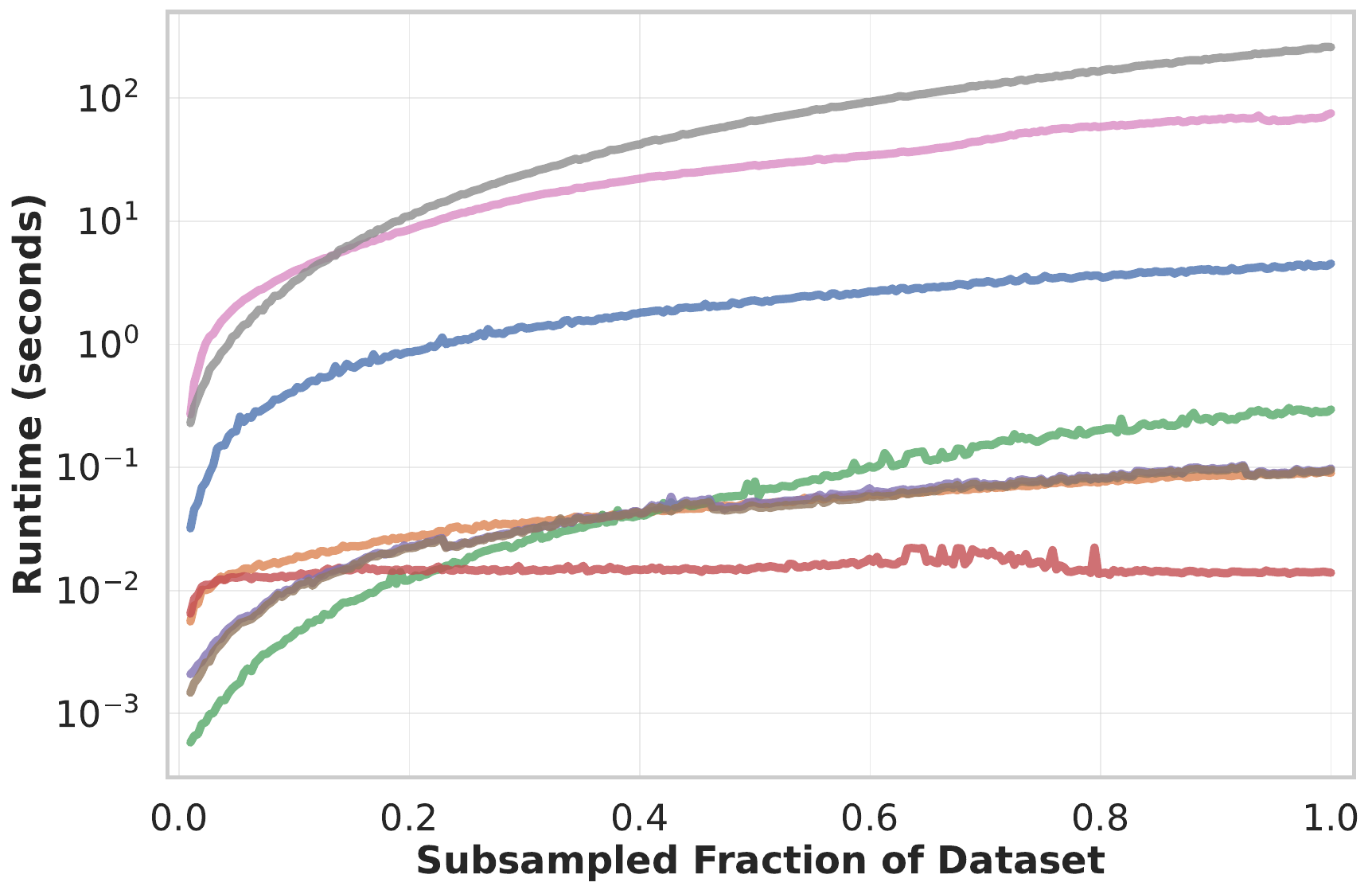}
    \\[2mm]
    (c) coil100
\end{minipage}
\vspace{6pt}
\begin{minipage}{\linewidth}
    \centering
    \includegraphics[width=\linewidth]{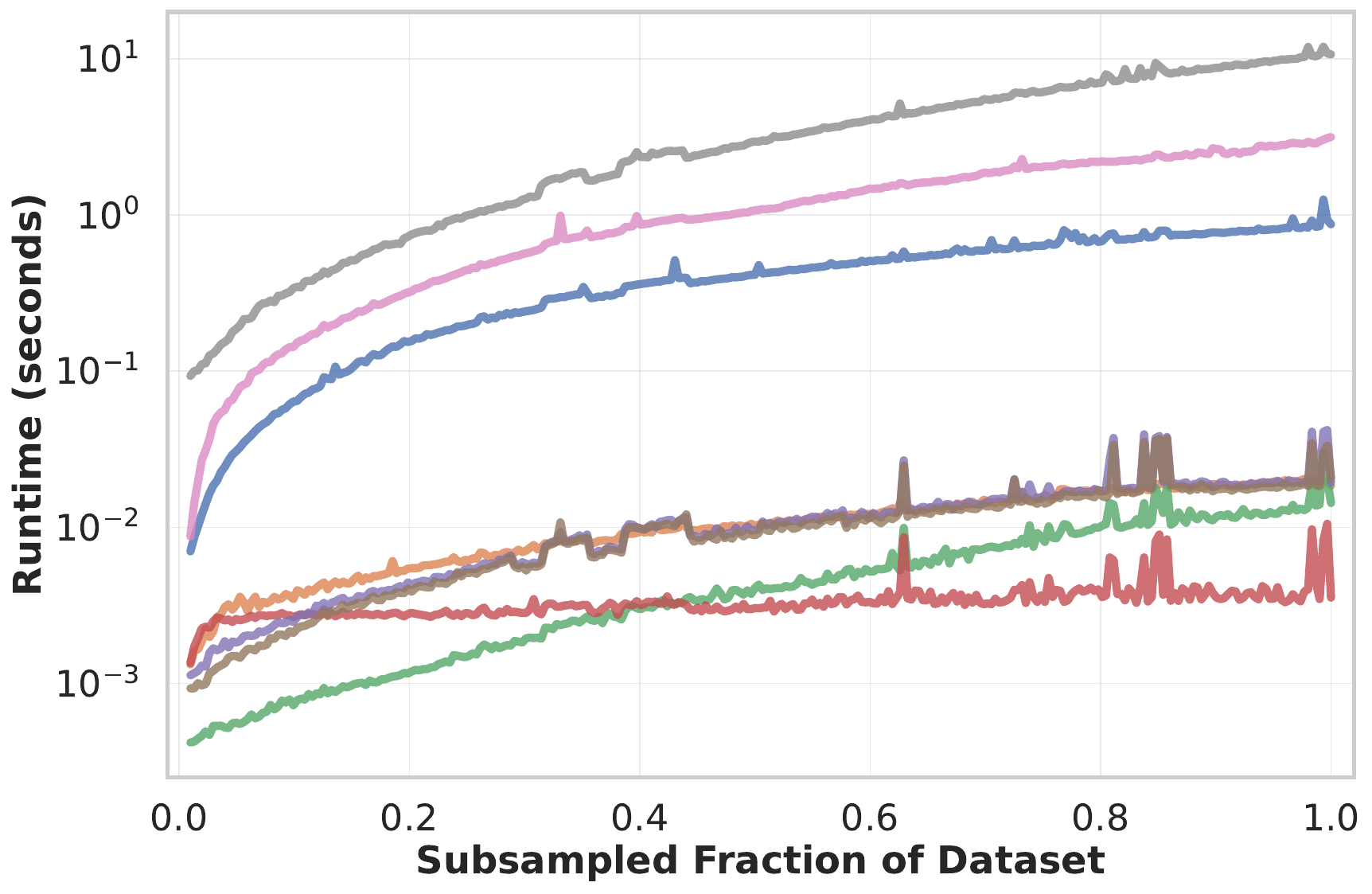}
    \\[2mm]
    (d) coil20
\end{minipage}
\vspace{6pt}
\begin{minipage}{\linewidth}
    \centering
    \includegraphics[width=\linewidth]{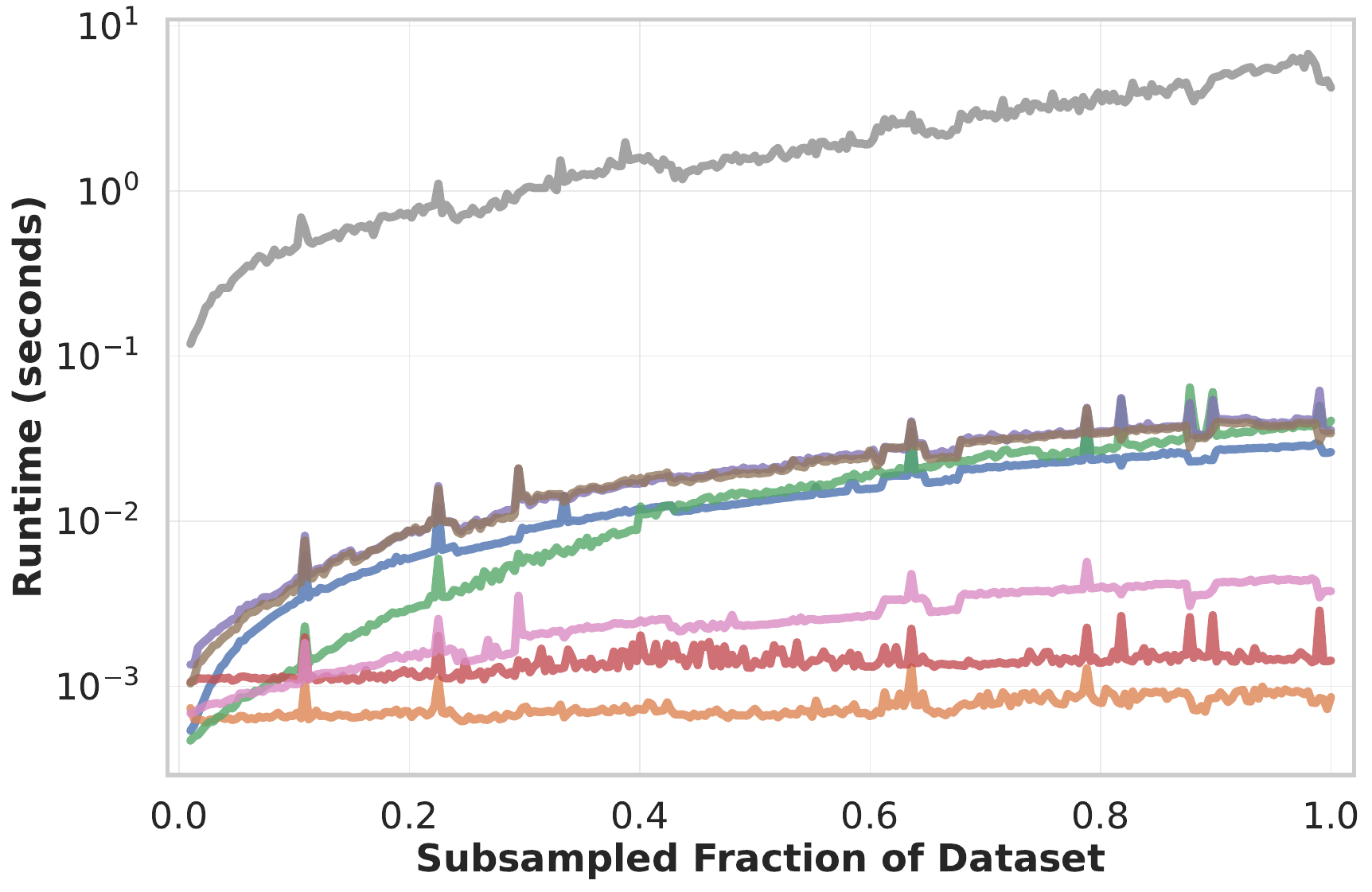}
    \\[2mm]
    (e) concentric3
\end{minipage}
\vspace{6pt}
\begin{minipage}{\linewidth}
    \centering
    \includegraphics[width=\linewidth]{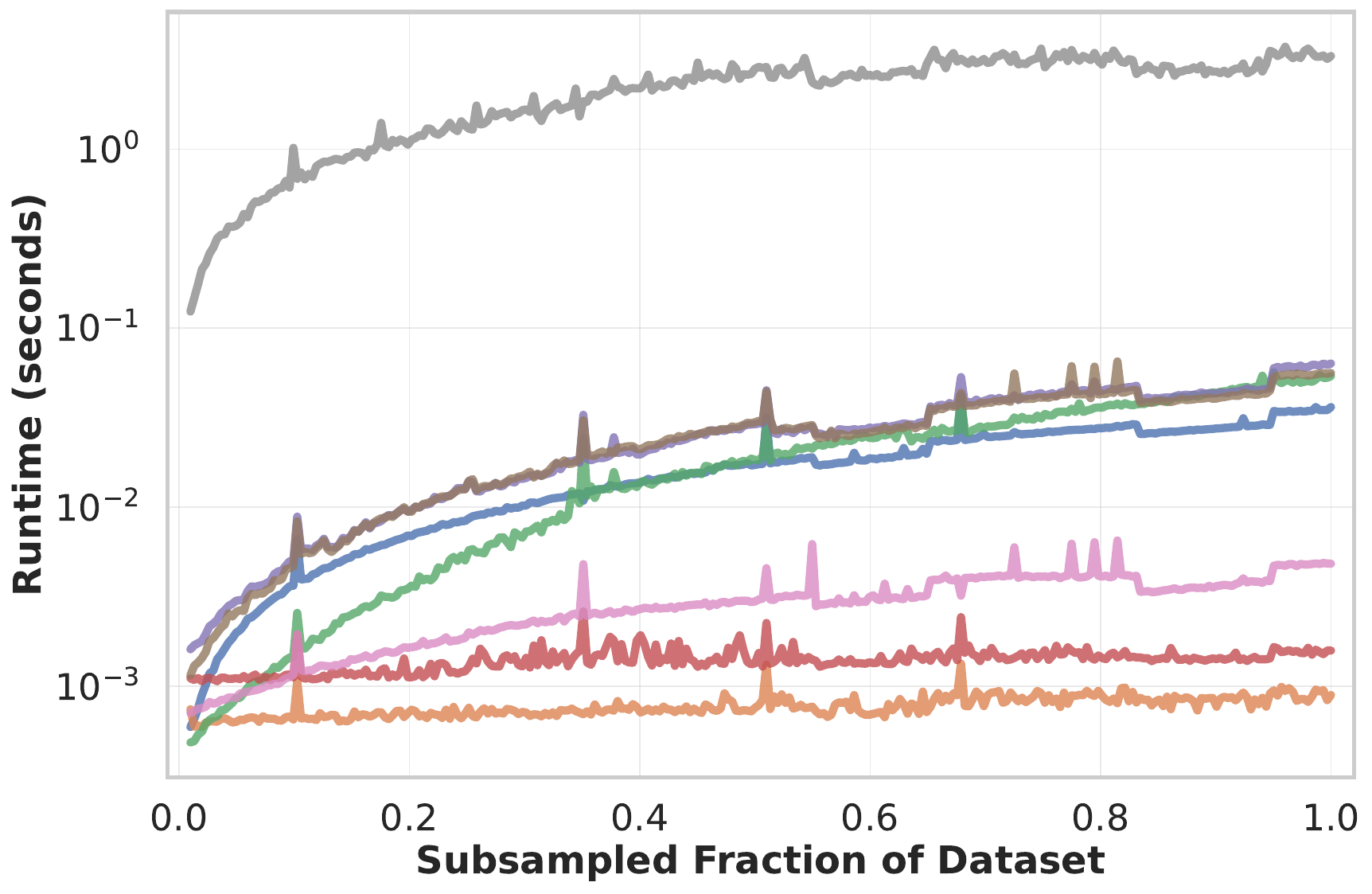}
    \\[2mm]
    (f) concentric4
\end{minipage}
\vspace{6pt}
\begin{minipage}{\linewidth}
    \centering
    \includegraphics[width=\linewidth]{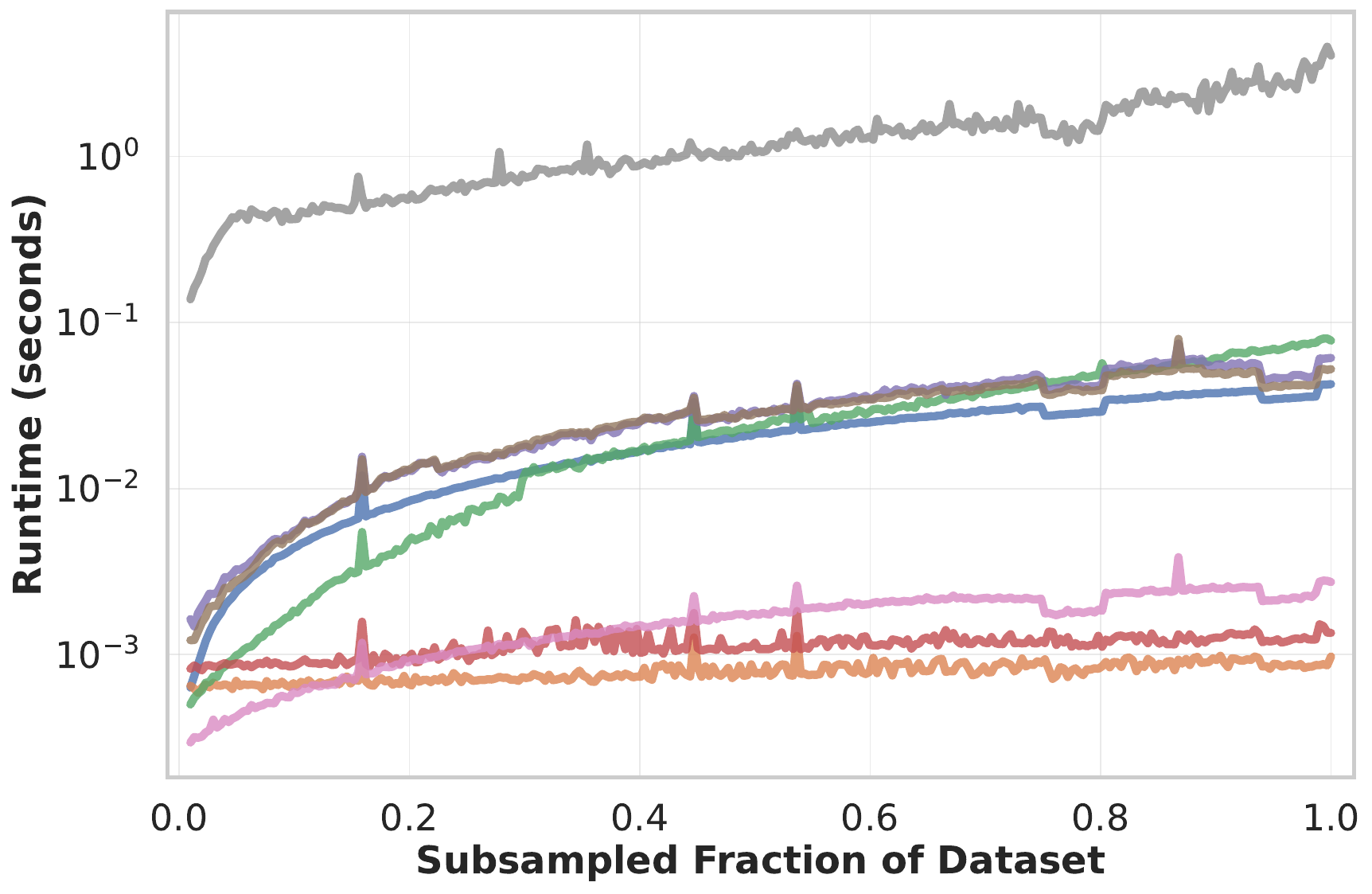}
    \\[2mm]
    (g) donuts
\end{minipage}
\vspace{6pt}
\begin{minipage}{\linewidth}
    \centering
    \includegraphics[width=\linewidth]{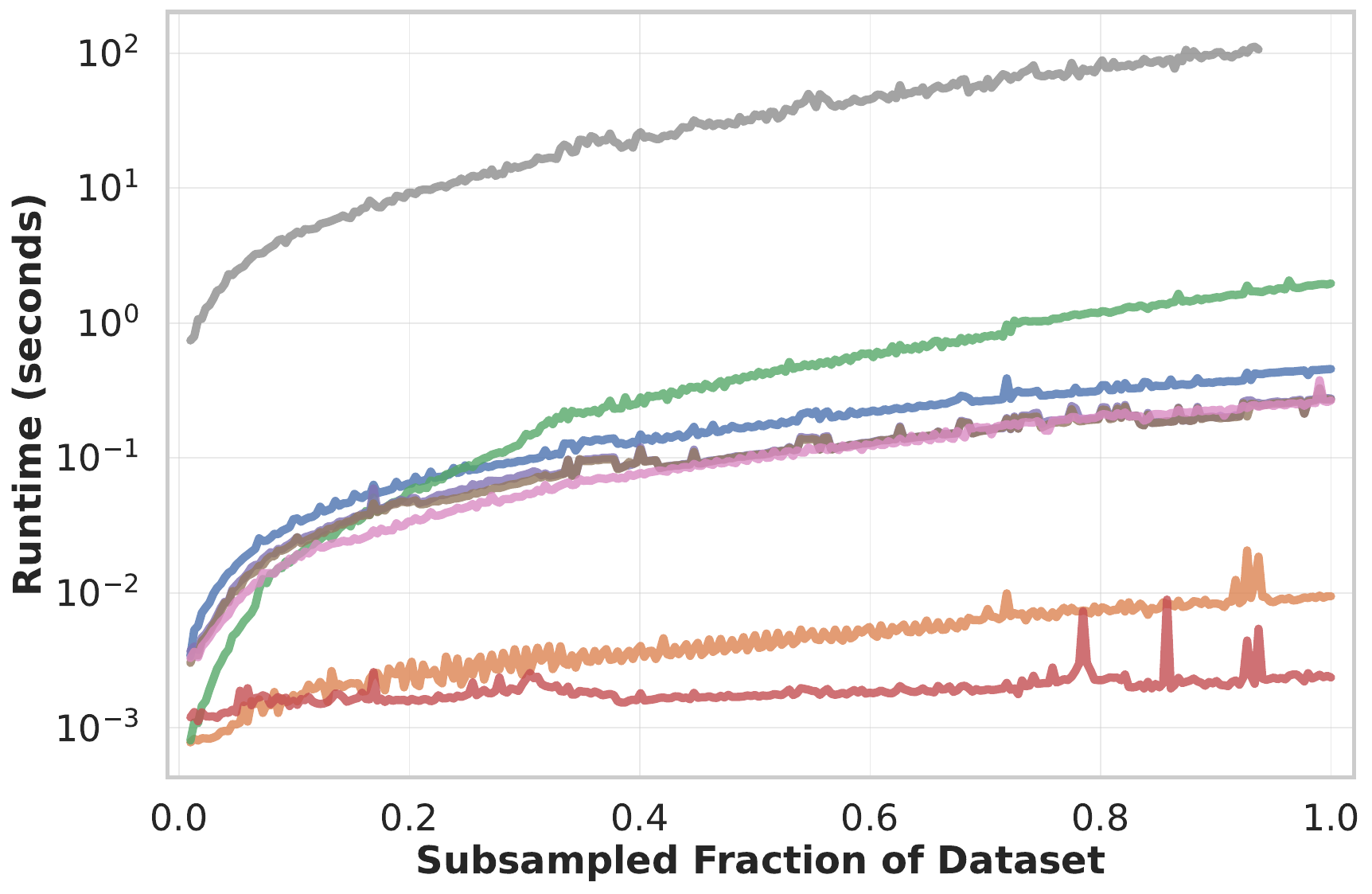}
    \\[2mm]
    (h) emotion
\end{minipage}
\vspace{6pt}
\begin{minipage}{\linewidth}
    \centering
    \includegraphics[width=\linewidth]{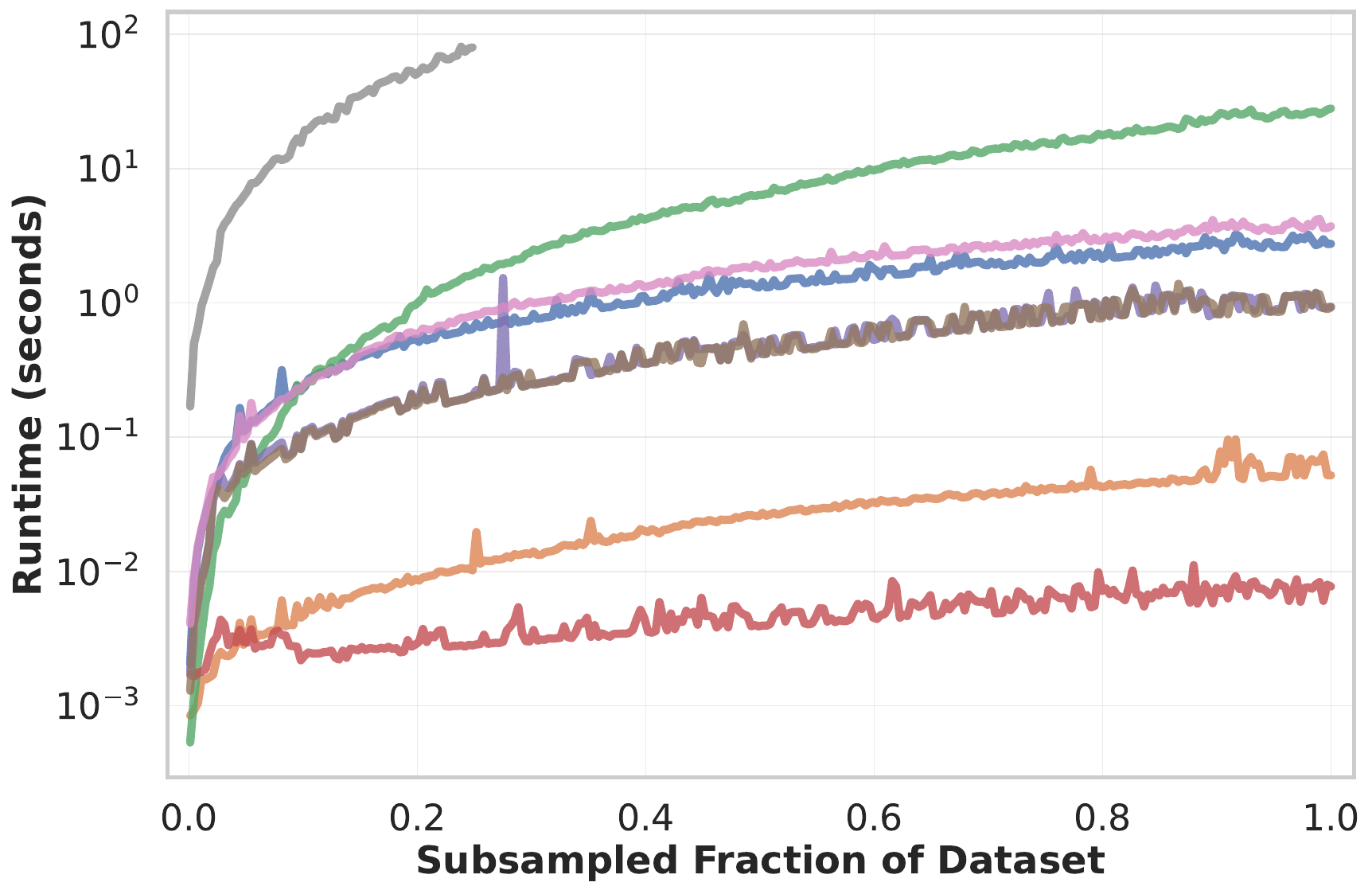}
    \\[2mm]
    (i) F-MNIST
\end{minipage}
\vspace{6pt}
\begin{minipage}{\linewidth}
    \centering
    \includegraphics[width=\linewidth]{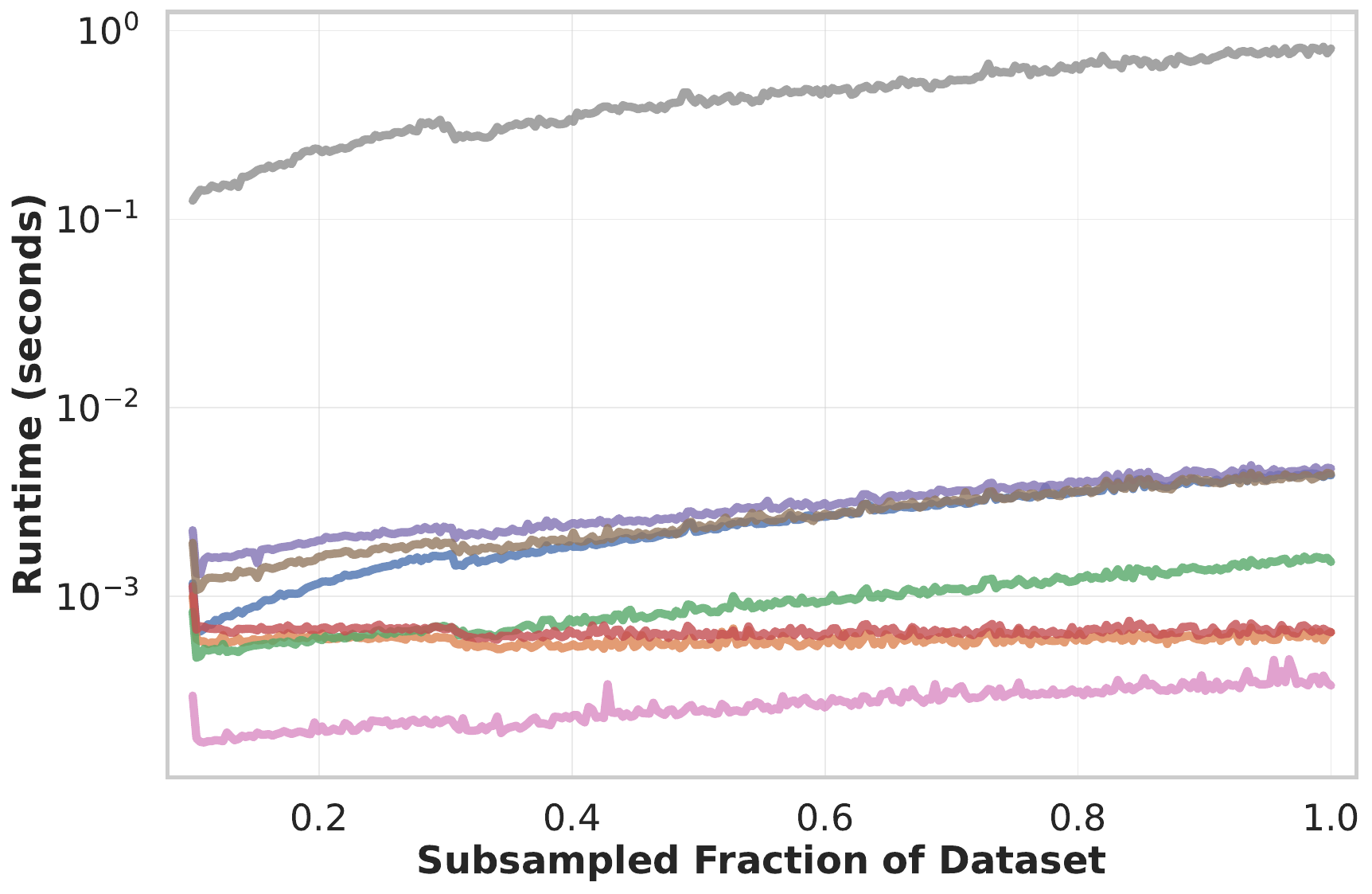}
    \\[2mm]
    (j) liver
\end{minipage}
\vspace{6pt}
\begin{minipage}{\linewidth}
    \centering
    \includegraphics[width=\linewidth]{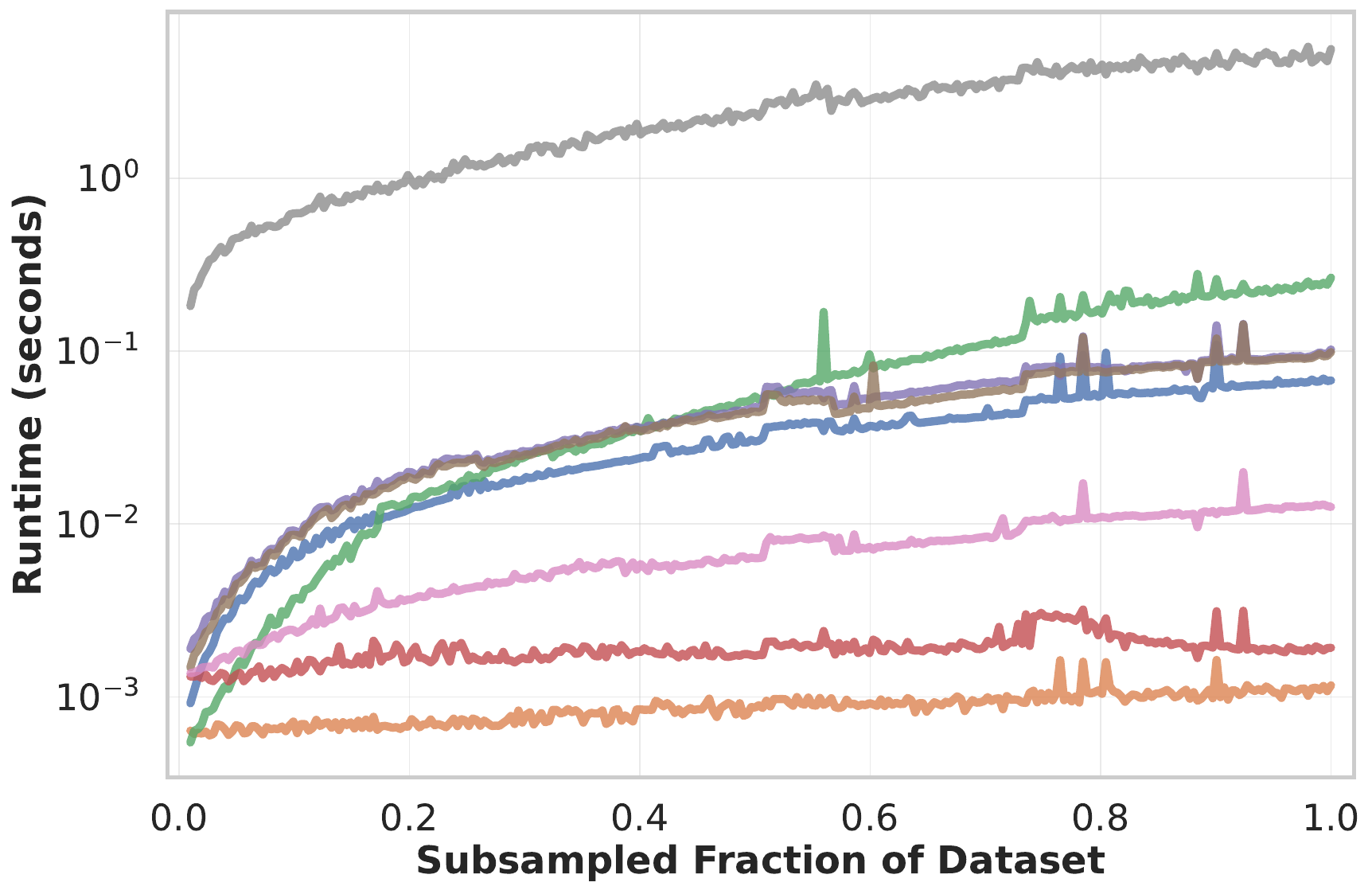}
    \\[2mm]
    (k) matryoshka
\end{minipage}
\vspace{6pt}
\begin{minipage}{\linewidth}
    \centering
    \includegraphics[width=\linewidth]{images/time_taken/MNIST.pdf}
    \\[2mm]
    (l) MNIST
\end{minipage}
\vspace{6pt}
\begin{minipage}{\linewidth}
    \centering
    \includegraphics[width=\linewidth]{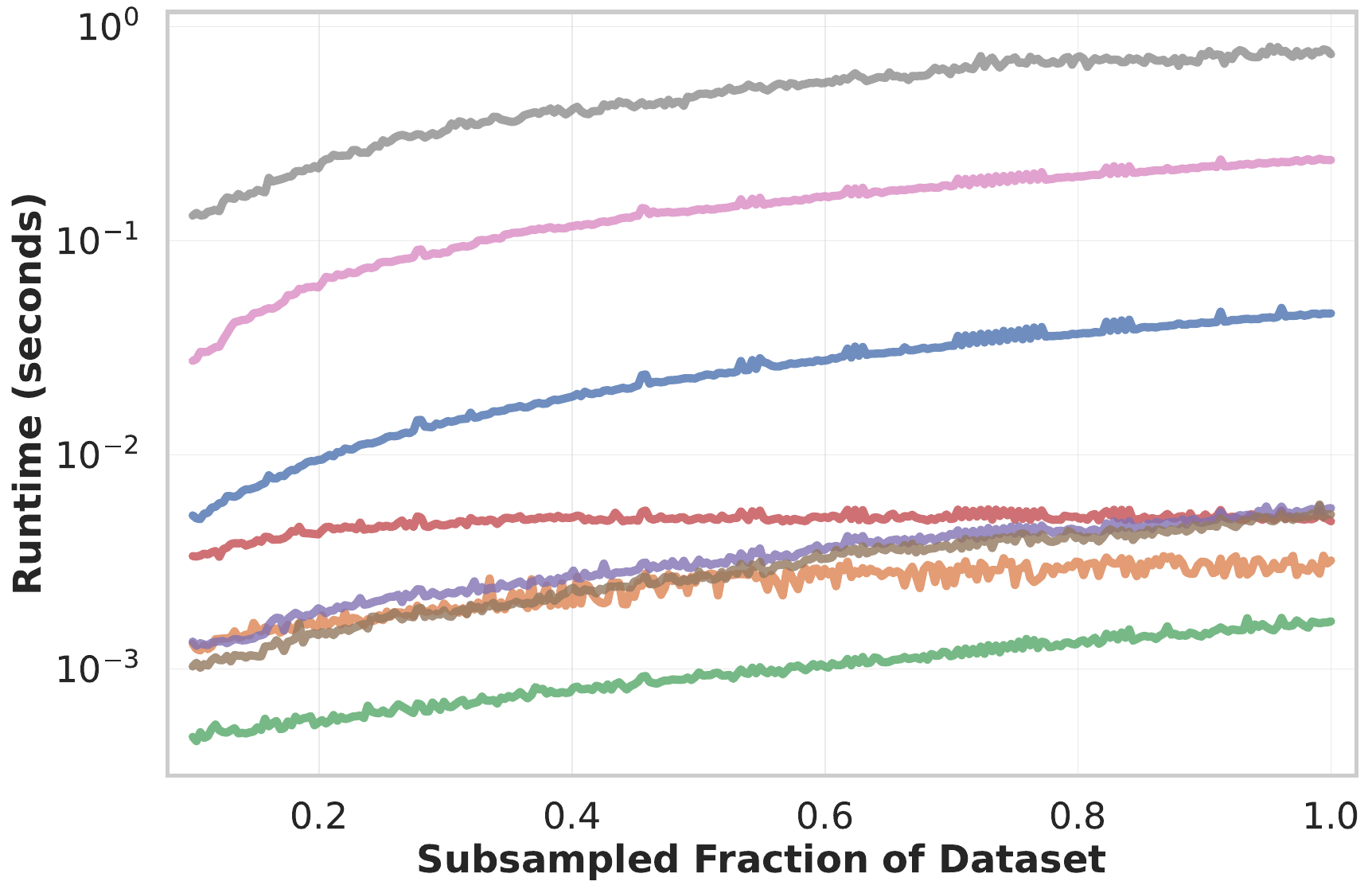}
    \\[2mm]
    (m) olivetti
\end{minipage}
\vspace{6pt}
\begin{minipage}{\linewidth}
    \centering
    \includegraphics[width=\linewidth]{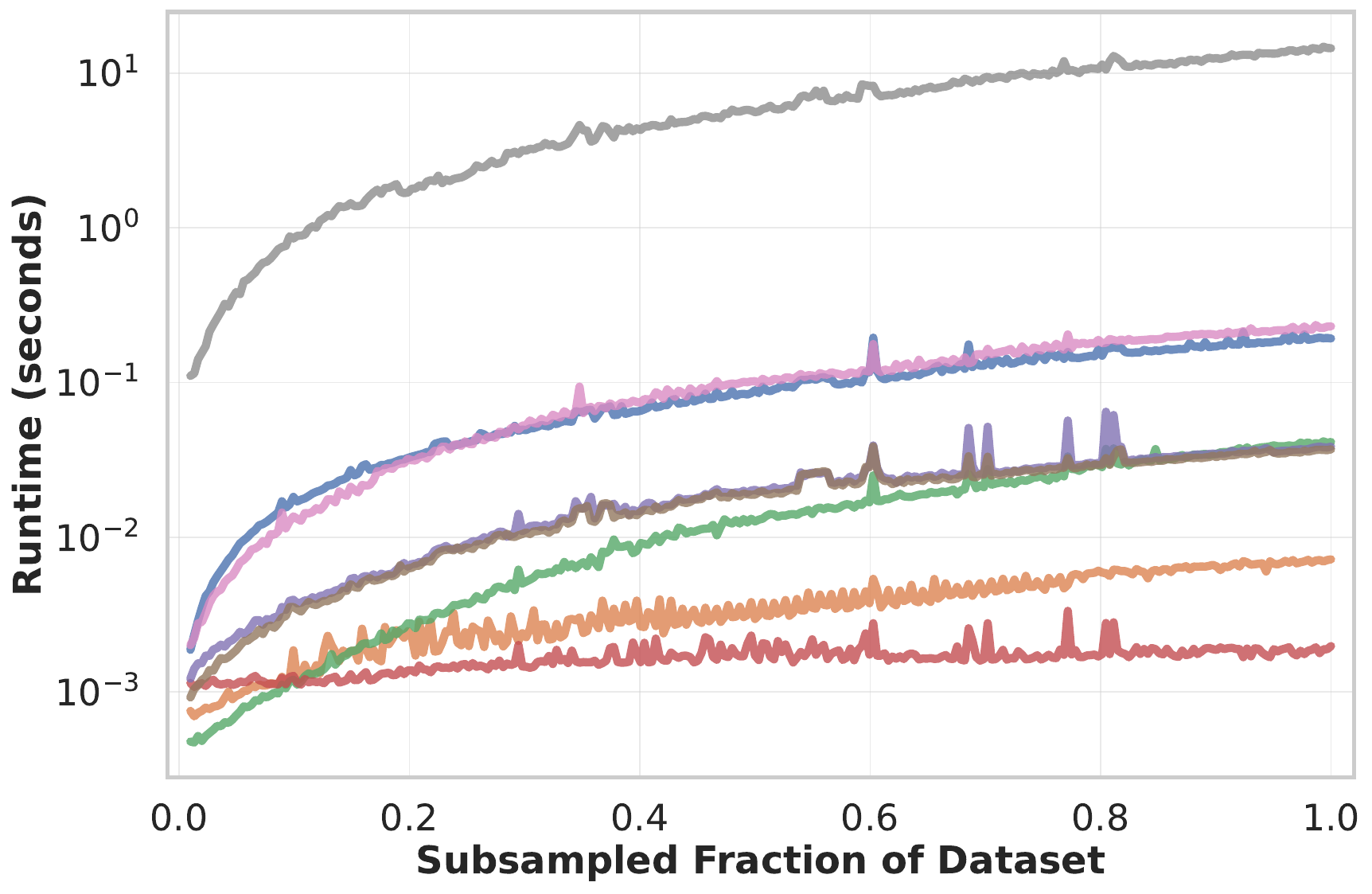}
    \\[2mm]
    (n) pbmc3k
\end{minipage}
\vspace{6pt}
\begin{minipage}{\linewidth}
    \centering
    \includegraphics[width=\linewidth]{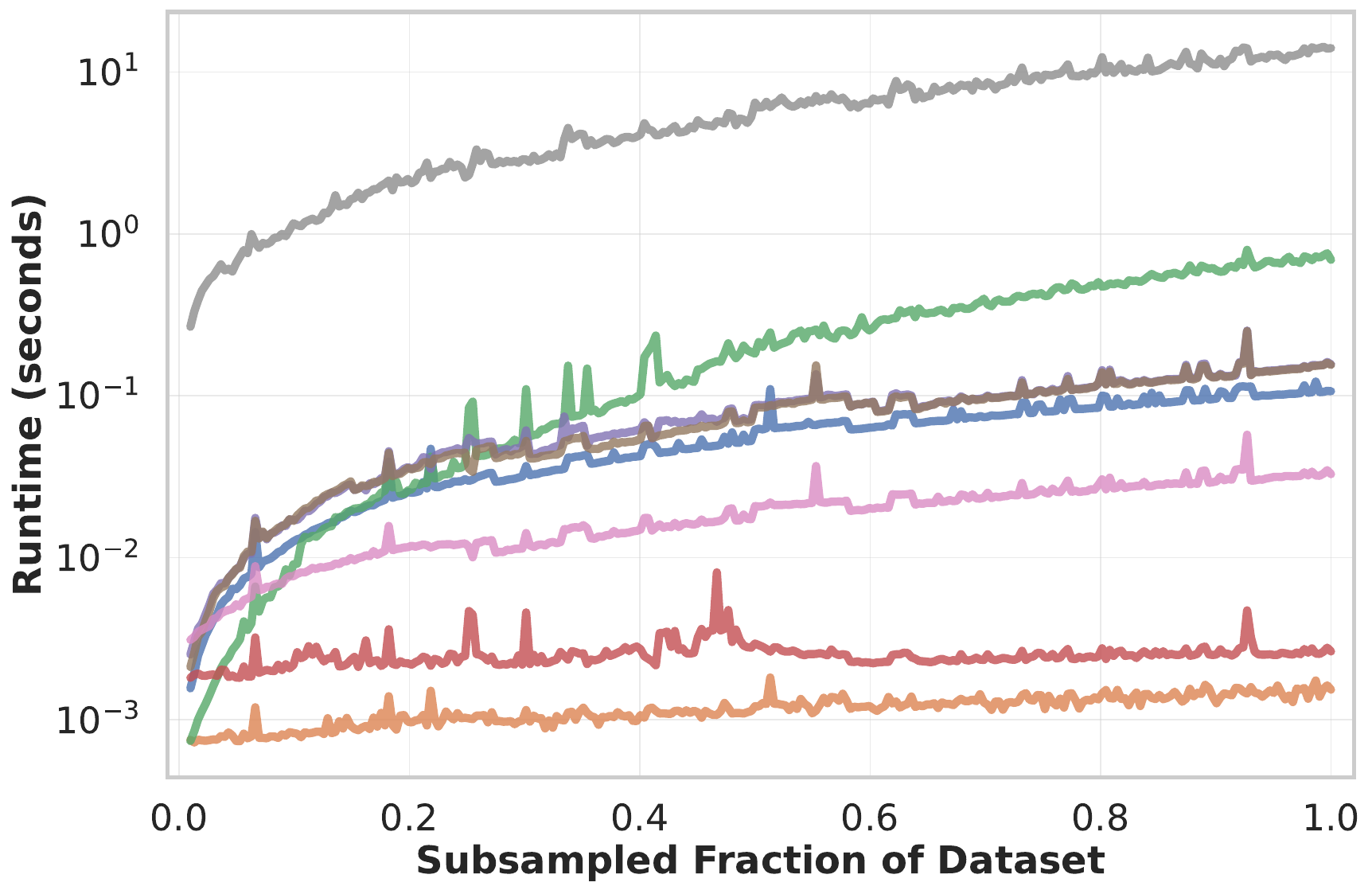}
    \\[2mm]
    (o) pendigits
\end{minipage}
\vspace{6pt}
\begin{minipage}{\linewidth}
    \centering
    \includegraphics[width=\linewidth]{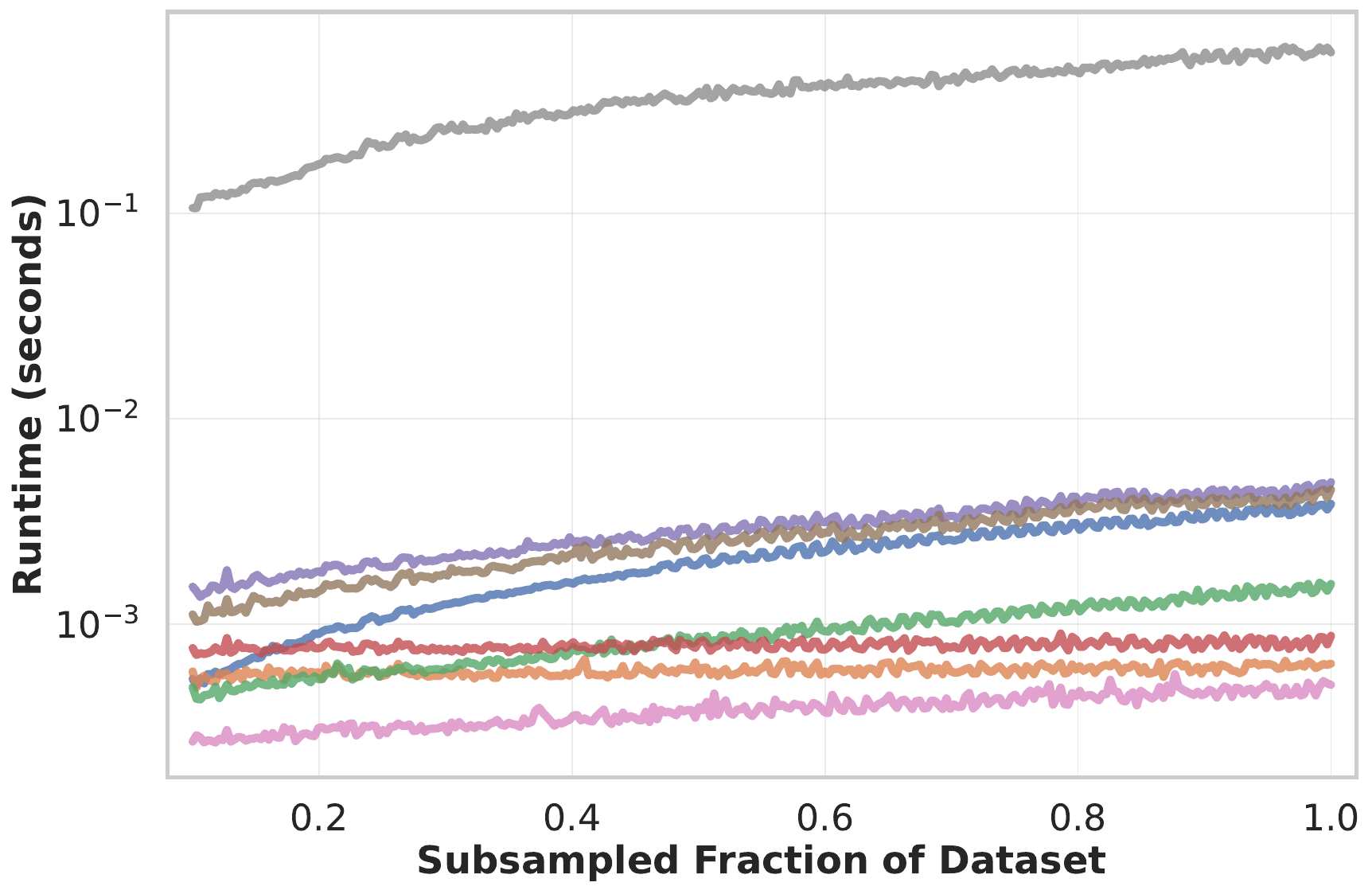}
    \\[2mm]
    (p) penguins
\end{minipage}
\vspace{6pt}
\begin{minipage}{\linewidth}
    \centering
    \includegraphics[width=\linewidth]{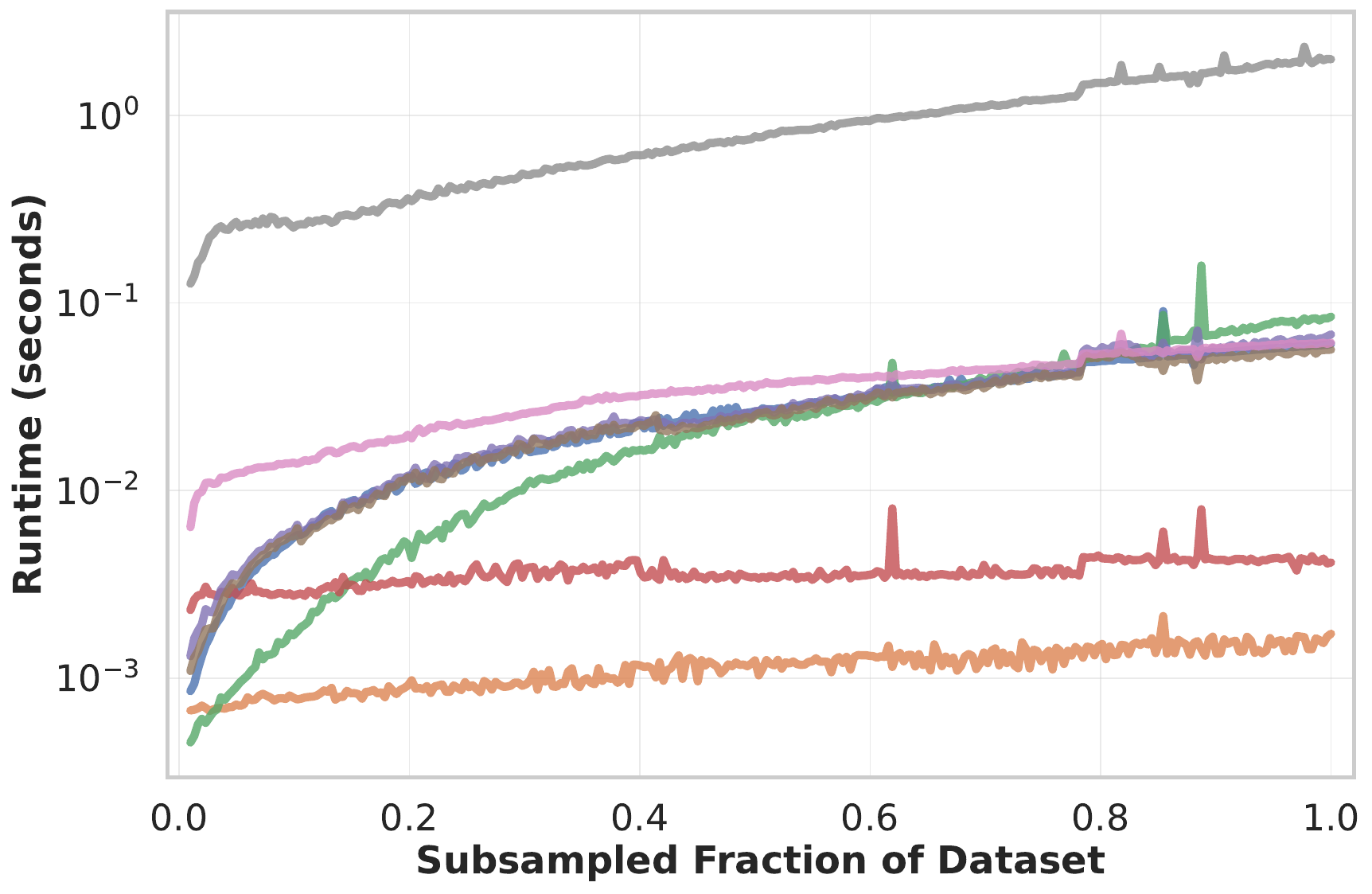}
    \\[2mm]
    (q) rings
\end{minipage}
\vspace{6pt}
\begin{minipage}{\linewidth}
    \centering
    \includegraphics[width=\linewidth]{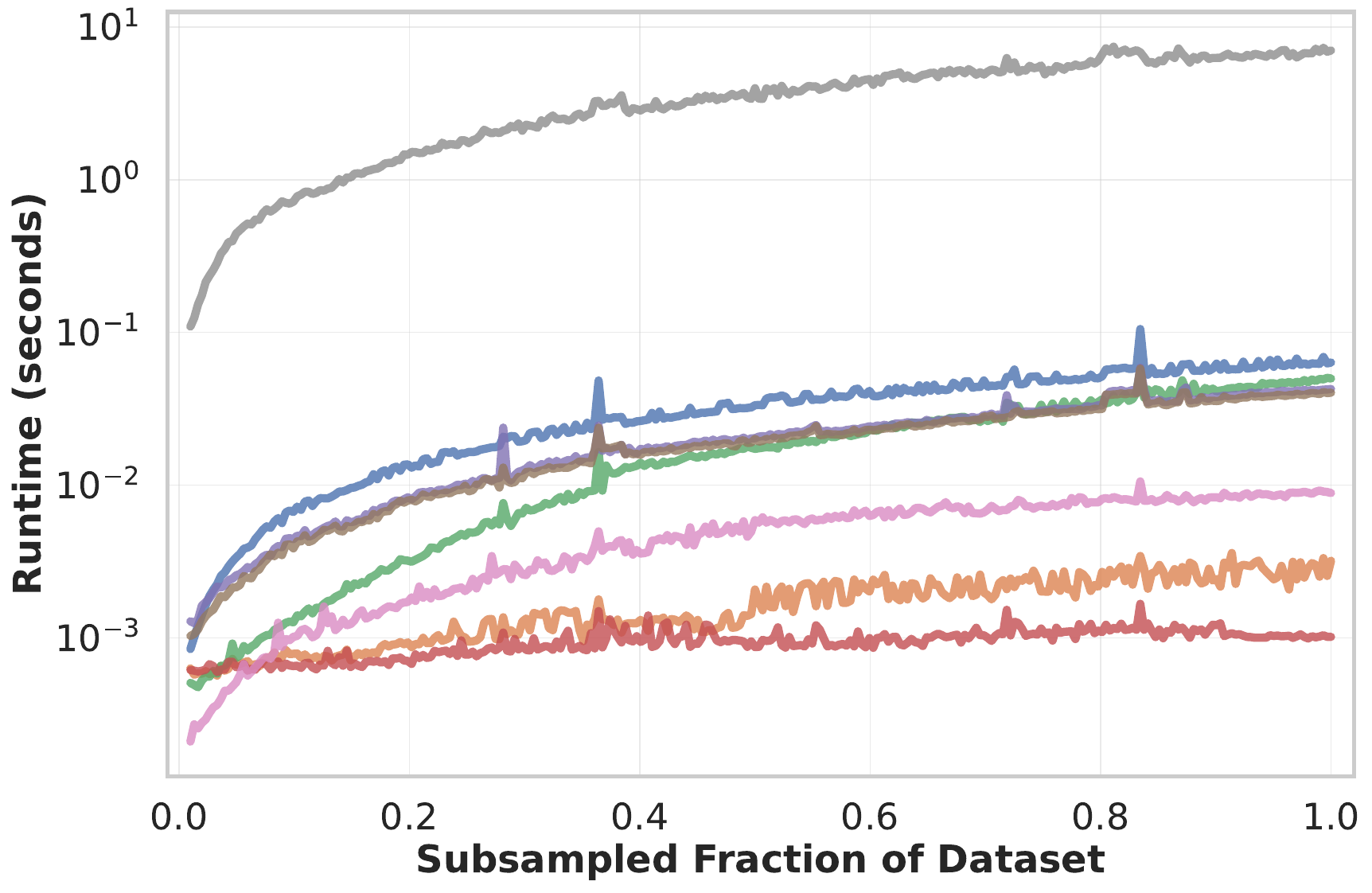}
    \\[2mm]
    (r) sentiment
\end{minipage}
\vspace{6pt}
\begin{minipage}{\linewidth}
    \centering
    \includegraphics[width=\linewidth]{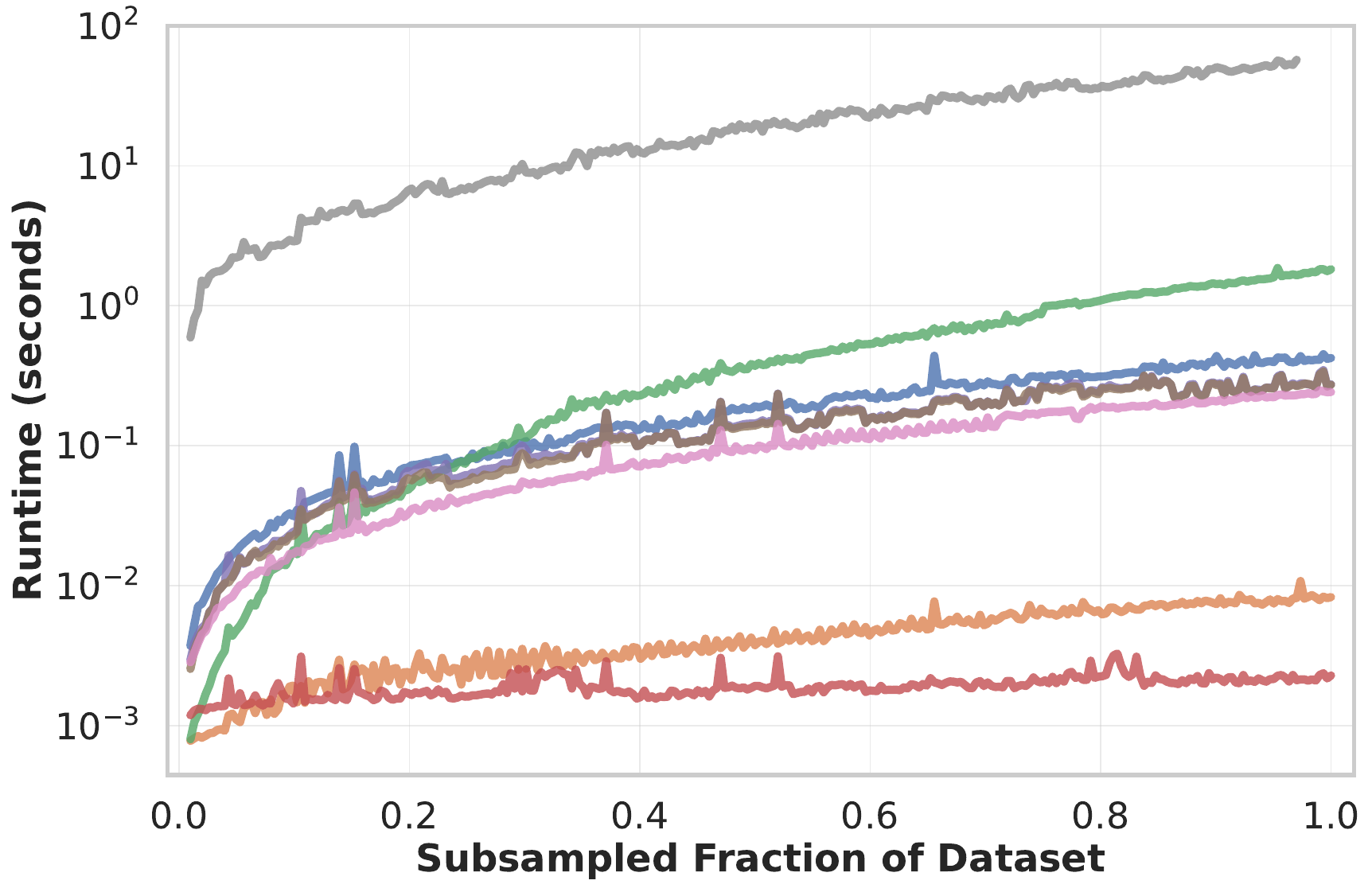}
    \\[2mm]
    (s) trec
\end{minipage}
\vspace{6pt}
\begin{minipage}{\linewidth}
    \centering
    \includegraphics[width=\linewidth]{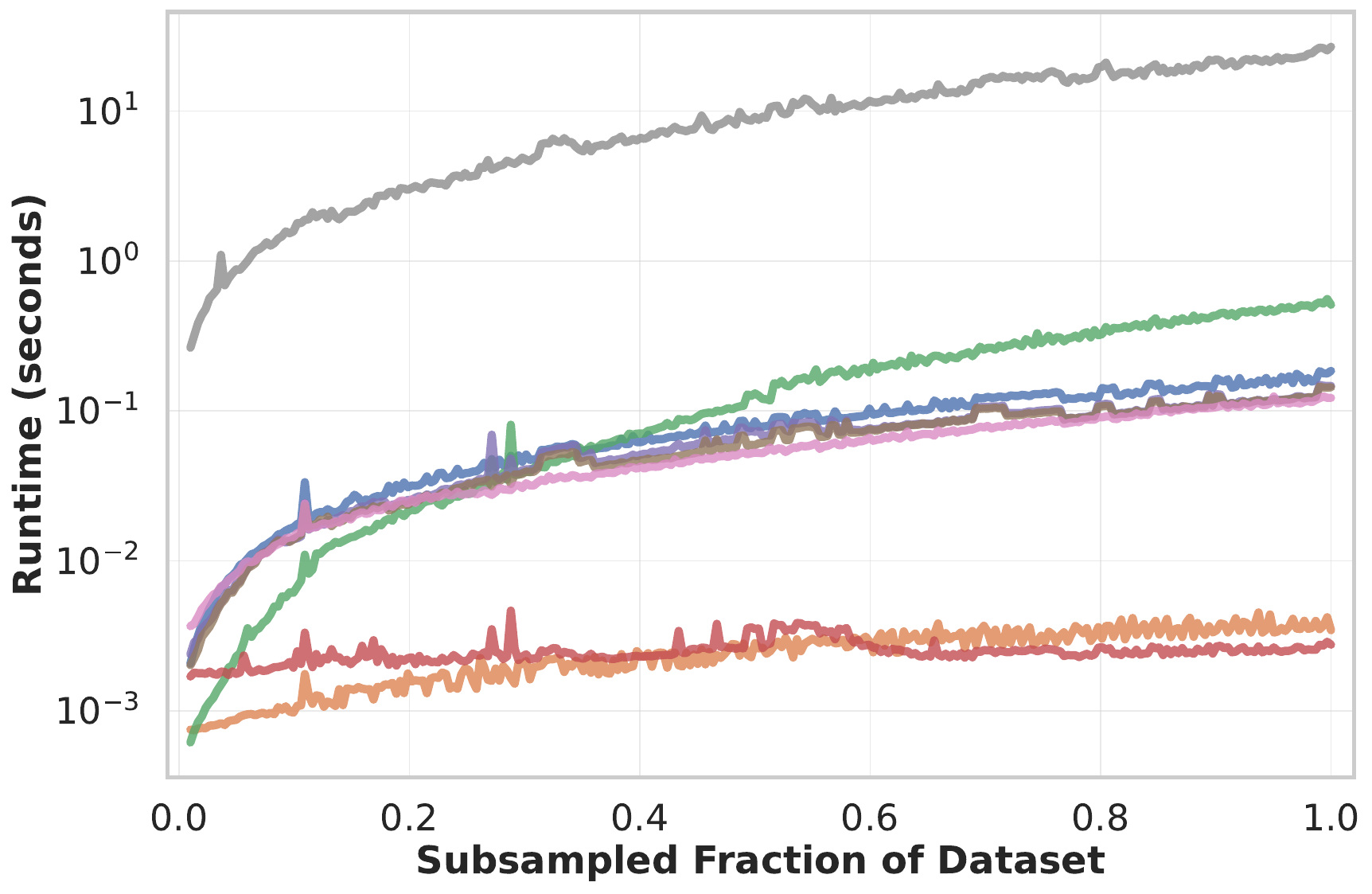}
    \\[2mm]
    (t) usps
\end{minipage}
\vspace{6pt}
\captionof{figure}{Times taken to compute each metric on the t-SNE projection of each dataset.}
\label{fig:times-taken}
}

\section{Stability of \ourmetricAbbrev}
\label{sec:stability}

We verify that we can achieve reasonable accuracy on \ourmetricAbbrev\ by sampling $k = \mathcal{O}(n)$ triplets (where $n$ is the number of samples in the dataset) by plotting the distribution of values \ourmetricAbbrev\ takes for each dataset in \autoref{fig:stability-all}.

\begin{center}
\begin{minipage}{\linewidth}
    \centering
    \includegraphics[width=\linewidth]{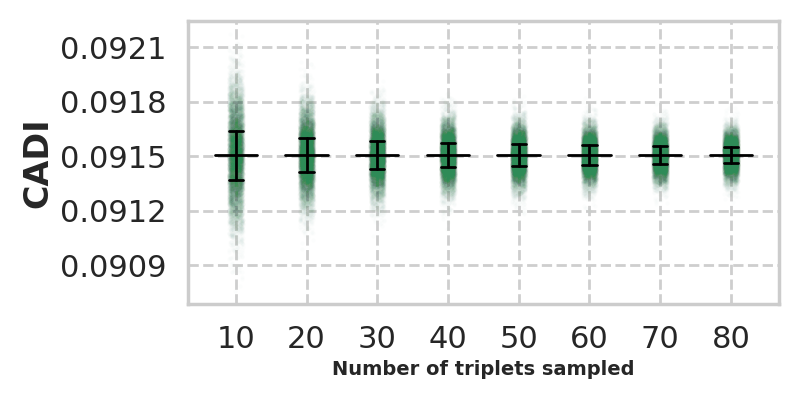}
    \\[2mm]
    (a) acl\_imdb
\end{minipage}
\par\medskip
\begin{minipage}{\linewidth}
    \centering
    \includegraphics[width=\linewidth]{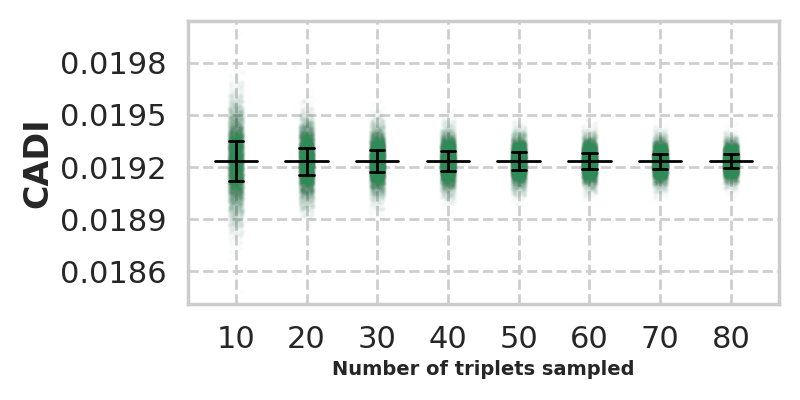}
    \\[2mm]
    (b) coil100
\end{minipage}
\par\medskip
\begin{minipage}{\linewidth}
    \centering
    \includegraphics[width=\linewidth]{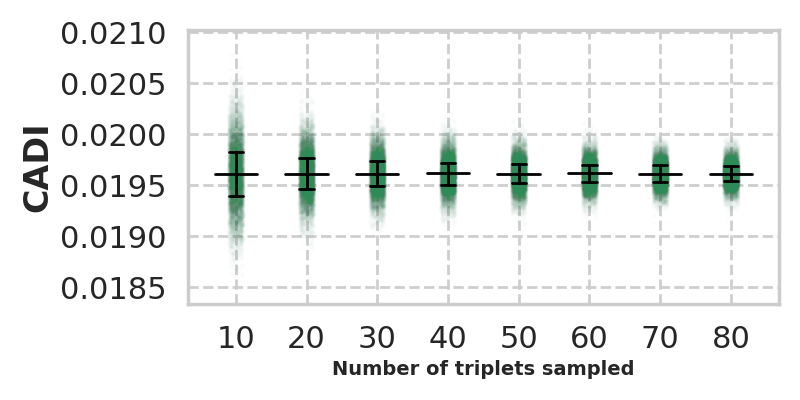}
    \\[2mm]
    (c) coil20
\end{minipage}
\par\medskip
\begin{minipage}{\linewidth}
    \centering
    \includegraphics[width=\linewidth]{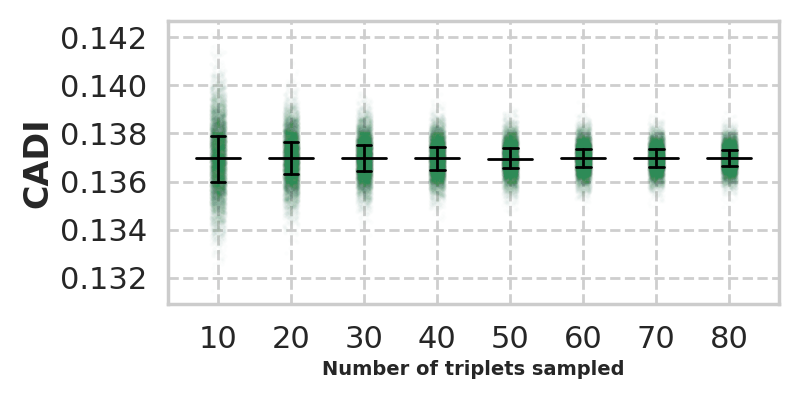}
    \\[2mm]
    (d) concentric3
\end{minipage}
\par\medskip
\begin{minipage}{\linewidth}
    \centering
    \includegraphics[width=\linewidth]{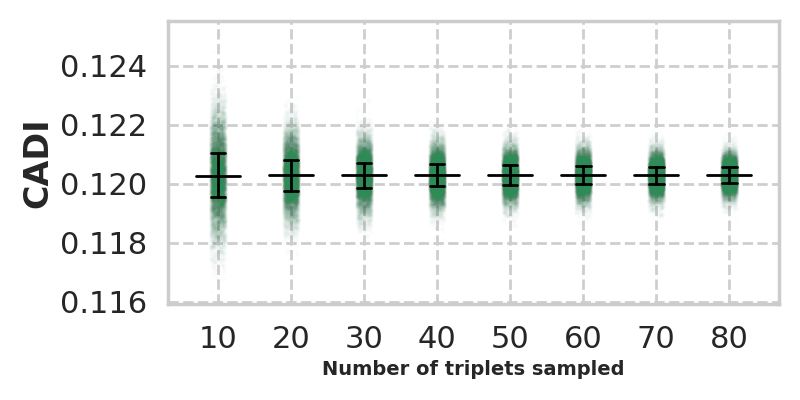}
    \\[2mm]
    (e) concentric4
\end{minipage}
\par\medskip
\begin{minipage}{\linewidth}
    \centering
    \includegraphics[width=\linewidth]{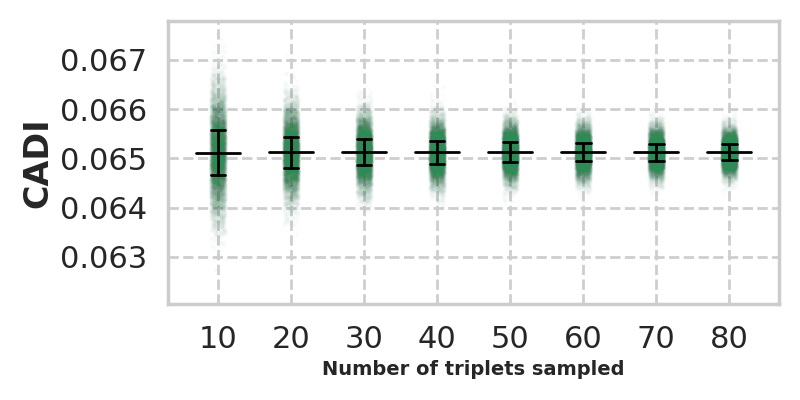}
    \\[2mm]
    (f) donuts
\end{minipage}
\par\medskip
\begin{minipage}{\linewidth}
    \centering
    \includegraphics[width=\linewidth]{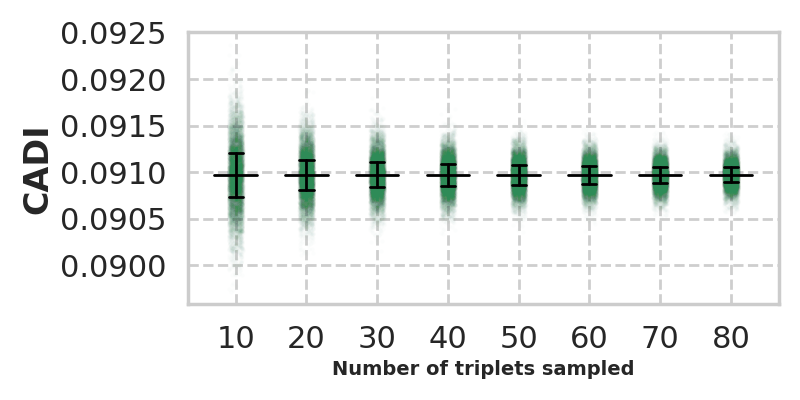}
    \\[2mm]
    (g) emotion
\end{minipage}
\par\medskip
\begin{minipage}{\linewidth}
    \centering
    \includegraphics[width=\linewidth]{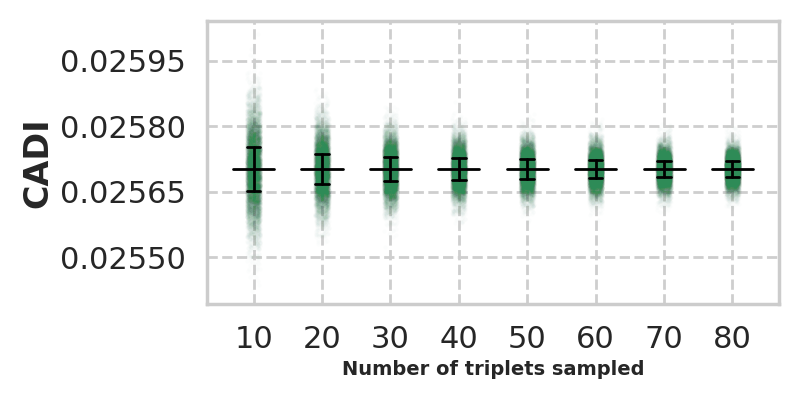}
    \\[2mm]
    (h) F-MNIST
\end{minipage}
\par\medskip
\begin{minipage}{\linewidth}
    \centering
    \includegraphics[width=\linewidth]{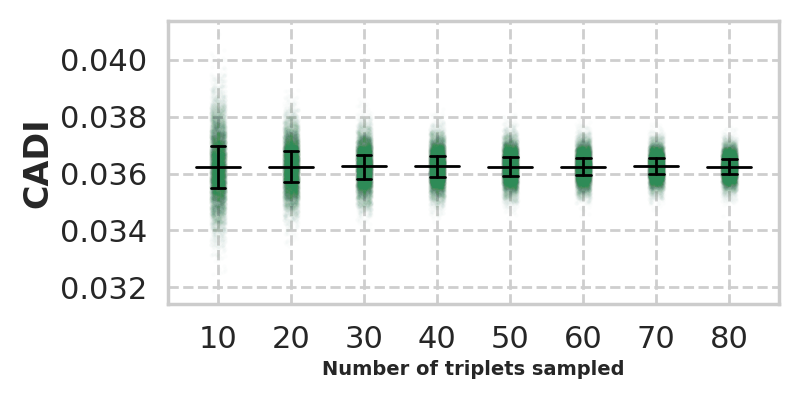}
    \\[2mm]
    (i) liver
\end{minipage}
\par\medskip
\begin{minipage}{\linewidth}
    \centering
    \includegraphics[width=\linewidth]{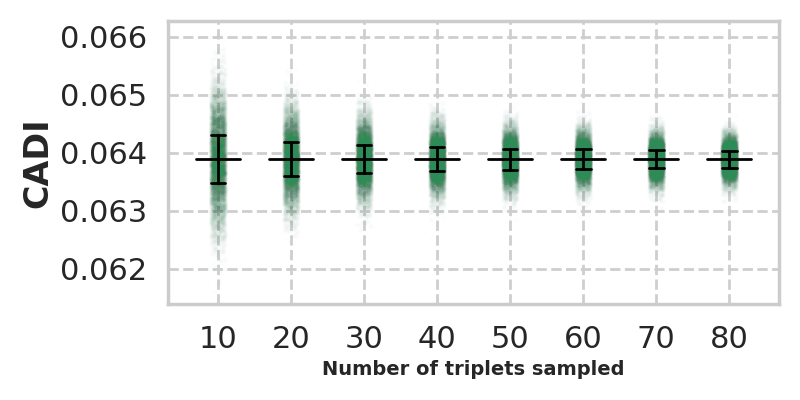}
    \\[2mm]
    (j) matryoshka
\end{minipage}
\par\medskip
\begin{minipage}{\linewidth}
    \centering
    \includegraphics[width=\linewidth]{images/stability10/MNIST.png}
    \\[2mm]
    (k) MNIST
\end{minipage}
\par\medskip
\begin{minipage}{\linewidth}
    \centering
    \includegraphics[width=\linewidth]{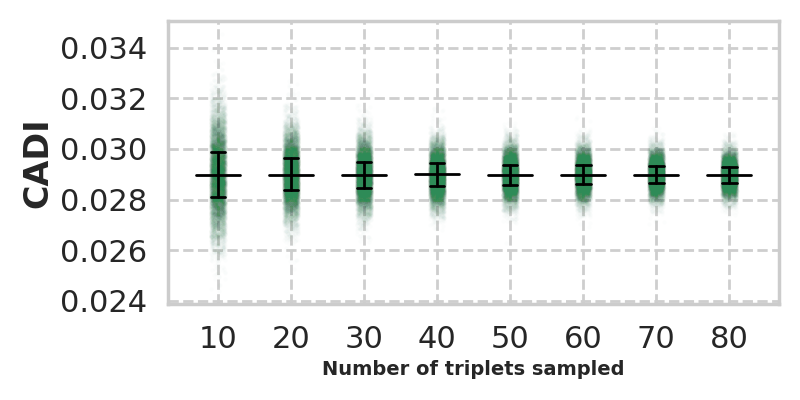}
    \\[2mm]
    (l) olivetti
\end{minipage}
\par\medskip
\begin{minipage}{\linewidth}
    \centering
    \includegraphics[width=\linewidth]{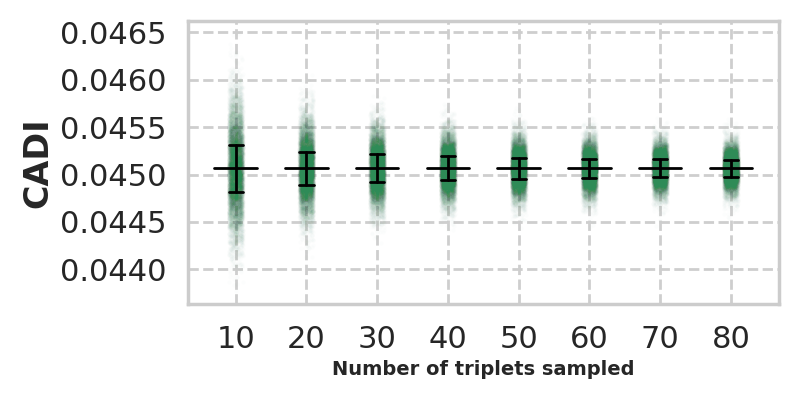}
    \\[2mm]
    (m) pbmc3k
\end{minipage}
\par\medskip
\begin{minipage}{\linewidth}
    \centering
    \includegraphics[width=\linewidth]{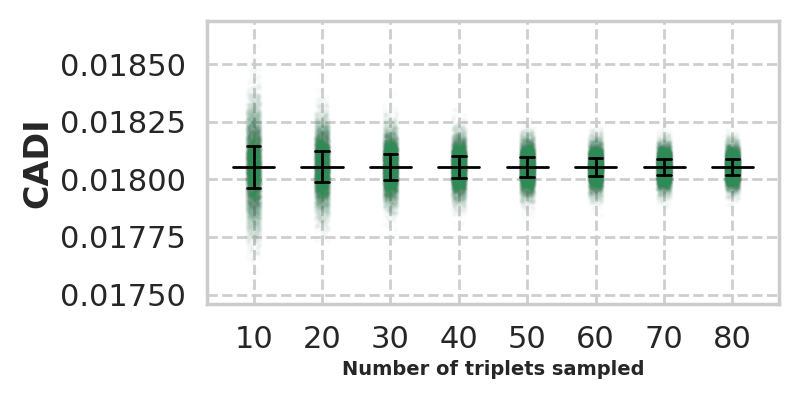}
    \\[2mm]
    (n) pendigits
\end{minipage}
\par\medskip
\begin{minipage}{\linewidth}
    \centering
    \includegraphics[width=\linewidth]{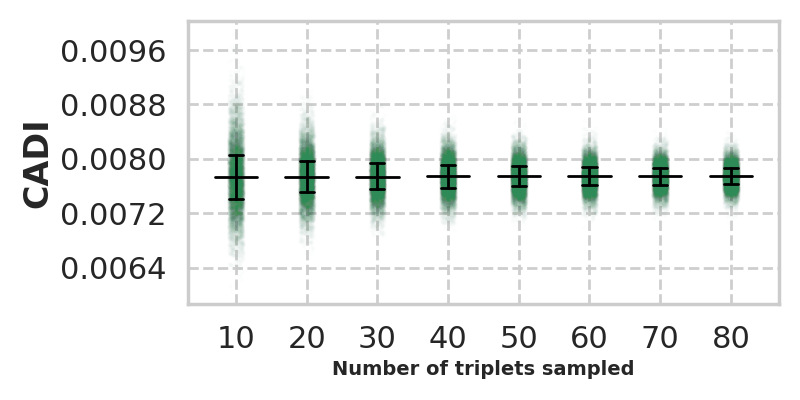}
    \\[2mm]
    (o) penguins
\end{minipage}
\par\medskip
\begin{minipage}{\linewidth}
    \centering
    \includegraphics[width=\linewidth]{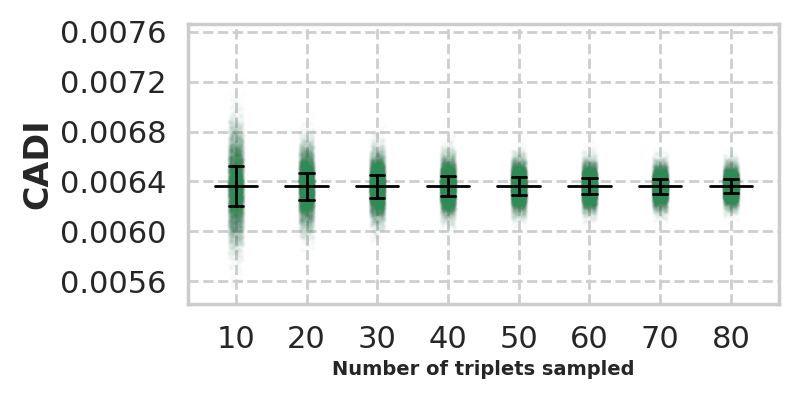}
    \\[2mm]
    (p) rings
\end{minipage}
\par\medskip
\begin{minipage}{\linewidth}
    \centering
    \includegraphics[width=\linewidth]{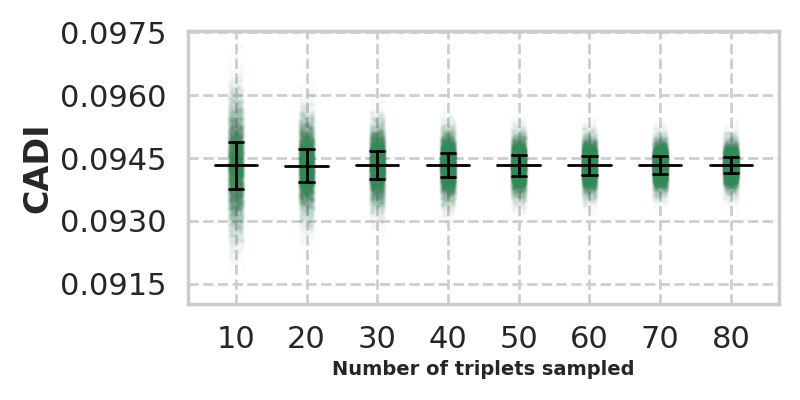}
    \\[2mm]
    (q) sentiment
\end{minipage}
\par\medskip
\begin{minipage}{\linewidth}
    \centering
    \includegraphics[width=\linewidth]{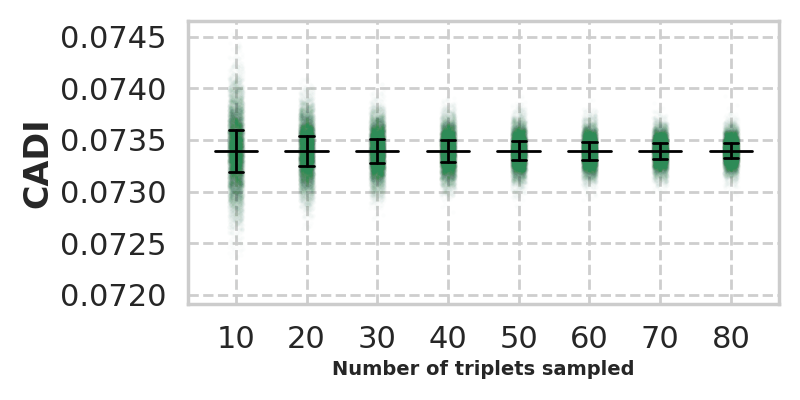}
    \\[2mm]
    (r) trec
\end{minipage}
\par\medskip
\begin{minipage}{\linewidth}
    \centering
    \includegraphics[width=\linewidth]{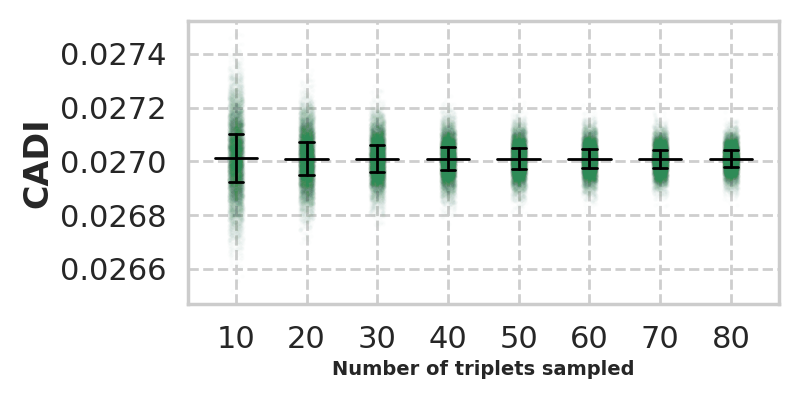}
    \\[2mm]
    (s) usps
\end{minipage}
\par\medskip
\captionof{figure}{ For different numbers of triplets sampled, we calculate \ourmetricAbbrev\ on the t-SNE projection of each dataset on 10,000 different runs each. The distribution of values \ourmetricAbbrev\ takes when sampling different multiples of each dataset is plotted here.}
\label{fig:stability-all}
\end{center}

\section{Analysis of the Liver Dataset}
\label{sec:liver-explain}

We can see in \autoref{fig:liver-analysis} that normal, healthy cells have much less genetic variation than cancer cells. This can be explained by the fact that accumulating mutations result in higher genetic diversity between cancer cells, making them less uniform than normal, healthy cells.

Therefore, projections such as PCA and AngleEmbedding, which spread out the HCC cluster, most faithfully represent the structure of the dataset at the cluster level. While the MDS projection also spreads out the HCC cluster, there is no clear separation of the two clusters. UMATO also does not separate the two clusters well.

UMAP, PaCMAP, and t-SNE separate the two clusters, but the cluster sizes are not interpretable. However, we also note that the red edges (which encode shorter distances) in the epsilon tend to be quite short, and longer edges are yellower (which encode longer distances in the dataset). This is in contrast to AngleEmbedding and PCA, where red edges aren't meaningfully shorter than the yellow edges. Therefore, it appears that, while these projections do not perfectly capture cluster-level structure, they are, as expected, superior in neighborhood preservation.

\begin{center}
    
\begin{minipage}{\linewidth}
    \centering
    \includegraphics[width=\linewidth]{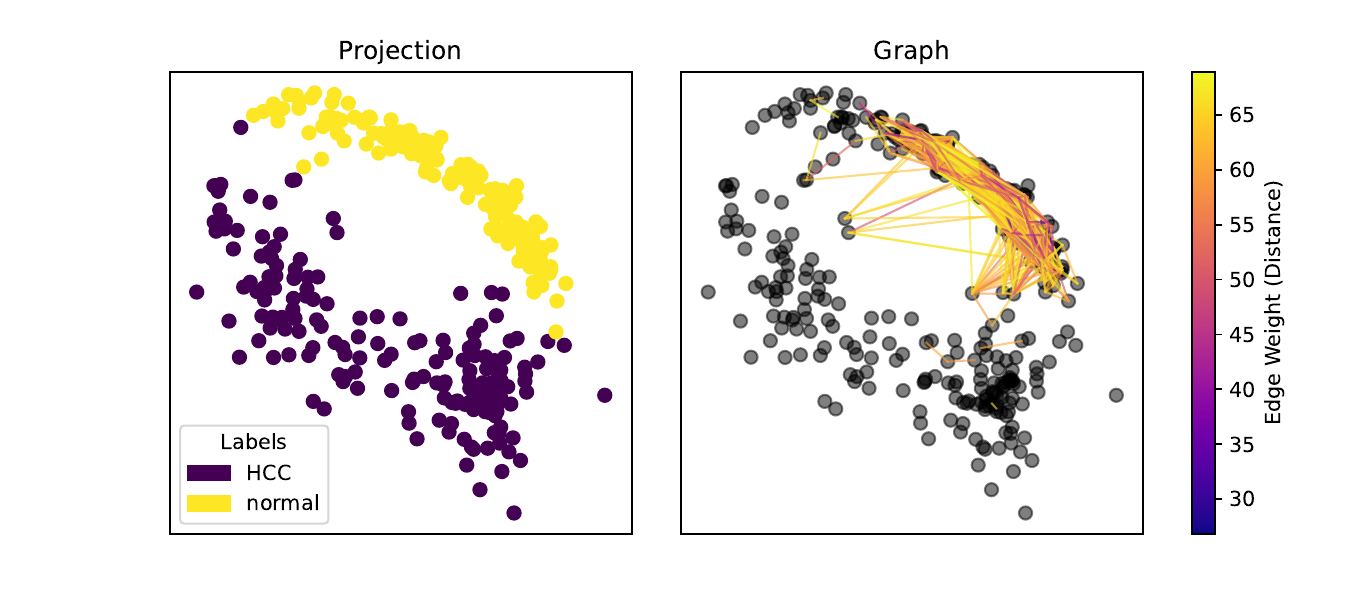}
    \\[2mm]
    (a) AngleEmbedding
\end{minipage}
\par\medskip

\begin{minipage}{\linewidth}
    \centering
    \includegraphics[width=\linewidth]{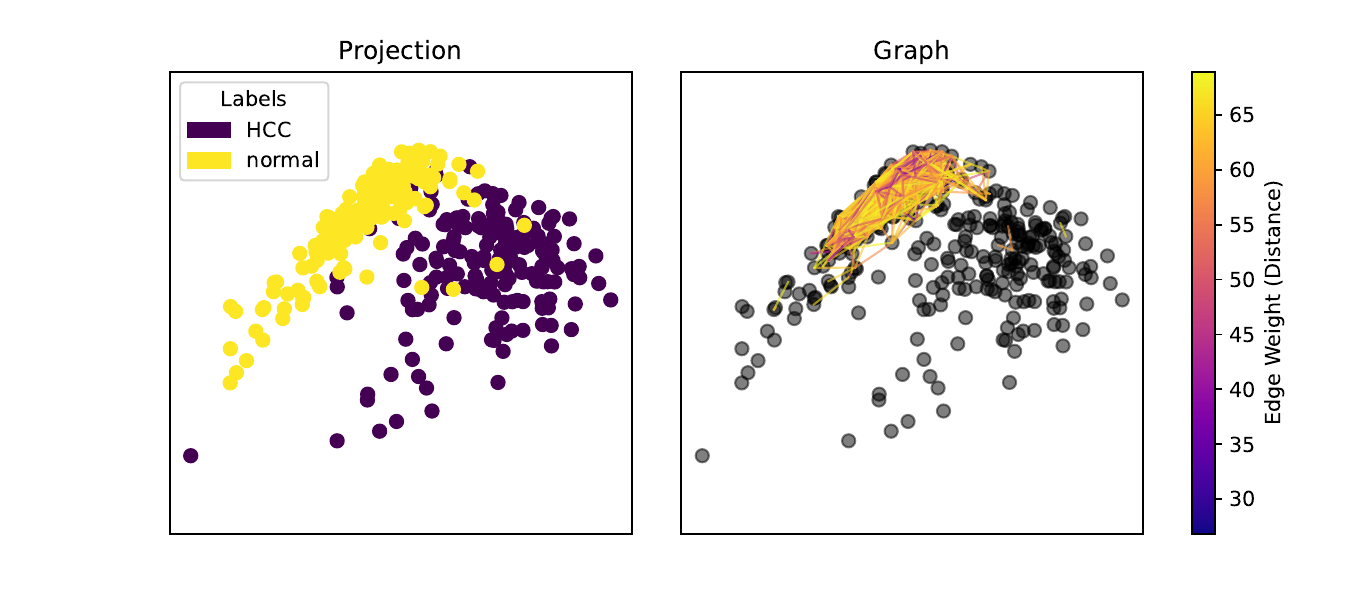}
    \\[2mm]
    (b) PCA
\end{minipage}
\par\medskip

\begin{minipage}{\linewidth}
    \centering
    \includegraphics[width=\linewidth]{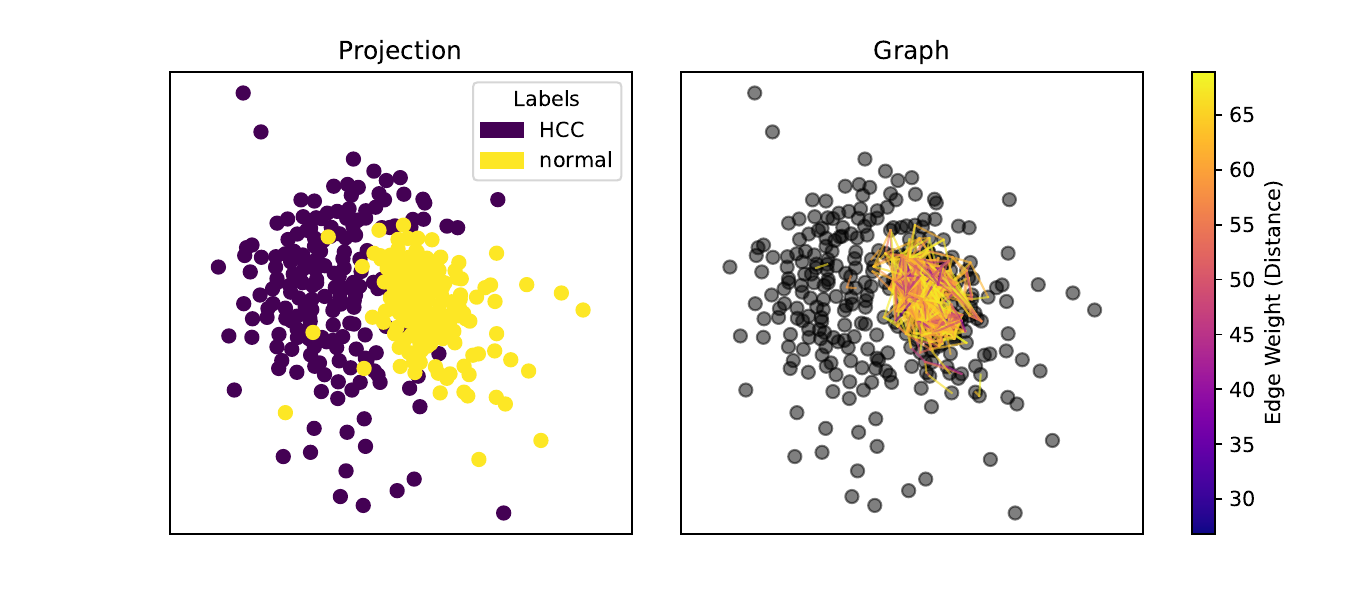}
    \\[2mm]
    (c) MDS
\end{minipage}
\par\medskip

\begin{minipage}{\linewidth}
    \centering
    \includegraphics[width=\linewidth]{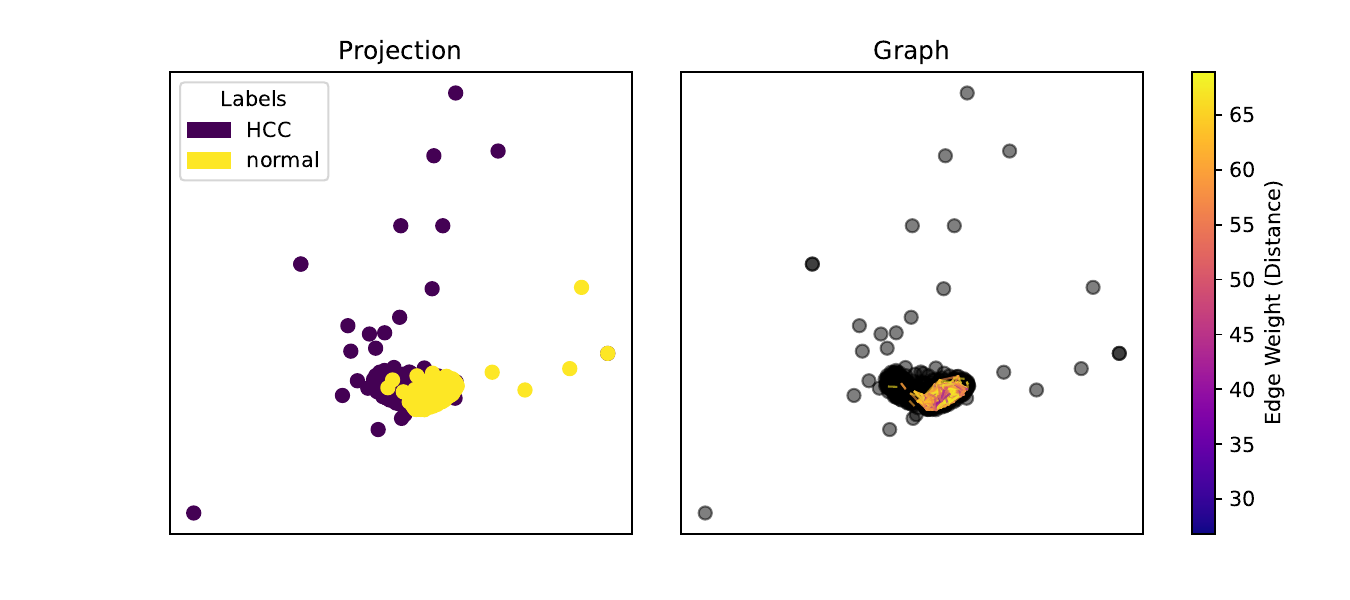}
    \\[2mm]
    (d) UMATO
\end{minipage}
\par\medskip

\begin{minipage}{\linewidth}
    \centering
    \includegraphics[width=\linewidth]{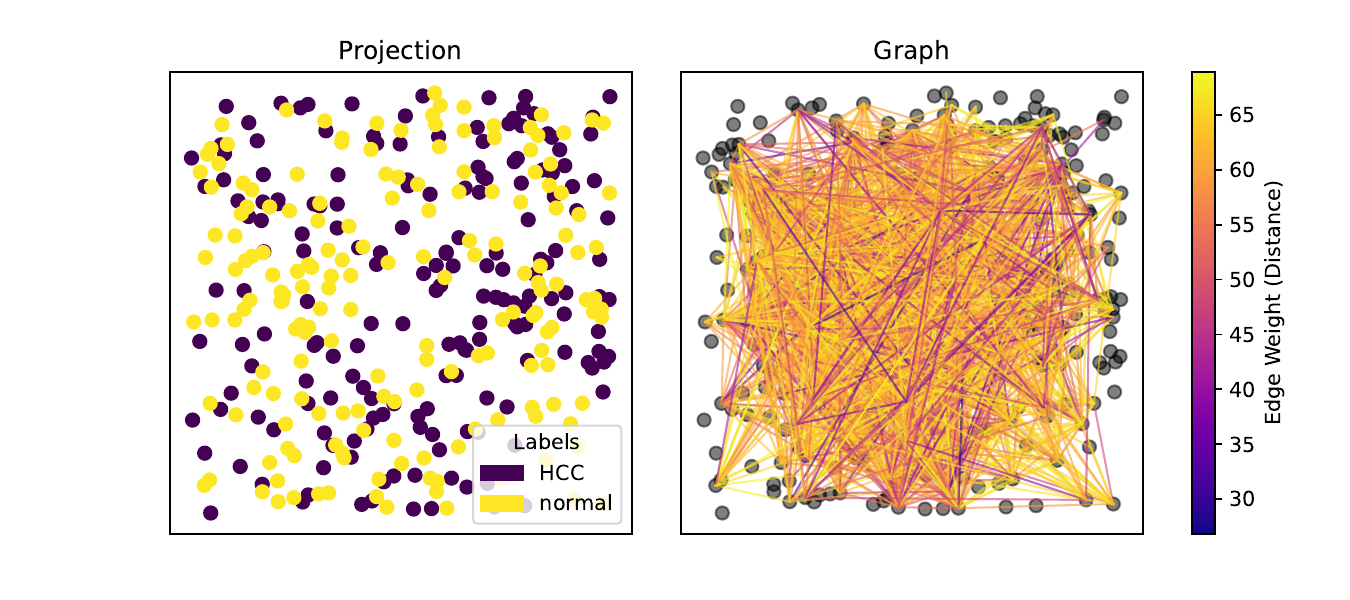}
    \\[2mm]
    (e) Random
\end{minipage}
\par\medskip

\begin{minipage}{\linewidth}
    \centering
    \includegraphics[width=\linewidth]{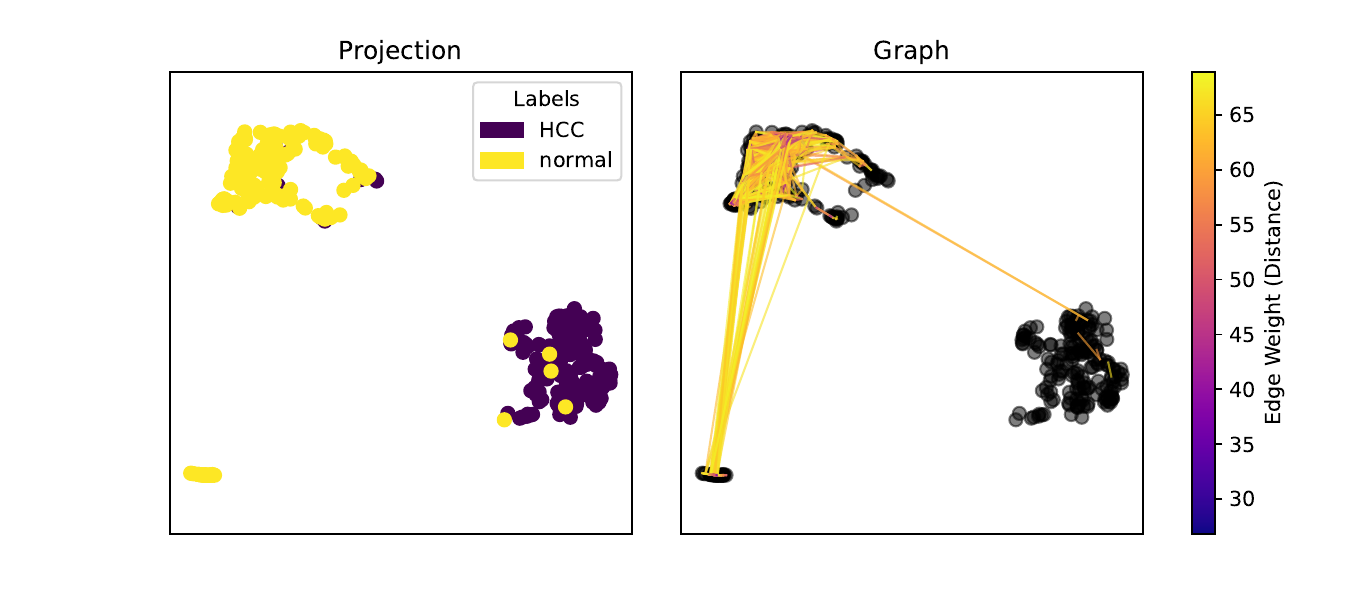}
    \\[2mm]
    (f) PaCMAP
\end{minipage}
\par\medskip

\begin{minipage}{\linewidth}
    \centering
    \includegraphics[width=\linewidth]{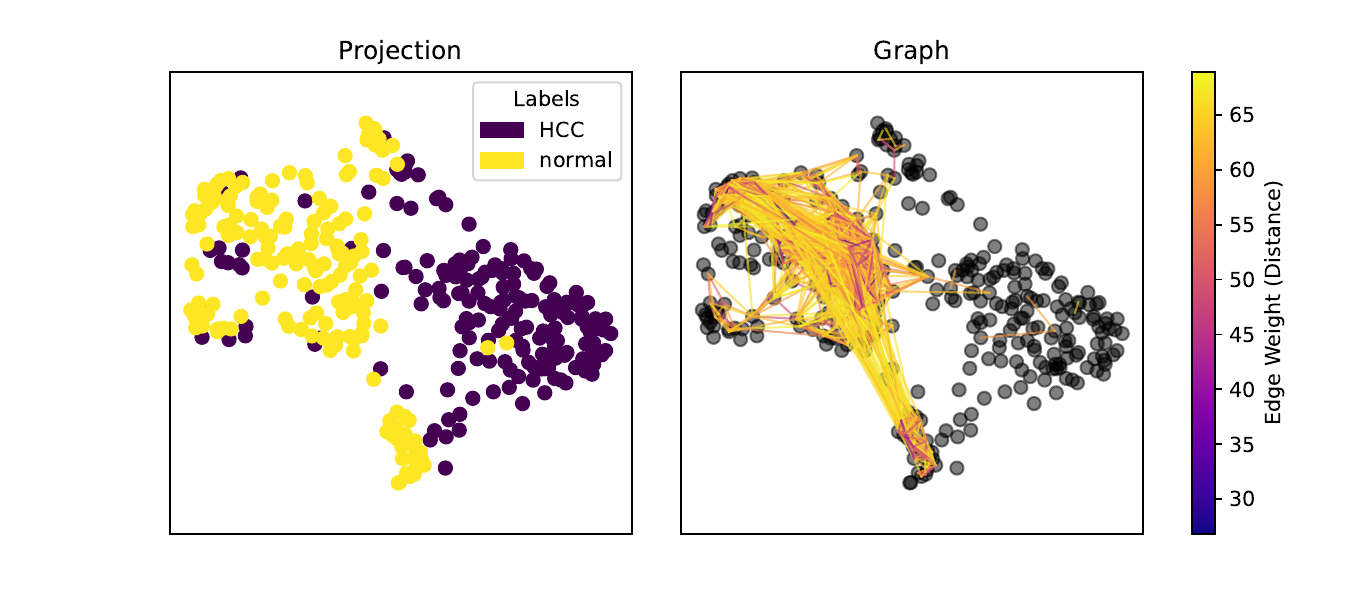}
    \\[2mm]
    (g) UMAP
\end{minipage}
\par\medskip

\begin{minipage}{\linewidth}
    \centering
    \includegraphics[width=\linewidth]{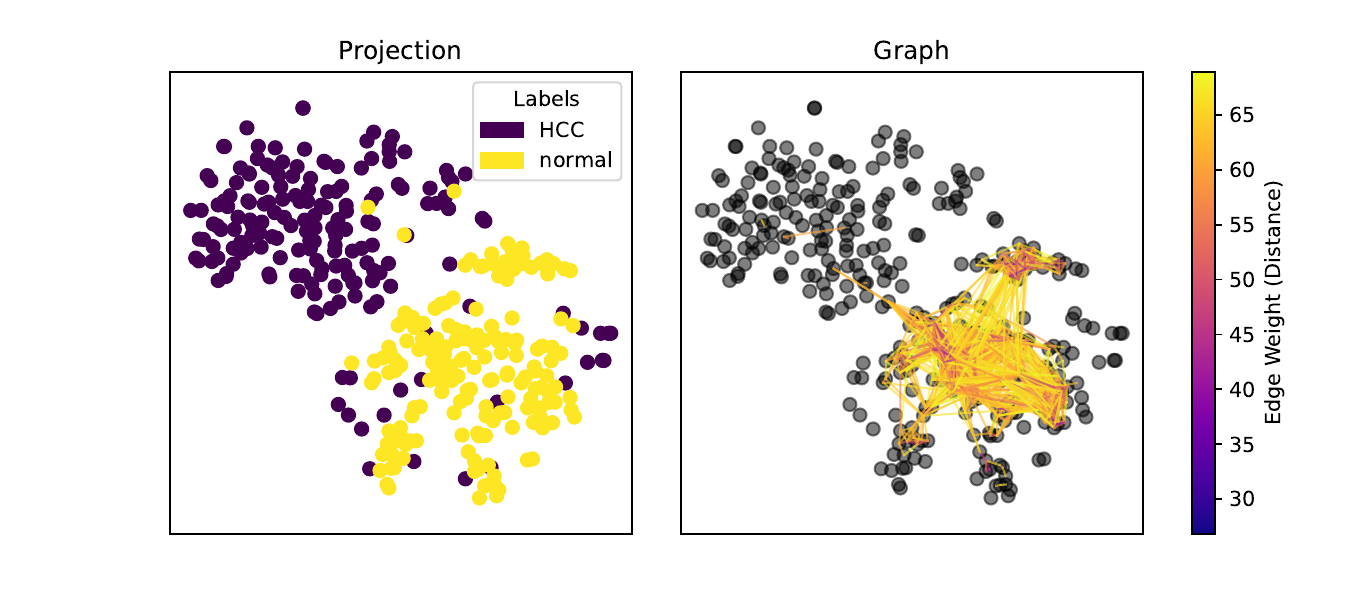}
    \\[2mm]
    (h) t-SNE
\end{minipage}
\par\medskip

\captionof{figure}{
The epsilon graphs of the liver dataset are plotted using each DR projection. Epsilon was selected as the 30th percentile of the distance to the 10th nearest neighbor in the dataset. We can clearly see that the cluster of normal cells is far more tightly packed, while there are barely any edges between any two cancer cells.
}
\label{fig:liver-analysis}

\end{center}

\section{Analysis of the USPS dataset}
\label{sec:usps-explain}

\begin{figure}
    \centering
    \includegraphics[width=\linewidth]{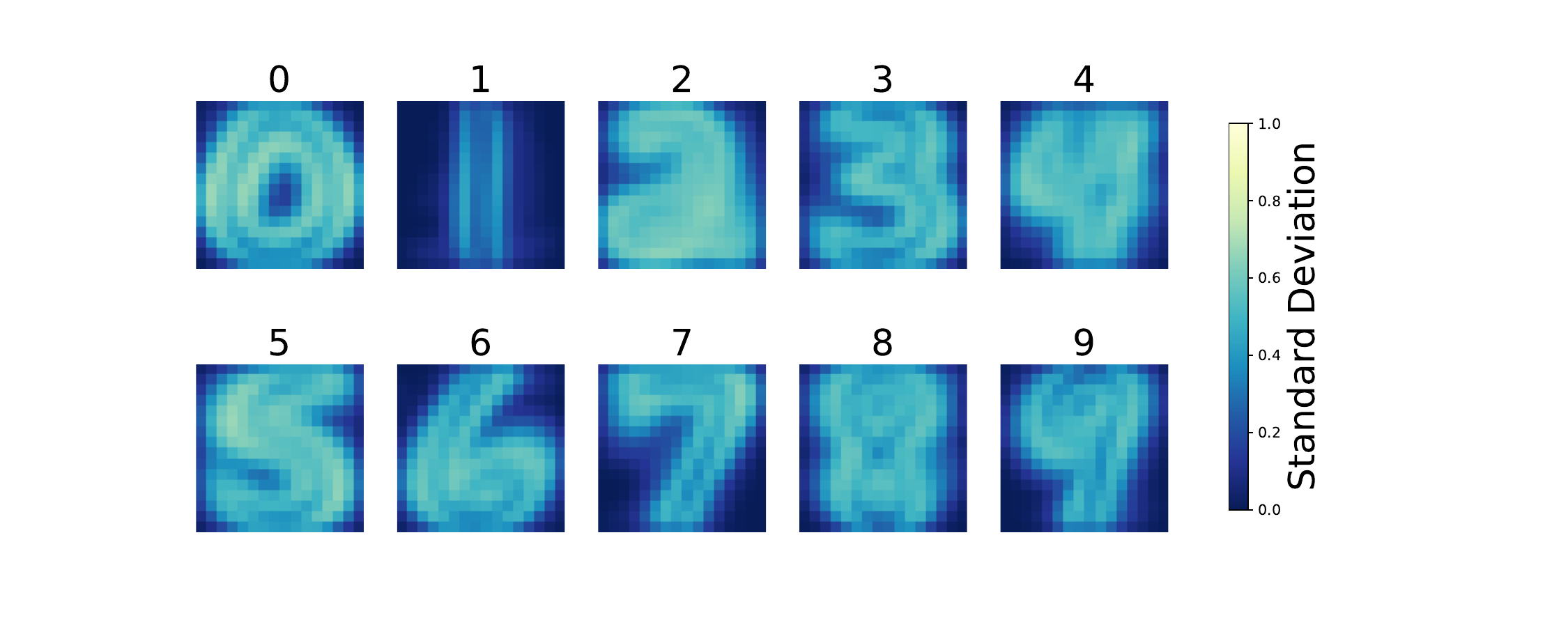}
    \caption{The per-pixel standard deviation across images within each class is visualized for each of the 10 classes in the USPS dataset. We can clearly see that the '1' class, with the darkest figure, has the least variation within its images.}
    \label{fig:usps-class-std}
\end{figure}

The USPS dataset is composed of $16\times16$ px images of handwritten digits from 0 to 9. We note in \autoref{tab:usps-stats} that each class has different sizes. The '0' class is by far the largest in number, with around twice as many points as most of the other classes. It is also the largest in extent, with the greatest distance between two points. The '1' class is also quite large in number compared to the others, but it is also the smallest in extent. We also visualize the per-pixel standard deviation within each class in \autoref{fig:usps-class-std}, where it is quite clear that there is very little variation between the images in the '1' class compared to the other classes. 

Therefore, we claim that a projection that preserves structure at the cluster level should create differently sized clusters for each class in the USPS dataset, and note that the AngleEmbedding projection, visualized in \autoref{tab:usps-full}, plots the '0' class the largest and the '1' class the smallest.

\begin{table}[h!]
    \centering
    \begin{tabular}{lcc}
    \toprule
     & Max Distance & Number of Points \\
    Class &  &  \\
    \midrule
    \textbf{0} & 20.728103 & 1553 \\
    \textbf{1} & 15.310417 & 1269 \\
    \textbf{2} & 20.363129 & 929 \\
    \textbf{3} & 18.374678 & 824 \\
    \textbf{4} & 18.278038 & 852 \\
    \textbf{5} & 20.666971 & 716 \\
    \textbf{6} & 17.877049 & 834 \\
    \textbf{7} & 17.863816 & 792 \\
    \textbf{8} & 19.844467 & 708 \\
    \textbf{9} & 17.541731 & 821 \\
    \bottomrule
    \end{tabular}
    \caption{For each class in the USPS dataset, we report the maximum distance between two points in the same class, as well as the total number of points in that class.}
    \label{tab:usps-stats}
\end{table}

\section{Analysis of the TREC dataset}

The TREC dataset consists of open-domain factoid questions, which are classified into 6 classes based on the semantic type of answer expected: ABBR, ENTY, DESC, HUM, LOC, and NUM. ABBR includes questions about abbreviations or their expansions (e.g., acronyms); ENTY covers questions seeking a specific entity such as an object, substance, animal, event, or concept; DESC contains definition or explanatory questions that require descriptive answers rather than short facts; HUM encompasses questions about individuals or groups of people; LOC refers to questions whose answers are geographical locations or places; and NUM includes questions expecting numerical responses such as dates, counts, measurements, or other quantities.

\begin{figure}
    \centering
    \includegraphics[width=0.49\linewidth]{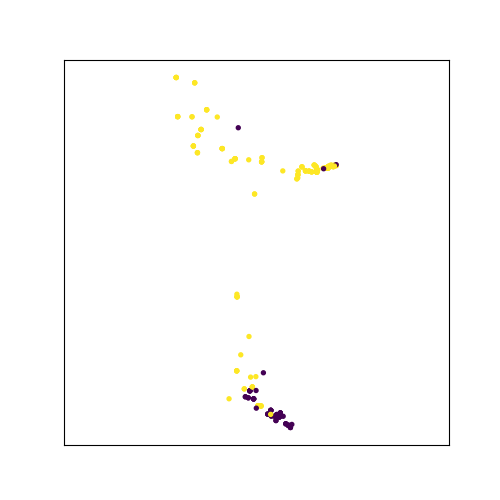}
    \includegraphics[width=0.49\linewidth]{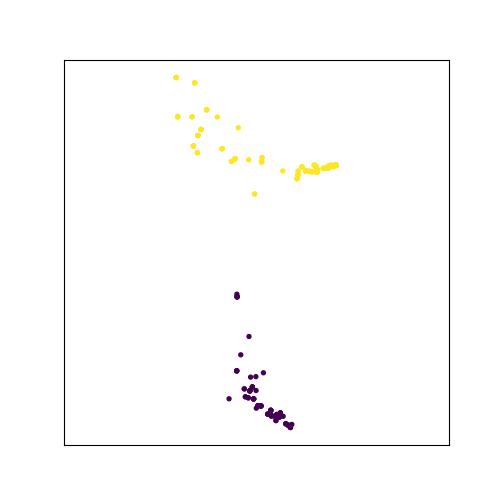}
    \parbox[c]{0.49\linewidth}{\centering (a) K-Means on dataset}
    \parbox[c]{0.49\linewidth}{\centering (b) K-Means on AngleEmbedding projection}
    \caption{The ABBR class of the AngleEmbedding projection, isolated, and labelled with K-Means clustering on (a) the original dataset and (b) the AngleEmbedding projection.}
    \label{fig:trec-kmeans}
\end{figure}

We note that the AngleEmbedding projection splits the 'ABBR' class into two clusters. We ran k-means clustering with k = 2 on the 'ABBR' class in both the projection and dataset; we observe that the two labellings agree 86\% of the time, as we visualize in \autoref{fig:trec-kmeans}.

\begin{figure}
    \centering
    \includegraphics[width=0.32\linewidth]{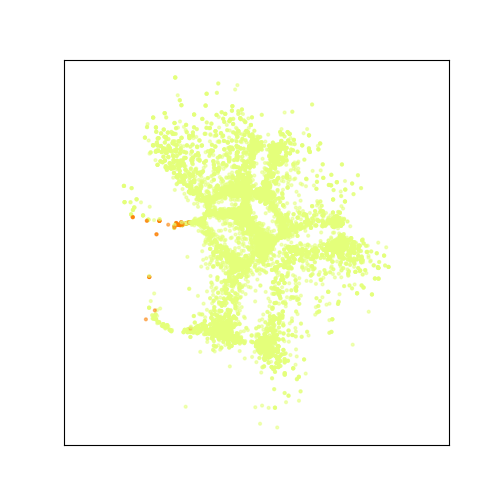}
    \includegraphics[width=0.32\linewidth]{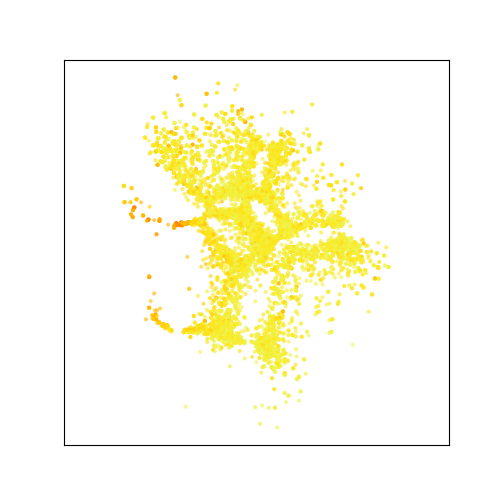}  
    \includegraphics[width=0.32\linewidth]{images/emb_figs/trec/AngleEmbedding.png}

    \parbox[c]{0.32\linewidth}{\centering (a) Explicitly contains derivatives of 'abbreviate' and 'acronym'}
    \parbox[c]{0.32\linewidth}{\centering (b) Politics \& Business related terms}
    \parbox[c]{0.32\linewidth}{\centering (c) Class labels}
    \caption{AngleEmbedding projection of the TREC dataset. (a) Instances explicitly containing derivatives of “abbreviate” or “acronym” are highlighted in dark. (b) Points are shaded according to cosine similarity to politics- and business-related queries, with darker shades indicating higher similarity. (c) Points are colored by class label; the ABBR class (purple) forms two distinct clusters.}
    \label{fig:trec-heatmap}
\end{figure}

To analyze why this occurs, we analyzed the two clusters formed by AngleEmbedding and made the following observations, which we visualize in \autoref{fig:trec-heatmap}:
\begin{itemize}
    \item While all questions in this class relate to abbreviations, the sentences of one cluster explicitly contain some derivative of the words "abbreviation" or "acronym" 43\% (65 instances) of the time. While the second cluster also contains such questions, they only occur 6.7\% (6 instances) of the time.
    \item The same first cluster also seems to contain more sentences pertaining to topics that are related to world politics and business, e.g. "What is the abbreviation of General Motors ?", "CNN is the abbreviation for what ?" We observe that this cluster has greater cosine similarity to such topics than others.
\end{itemize}

\section{Analysis of Emotion Dataset}

\begin{figure}
    \centering
    \includegraphics[width=0.49\linewidth]{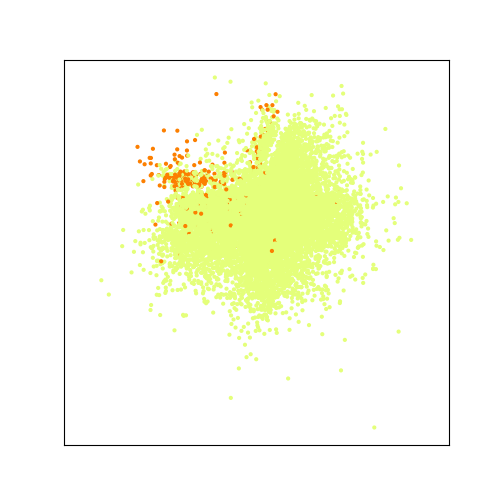}
    \includegraphics[width=0.49\linewidth]{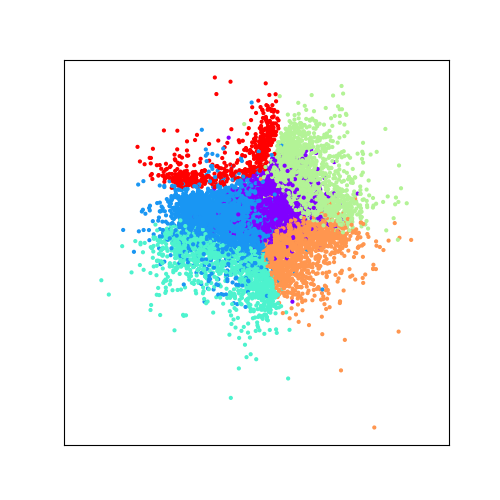}
    \parbox[c]{0.49\linewidth}{\centering (a) Contains derivatives of 'amaze' and 'impress'}
    \parbox[c]{0.49\linewidth}{\centering (b) Class labels}
    \caption{AngleEmbedding projection of the Emotion dataset. (a) Instances containing derivatives of the words 'amaze' and 'impress' are highlighted in dark orange. (b) Points are colored by class label; the 'surprise' class (red) forms two distinct clusters.}
    \label{fig:emotion-heatmap}
\end{figure}

The Emotion dataset, intended for sentiment analysis, is composed of various sentences classified by emotional tone: sadness, joy, love, anger, fear, and surprise. We notice that while AngleEmbedding generally splits the text data into 6 classes, it also splits the 'surprise' class into 2 clusters. It turns out that while both clusters contain sentences that convey surprise, one of them explicitly contains words semantically related to surprise, such as derivatives of 'amaze' and 'impress'. We visualize this observation in \autoref{fig:emotion-heatmap}.

\section{Results of Evaluation}
Here, we include results on all datasets, computing all metric values for all techniques, displaying all projections and values in a table.

\begin{figure*}[t!]
    \centering
    \includegraphics[width=\linewidth]{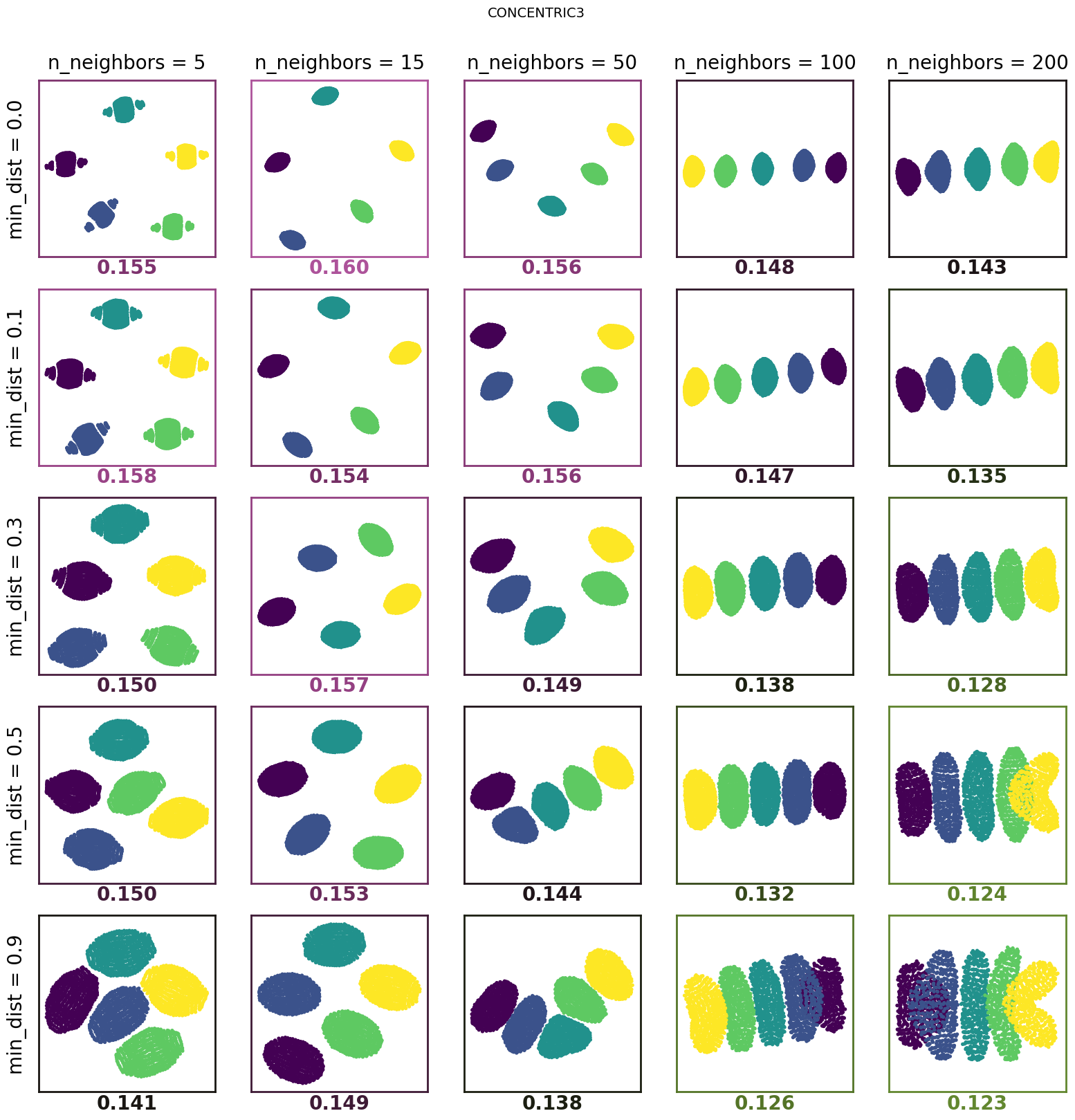}
    \caption{\ourmetricAbbrev\ scores for UMAP projections of the concentric3 dataset with different values for the hyperparameters \texttt{min\_dist} and \texttt{n\_neighbors}. Green borders indicate better scores, while red borders indicate worse scores. Note that all UMAP projections still perform relatively badly on \ourmetricAbbrev; scores for MDS, AngleEmbedding, and PCA are an order of magnitude lower. }\vspace{-0.4cm}
    \label{fig:umap-hparam2}
\end{figure*}

\begin{table*}[ht!]
\centering
\caption{Embedding quality metrics for \textbf{rings} dataset.}
\vspace{0.5em}
\begin{tabular}{lccc}
\toprule
\textbf{Metric} & \textbf{AngleEmbedding} & \textbf{MDS} & \textbf{PCA} \\
\midrule
 & \includegraphics[width=0.18\textwidth]{images/emb_figs/rings/AngleEmbedding.png} & \includegraphics[width=0.18\textwidth]{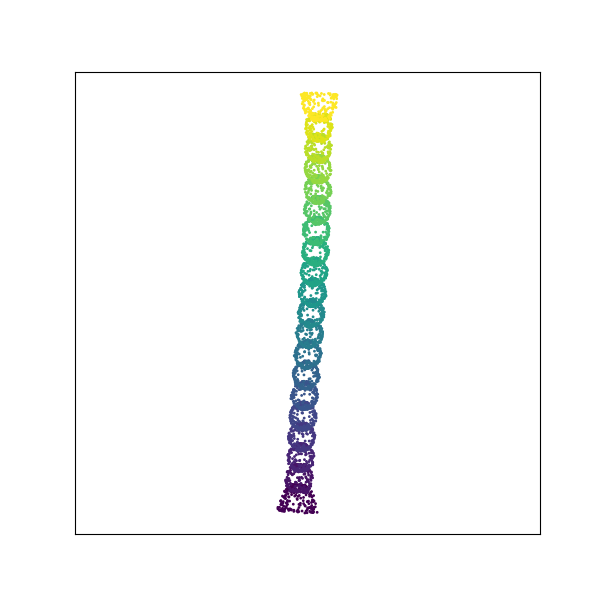} & \includegraphics[width=0.18\textwidth]{images/emb_figs/rings/PCA.png} \\
\midrule
{\ourmetricFull} & \cellcolor{myblue!70}\textcolor{white}{\textbf{0.0012}} & \cellcolor{myblue!60}\textcolor{white}{0.0037} & \cellcolor{myblue!50}\textcolor{white}{0.0040} \\
{Cluster Distance Score} & \cellcolor{myblue!50}\textcolor{white}{0.0013} & \cellcolor{myblue!60}\textcolor{white}{0.0004} & \cellcolor{myblue!70}\textcolor{white}{\textbf{0.0000}} \\
{Label-T\&C} & \cellcolor{myblue!42}\textcolor{black}{0.9994} & \cellcolor{myblue!56}\textcolor{white}{0.9998} & \cellcolor{myblue!70}\textcolor{white}{\textbf{1.0000}} \\
{Steadiness \& Cohesiveness} & \cellcolor{myblue!50}\textcolor{white}{0.5725} & \cellcolor{myblue!70}\textcolor{white}{\textbf{0.6321}} & \cellcolor{myblue!60}\textcolor{white}{0.6009} \\
{Silhouette Score} & \cellcolor{myblue!20}\textcolor{black}{0.0768} & \cellcolor{myblue!10}\textcolor{black}{0.0723} & \cellcolor{myblue!30}\textcolor{black}{0.1093} \\
{Davies Bouldin Index} & \cellcolor{myblue!10}\textcolor{black}{1.2882} & \cellcolor{myblue!20}\textcolor{black}{1.1725} & \cellcolor{myblue!30}\textcolor{black}{0.9086} \\
{Normalized Mutual Information} & \cellcolor{myblue!28}\textcolor{black}{0.0074} & \cellcolor{myblue!42}\textcolor{black}{0.0095} & \cellcolor{myblue!0}\textcolor{black}{0.0000} \\
{Adjusted Rand Index} & \cellcolor{myblue!23}\textcolor{black}{0.0002} & \cellcolor{myblue!23}\textcolor{black}{0.0003} & \cellcolor{myblue!0}\textcolor{black}{0.0000} \\
\bottomrule
\end{tabular}
\vspace{0.5em}
\begin{tabular}{lccc}
\toprule
\textbf{Metric} & \textbf{PaCMap} & \textbf{Random} & \textbf{TSNE} \\
\midrule
 & \includegraphics[width=0.18\textwidth]{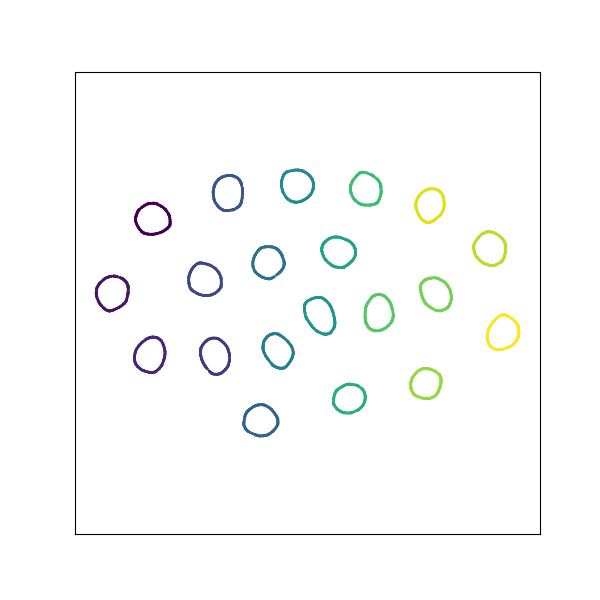} & \includegraphics[width=0.18\textwidth]{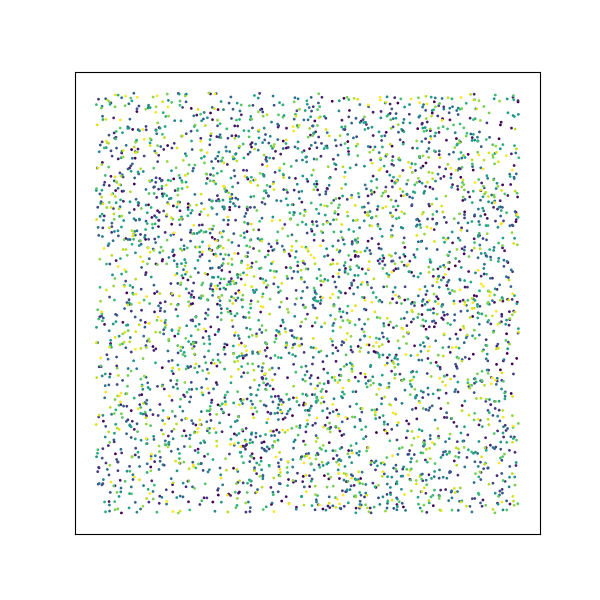} & \includegraphics[width=0.18\textwidth]{images/emb_figs/rings/TSNE.png} \\
\midrule
{\ourmetricFull} & \cellcolor{myblue!10}\textcolor{black}{0.0066} & \cellcolor{myblue!0}\textcolor{black}{0.1753} & \cellcolor{myblue!20}\textcolor{black}{0.0063} \\
{Cluster Distance Score} & \cellcolor{myblue!30}\textcolor{black}{0.3086} & \cellcolor{myblue!0}\textcolor{black}{0.6682} & \cellcolor{myblue!20}\textcolor{black}{0.3741} \\
{Label-T\&C} & \cellcolor{myblue!13}\textcolor{black}{0.9883} & \cellcolor{myblue!0}\textcolor{black}{0.7124} & \cellcolor{myblue!13}\textcolor{black}{0.9883} \\
{Steadiness \& Cohesiveness} & \cellcolor{myblue!30}\textcolor{black}{0.4088} & \cellcolor{myblue!0}\textcolor{black}{0.2469} & \cellcolor{myblue!10}\textcolor{black}{0.3880} \\
{Silhouette Score} & \cellcolor{myblue!60}\textcolor{white}{0.6674} & \cellcolor{myblue!0}\textcolor{black}{-0.0568} & \cellcolor{myblue!50}\textcolor{white}{0.6634} \\
{Davies Bouldin Index} & \cellcolor{myblue!60}\textcolor{white}{0.4791} & \cellcolor{myblue!0}\textcolor{black}{83.7668} & \cellcolor{myblue!50}\textcolor{white}{0.4835} \\
{Normalized Mutual Information} & \cellcolor{myblue!70}\textcolor{white}{\textbf{1.0000}} & \cellcolor{myblue!13}\textcolor{black}{0.0024} & \cellcolor{myblue!70}\textcolor{white}{\textbf{1.0000}} \\
{Adjusted Rand Index} & \cellcolor{myblue!70}\textcolor{white}{\textbf{1.0000}} & \cellcolor{myblue!0}\textcolor{black}{0.0000} & \cellcolor{myblue!70}\textcolor{white}{\textbf{1.0000}} \\
\bottomrule
\end{tabular}
\vspace{0.5em}
\begin{tabular}{lcc}
\toprule
\textbf{Metric} & \textbf{UMAP} & \textbf{UMATO} \\
\midrule
 & \includegraphics[width=0.18\textwidth]{images/emb_figs/rings/UMAP.png} & \includegraphics[width=0.18\textwidth]{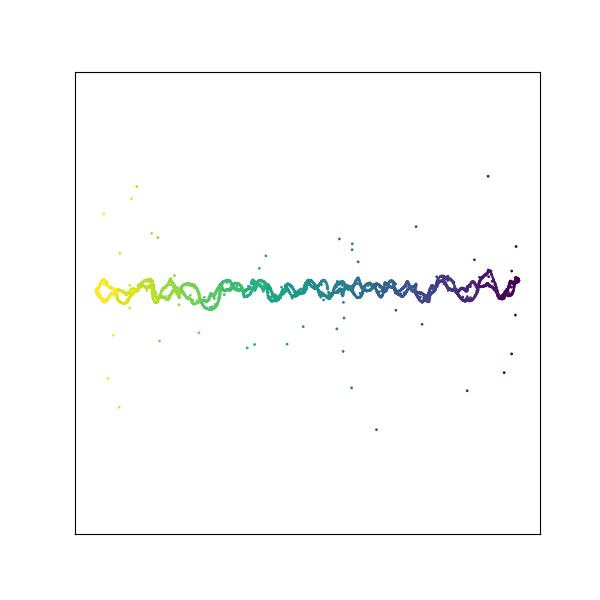} \\
\midrule
{\ourmetricFull} & \cellcolor{myblue!30}\textcolor{black}{0.0061} & \cellcolor{myblue!40}\textcolor{black}{0.0048} \\
{Cluster Distance Score} & \cellcolor{myblue!10}\textcolor{black}{0.4547} & \cellcolor{myblue!40}\textcolor{black}{0.0062} \\
{Label-T\&C} & \cellcolor{myblue!13}\textcolor{black}{0.9883} & \cellcolor{myblue!28}\textcolor{black}{0.9893} \\
{Steadiness \& Cohesiveness} & \cellcolor{myblue!20}\textcolor{black}{0.3895} & \cellcolor{myblue!40}\textcolor{black}{0.4723} \\
{Silhouette Score} & \cellcolor{myblue!70}\textcolor{white}{\textbf{0.6779}} & \cellcolor{myblue!40}\textcolor{black}{0.3767} \\
{Davies Bouldin Index} & \cellcolor{myblue!70}\textcolor{white}{\textbf{0.4771}} & \cellcolor{myblue!40}\textcolor{black}{0.7933} \\
{Normalized Mutual Information} & \cellcolor{myblue!70}\textcolor{white}{\textbf{1.0000}} & \cellcolor{myblue!56}\textcolor{white}{0.5924} \\
{Adjusted Rand Index} & \cellcolor{myblue!70}\textcolor{white}{\textbf{1.0000}} & \cellcolor{myblue!46}\textcolor{white}{0.0638} \\
\bottomrule
\end{tabular}
\vspace{0.5em}
\end{table*}

\begin{table*}[ht!]
\centering
\caption{Embedding quality metrics for \textbf{matryoshka} dataset.}
\vspace{0.5em}
\begin{tabular}{lccc}
\toprule
\textbf{Metric} & \textbf{AngleEmbedding} & \textbf{MDS} & \textbf{PCA} \\
\midrule
 & \includegraphics[width=0.18\textwidth]{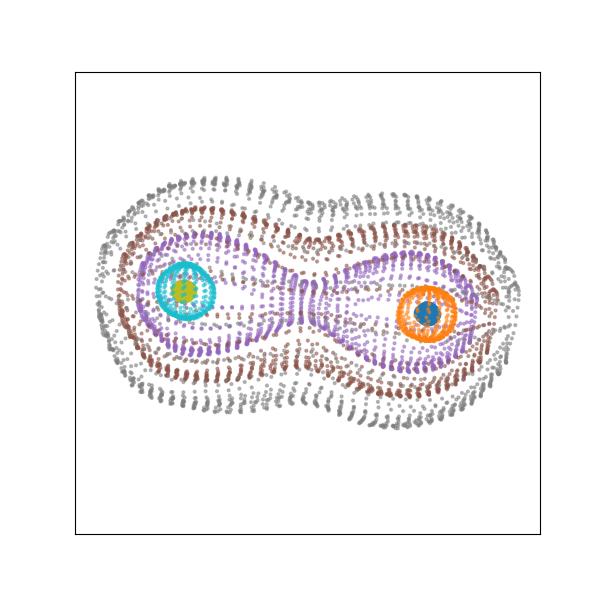} & \includegraphics[width=0.18\textwidth]{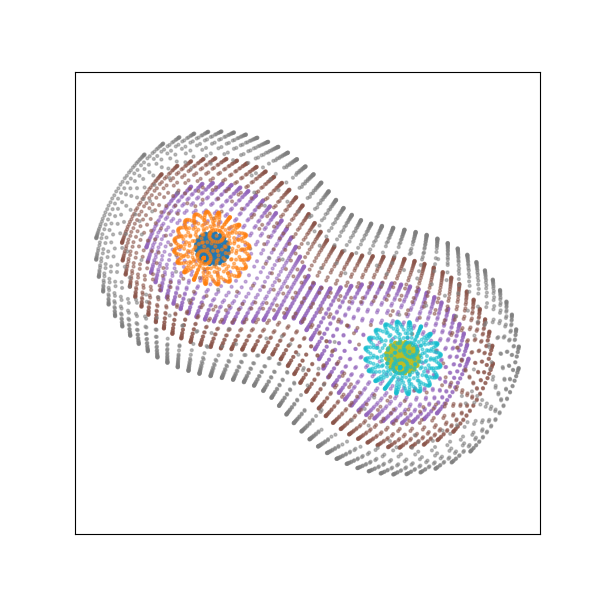} & \includegraphics[width=0.18\textwidth]{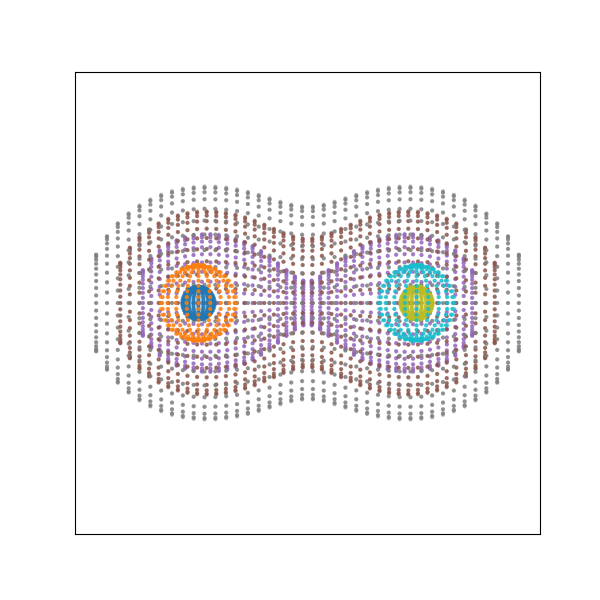} \\
\midrule
{\ourmetricFull} & \cellcolor{myblue!70}\textcolor{white}{\textbf{0.0130}} & \cellcolor{myblue!50}\textcolor{white}{0.0173} & \cellcolor{myblue!60}\textcolor{white}{0.0153} \\
{Cluster Distance Score} & \cellcolor{myblue!60}\textcolor{white}{0.0033} & \cellcolor{myblue!50}\textcolor{white}{0.0053} & \cellcolor{myblue!70}\textcolor{white}{\textbf{0.0000}} \\
{Label-T\&C} & \cellcolor{myblue!50}\textcolor{white}{0.9961} & \cellcolor{myblue!60}\textcolor{white}{0.9963} & \cellcolor{myblue!70}\textcolor{white}{\textbf{0.9984}} \\
{Steadiness \& Cohesiveness} & \cellcolor{myblue!50}\textcolor{white}{0.6725} & \cellcolor{myblue!40}\textcolor{black}{0.6676} & \cellcolor{myblue!30}\textcolor{black}{0.6250} \\
{Silhouette Score} & \cellcolor{myblue!10}\textcolor{black}{-0.3157} & \cellcolor{myblue!30}\textcolor{black}{-0.3101} & \cellcolor{myblue!0}\textcolor{black}{-0.3265} \\
{Davies Bouldin Index} & \cellcolor{myblue!30}\textcolor{black}{324.5935} & \cellcolor{myblue!10}\textcolor{black}{1381.0927} & \cellcolor{myblue!0}\textcolor{black}{23202969106.9720} \\
{Normalized Mutual Information} & \cellcolor{myblue!30}\textcolor{black}{0.3064} & \cellcolor{myblue!40}\textcolor{black}{0.3103} & \cellcolor{myblue!10}\textcolor{black}{0.2141} \\
{Adjusted Rand Index} & \cellcolor{myblue!0}\textcolor{black}{-0.0246} & \cellcolor{myblue!20}\textcolor{black}{-0.0048} & \cellcolor{myblue!10}\textcolor{black}{-0.0106} \\
\bottomrule
\end{tabular}
\vspace{0.5em}
\begin{tabular}{lccc}
\toprule
\textbf{Metric} & \textbf{PaCMap} & \textbf{Random} & \textbf{TSNE} \\
\midrule
 & \includegraphics[width=0.18\textwidth]{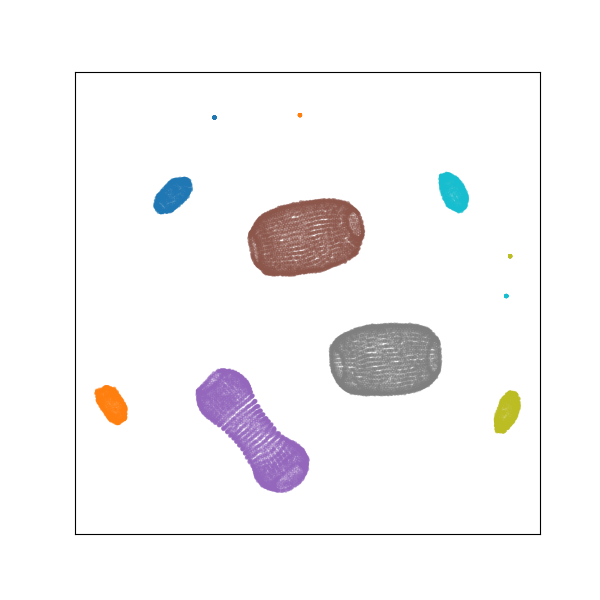} & \includegraphics[width=0.18\textwidth]{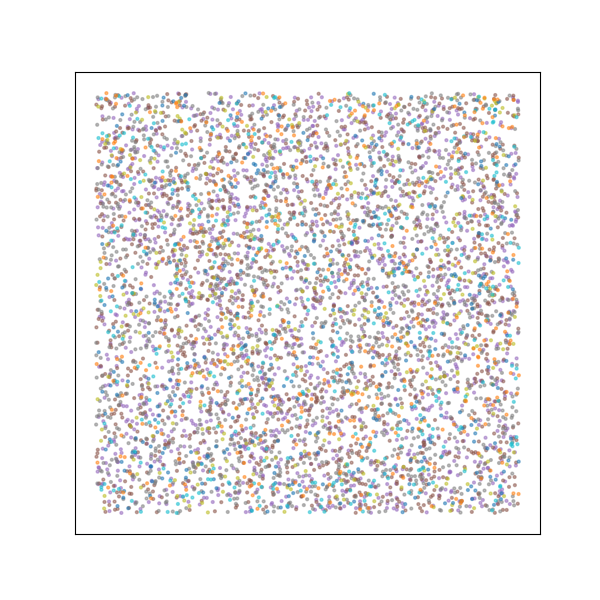} & \includegraphics[width=0.18\textwidth]{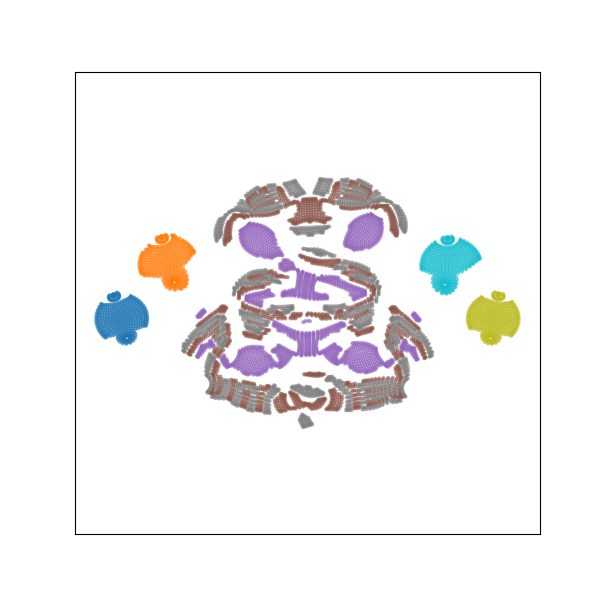} \\
\midrule
{\ourmetricFull} & \cellcolor{myblue!10}\textcolor{black}{0.0961} & \cellcolor{myblue!0}\textcolor{black}{0.1780} & \cellcolor{myblue!30}\textcolor{black}{0.0637} \\
{Cluster Distance Score} & \cellcolor{myblue!20}\textcolor{black}{0.4222} & \cellcolor{myblue!0}\textcolor{black}{0.6041} & \cellcolor{myblue!30}\textcolor{black}{0.1554} \\
{Label-T\&C} & \cellcolor{myblue!0}\textcolor{black}{0.8366} & \cellcolor{myblue!20}\textcolor{black}{0.8839} & \cellcolor{myblue!30}\textcolor{black}{0.9072} \\
{Steadiness \& Cohesiveness} & \cellcolor{myblue!10}\textcolor{black}{0.5663} & \cellcolor{myblue!0}\textcolor{black}{0.3572} & \cellcolor{myblue!60}\textcolor{white}{0.6787} \\
{Silhouette Score} & \cellcolor{myblue!70}\textcolor{white}{\textbf{0.6310}} & \cellcolor{myblue!40}\textcolor{black}{-0.0248} & \cellcolor{myblue!50}\textcolor{white}{0.0809} \\
{Davies Bouldin Index} & \cellcolor{myblue!70}\textcolor{white}{\textbf{0.4752}} & \cellcolor{myblue!40}\textcolor{black}{88.4908} & \cellcolor{myblue!60}\textcolor{white}{47.2568} \\
{Normalized Mutual Information} & \cellcolor{myblue!70}\textcolor{white}{\textbf{0.9859}} & \cellcolor{myblue!0}\textcolor{black}{0.0007} & \cellcolor{myblue!50}\textcolor{white}{0.6881} \\
{Adjusted Rand Index} & \cellcolor{myblue!70}\textcolor{white}{\textbf{0.9954}} & \cellcolor{myblue!30}\textcolor{black}{0.0001} & \cellcolor{myblue!50}\textcolor{white}{0.3135} \\
\bottomrule
\end{tabular}
\vspace{0.5em}
\begin{tabular}{lcc}
\toprule
\textbf{Metric} & \textbf{UMAP} & \textbf{UMATO} \\
\midrule
 & \includegraphics[width=0.18\textwidth]{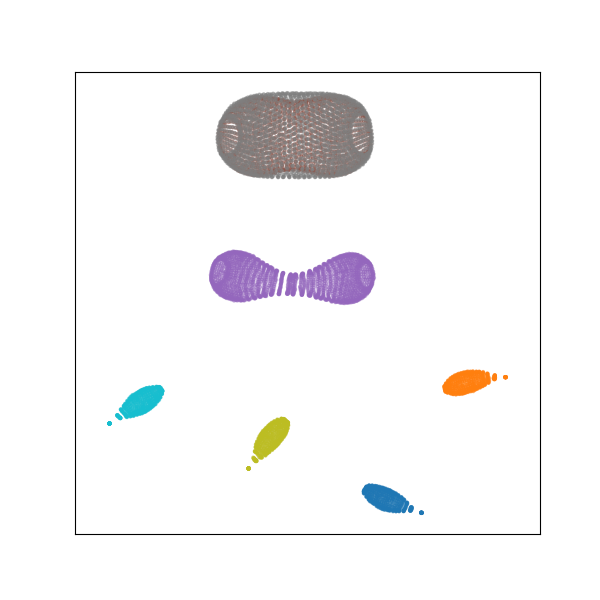} & \includegraphics[width=0.18\textwidth]{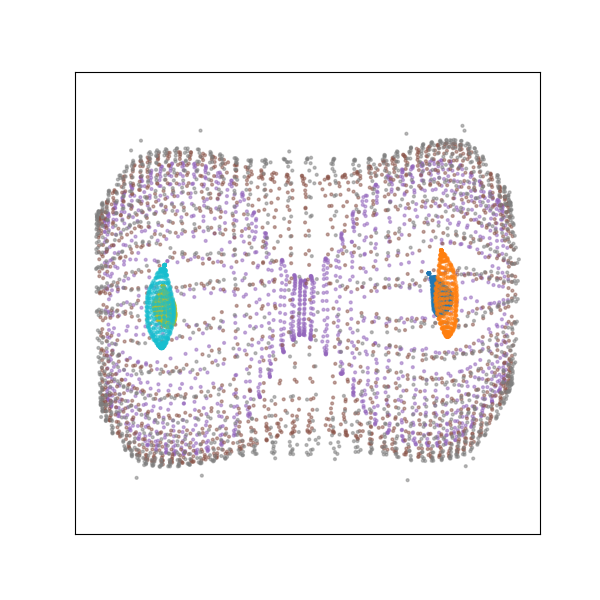} \\
\midrule
{\ourmetricFull} & \cellcolor{myblue!20}\textcolor{black}{0.0867} & \cellcolor{myblue!40}\textcolor{black}{0.0210} \\
{Cluster Distance Score} & \cellcolor{myblue!10}\textcolor{black}{0.5069} & \cellcolor{myblue!40}\textcolor{black}{0.0196} \\
{Label-T\&C} & \cellcolor{myblue!10}\textcolor{black}{0.8494} & \cellcolor{myblue!40}\textcolor{black}{0.9870} \\
{Steadiness \& Cohesiveness} & \cellcolor{myblue!20}\textcolor{black}{0.5671} & \cellcolor{myblue!70}\textcolor{white}{\textbf{0.6930}} \\
{Silhouette Score} & \cellcolor{myblue!60}\textcolor{white}{0.3523} & \cellcolor{myblue!20}\textcolor{black}{-0.3149} \\
{Davies Bouldin Index} & \cellcolor{myblue!50}\textcolor{white}{48.5437} & \cellcolor{myblue!20}\textcolor{black}{986.9179} \\
{Normalized Mutual Information} & \cellcolor{myblue!60}\textcolor{white}{0.8609} & \cellcolor{myblue!20}\textcolor{black}{0.2710} \\
{Adjusted Rand Index} & \cellcolor{myblue!60}\textcolor{white}{0.6771} & \cellcolor{myblue!40}\textcolor{black}{0.0839} \\
\bottomrule
\end{tabular}
\vspace{0.5em}
\end{table*}

\begin{table*}[ht!]
\centering
\caption{Embedding quality metrics for \textbf{donuts} dataset.}
\vspace{0.5em}
\begin{tabular}{lccc}
\toprule
\textbf{Metric} & \textbf{AngleEmbedding} & \textbf{MDS} & \textbf{PCA} \\
\midrule
 & \includegraphics[width=0.18\textwidth]{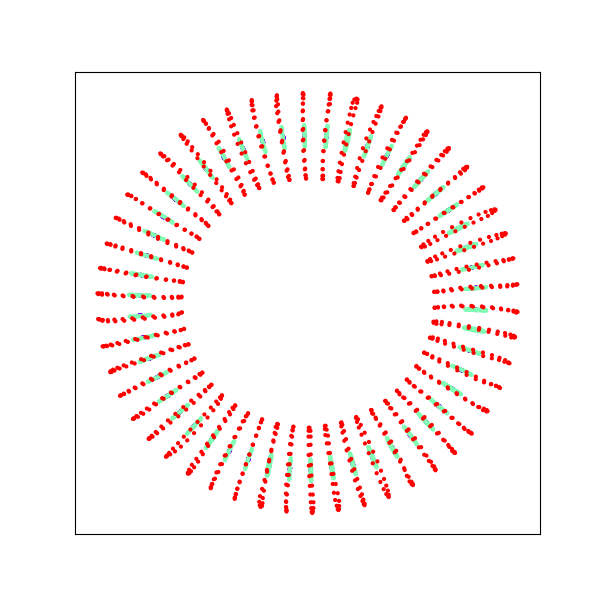} & \includegraphics[width=0.18\textwidth]{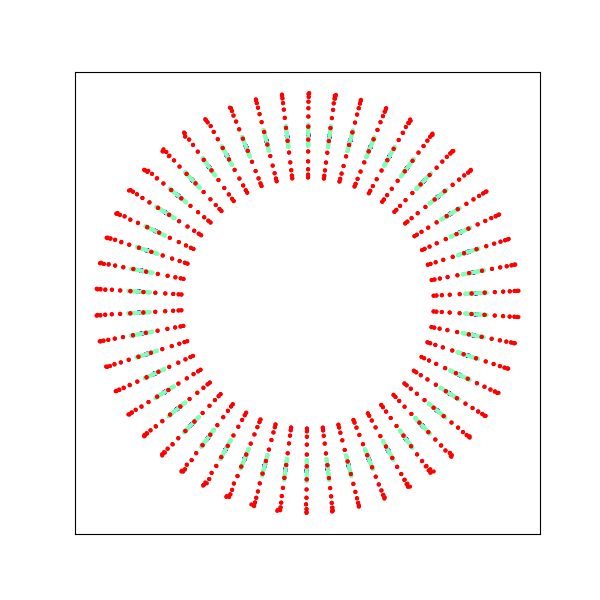} & \includegraphics[width=0.18\textwidth]{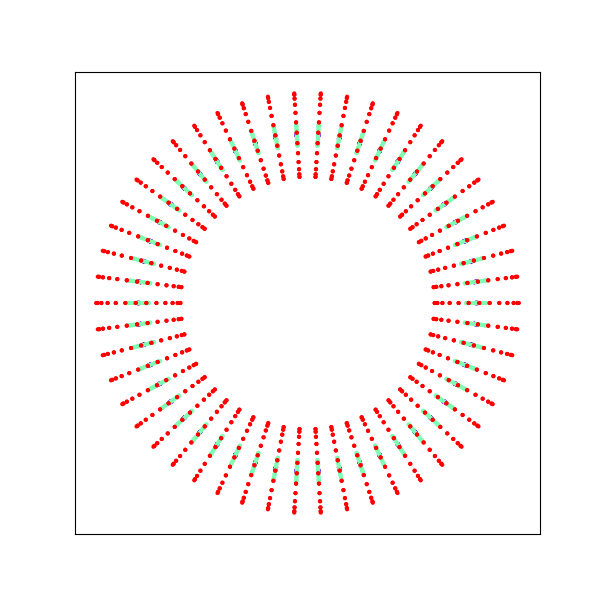} \\
\midrule
{\ourmetricFull} & \cellcolor{myblue!70}\textcolor{white}{0.0029} & \cellcolor{myblue!70}\textcolor{white}{\textbf{0.0029}} & \cellcolor{myblue!58}\textcolor{white}{0.0030} \\
{Cluster Distance Score} & \cellcolor{myblue!30}\textcolor{black}{0.7080} & \cellcolor{myblue!0}\textcolor{black}{0.7582} & \cellcolor{myblue!40}\textcolor{black}{0.5753} \\
{Label-T\&C} & \cellcolor{myblue!28}\textcolor{black}{0.9905} & \cellcolor{myblue!28}\textcolor{black}{0.9905} & \cellcolor{myblue!42}\textcolor{black}{0.9906} \\
{Steadiness \& Cohesiveness} & \cellcolor{myblue!70}\textcolor{white}{\textbf{0.8332}} & \cellcolor{myblue!60}\textcolor{white}{0.8296} & \cellcolor{myblue!50}\textcolor{white}{0.8237} \\
{Silhouette Score} & \cellcolor{myblue!11}\textcolor{black}{-0.0052} & \cellcolor{myblue!0}\textcolor{black}{-0.0057} & \cellcolor{myblue!35}\textcolor{black}{-0.0034} \\
{Davies Bouldin Index} & \cellcolor{myblue!20}\textcolor{black}{19787.4015} & \cellcolor{myblue!0}\textcolor{black}{5591361.4882} & \cellcolor{myblue!70}\textcolor{white}{\textbf{0.0000}} \\
{Normalized Mutual Information} & \cellcolor{myblue!50}\textcolor{white}{0.2504} & \cellcolor{myblue!40}\textcolor{black}{0.2388} & \cellcolor{myblue!20}\textcolor{black}{0.1930} \\
{Adjusted Rand Index} & \cellcolor{myblue!60}\textcolor{white}{0.0797} & \cellcolor{myblue!20}\textcolor{black}{0.0315} & \cellcolor{myblue!50}\textcolor{white}{0.0774} \\
\bottomrule
\end{tabular}
\vspace{0.5em}
\begin{tabular}{lccc}
\toprule
\textbf{Metric} & \textbf{PaCMap} & \textbf{Random} & \textbf{TSNE} \\
\midrule
 & \includegraphics[width=0.18\textwidth]{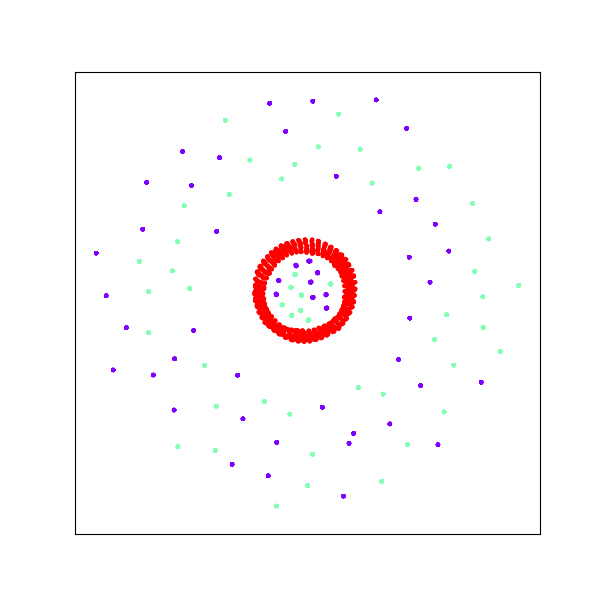} & \includegraphics[width=0.18\textwidth]{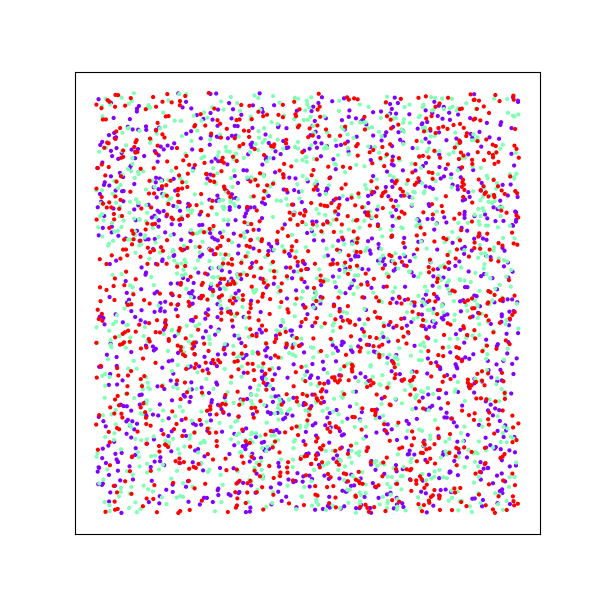} & \includegraphics[width=0.18\textwidth]{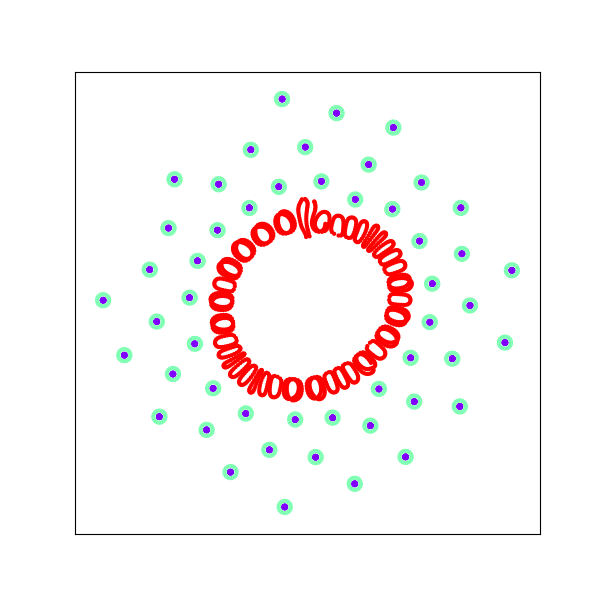} \\
\midrule
{\ourmetricFull} & \cellcolor{myblue!11}\textcolor{black}{0.1078} & \cellcolor{myblue!0}\textcolor{black}{0.1705} & \cellcolor{myblue!35}\textcolor{black}{0.0651} \\
{Cluster Distance Score} & \cellcolor{myblue!60}\textcolor{white}{0.3425} & \cellcolor{myblue!70}\textcolor{white}{\textbf{0.1088}} & \cellcolor{myblue!10}\textcolor{black}{0.7573} \\
{Label-T\&C} & \cellcolor{myblue!13}\textcolor{black}{0.9871} & \cellcolor{myblue!0}\textcolor{black}{0.9861} & \cellcolor{myblue!56}\textcolor{white}{0.9910} \\
{Steadiness \& Cohesiveness} & \cellcolor{myblue!10}\textcolor{black}{0.5069} & \cellcolor{myblue!0}\textcolor{black}{0.3396} & \cellcolor{myblue!30}\textcolor{black}{0.6768} \\
{Silhouette Score} & \cellcolor{myblue!58}\textcolor{white}{0.0211} & \cellcolor{myblue!23}\textcolor{black}{-0.0050} & \cellcolor{myblue!0}\textcolor{black}{-0.0057} \\
{Davies Bouldin Index} & \cellcolor{myblue!60}\textcolor{white}{12.5247} & \cellcolor{myblue!50}\textcolor{white}{142.4399} & \cellcolor{myblue!10}\textcolor{black}{44745.5822} \\
{Normalized Mutual Information} & \cellcolor{myblue!60}\textcolor{white}{0.3594} & \cellcolor{myblue!0}\textcolor{black}{0.0007} & \cellcolor{myblue!70}\textcolor{white}{\textbf{0.4554}} \\
{Adjusted Rand Index} & \cellcolor{myblue!40}\textcolor{black}{0.0390} & \cellcolor{myblue!0}\textcolor{black}{0.0000} & \cellcolor{myblue!70}\textcolor{white}{\textbf{0.4136}} \\
\bottomrule
\end{tabular}
\vspace{0.5em}
\begin{tabular}{lcc}
\toprule
\textbf{Metric} & \textbf{UMAP} & \textbf{UMATO} \\
\midrule
 & \includegraphics[width=0.18\textwidth]{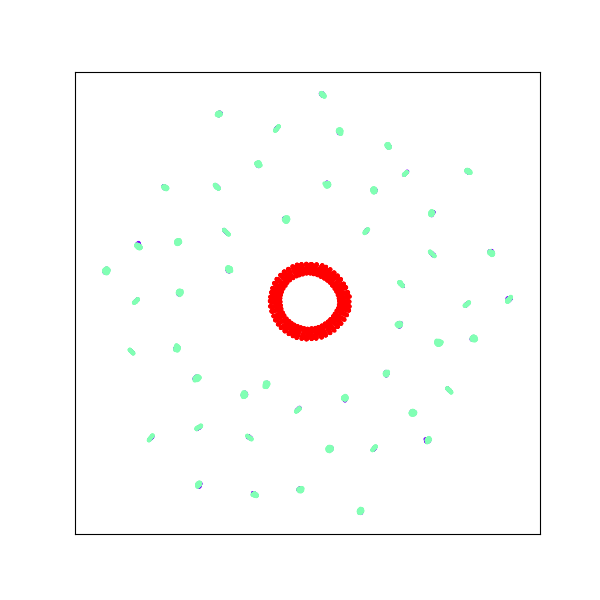} & \includegraphics[width=0.18\textwidth]{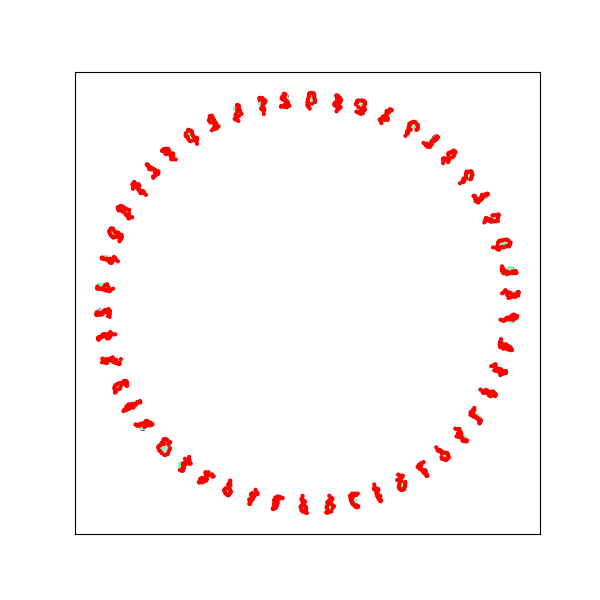} \\
\midrule
{\ourmetricFull} & \cellcolor{myblue!23}\textcolor{black}{0.1019} & \cellcolor{myblue!46}\textcolor{white}{0.0127} \\
{Cluster Distance Score} & \cellcolor{myblue!20}\textcolor{black}{0.7547} & \cellcolor{myblue!50}\textcolor{white}{0.3992} \\
{Label-T\&C} & \cellcolor{myblue!28}\textcolor{black}{0.9905} & \cellcolor{myblue!70}\textcolor{white}{\textbf{0.9914}} \\
{Steadiness \& Cohesiveness} & \cellcolor{myblue!20}\textcolor{black}{0.6290} & \cellcolor{myblue!40}\textcolor{black}{0.7379} \\
{Silhouette Score} & \cellcolor{myblue!70}\textcolor{white}{\textbf{0.0779}} & \cellcolor{myblue!46}\textcolor{white}{-0.0031} \\
{Davies Bouldin Index} & \cellcolor{myblue!40}\textcolor{black}{4675.0554} & \cellcolor{myblue!30}\textcolor{black}{19328.9877} \\
{Normalized Mutual Information} & \cellcolor{myblue!30}\textcolor{black}{0.2178} & \cellcolor{myblue!10}\textcolor{black}{0.0786} \\
{Adjusted Rand Index} & \cellcolor{myblue!10}\textcolor{black}{0.0154} & \cellcolor{myblue!30}\textcolor{black}{0.0334} \\
\bottomrule
\end{tabular}
\vspace{0.5em}
\end{table*}

\begin{table*}[ht!]
\centering
\caption{Embedding quality metrics for \textbf{concentric4} dataset.}
\vspace{0.5em}
\begin{tabular}{lccc}
\toprule
\textbf{Metric} & \textbf{AngleEmbedding} & \textbf{MDS} & \textbf{PCA} \\
\midrule
 & \includegraphics[width=0.18\textwidth]{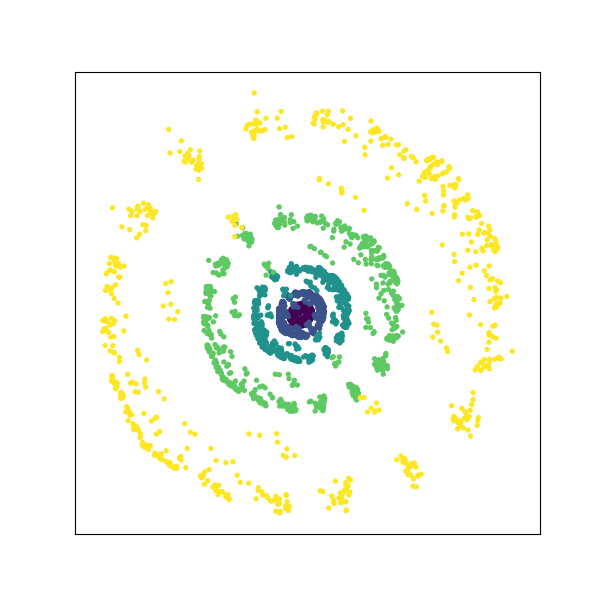} & \includegraphics[width=0.18\textwidth]{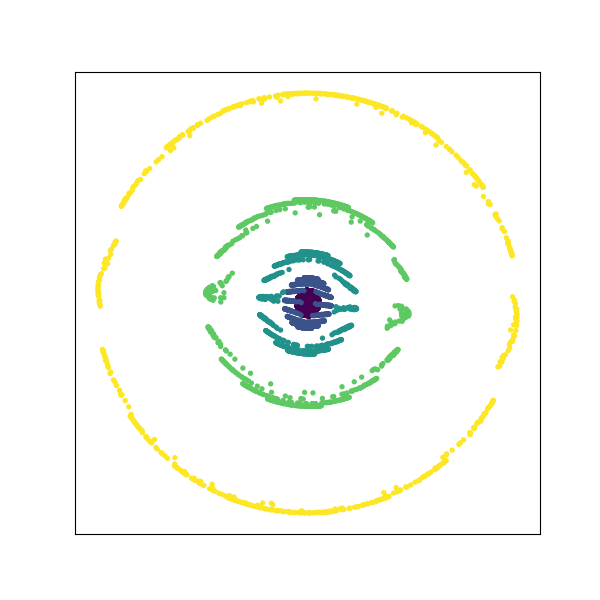} & \includegraphics[width=0.18\textwidth]{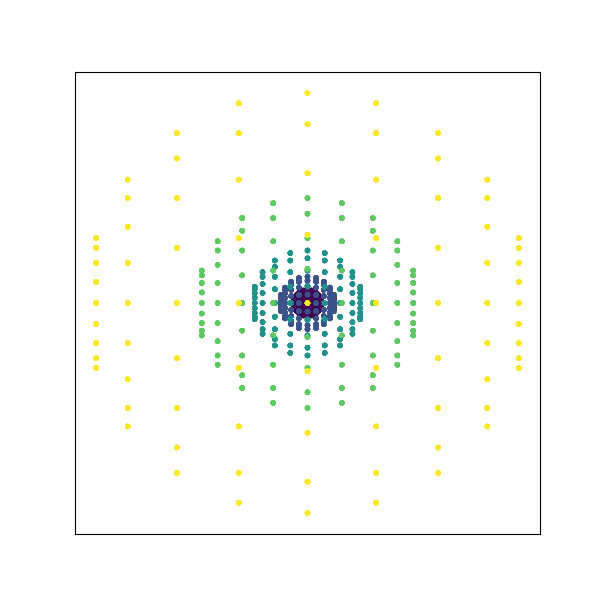} \\
\midrule
{\ourmetricFull} & \cellcolor{myblue!70}\textcolor{white}{\textbf{0.0197}} & \cellcolor{myblue!50}\textcolor{white}{0.0267} & \cellcolor{myblue!60}\textcolor{white}{0.0248} \\
{Cluster Distance Score} & \cellcolor{myblue!50}\textcolor{white}{0.3582} & \cellcolor{myblue!20}\textcolor{black}{0.5800} & \cellcolor{myblue!70}\textcolor{white}{\textbf{0.0020}} \\
{Label-T\&C} & \cellcolor{myblue!40}\textcolor{black}{0.9527} & \cellcolor{myblue!50}\textcolor{white}{0.9534} & \cellcolor{myblue!70}\textcolor{white}{\textbf{0.9958}} \\
{Steadiness \& Cohesiveness} & \cellcolor{myblue!50}\textcolor{white}{0.7193} & \cellcolor{myblue!40}\textcolor{black}{0.7187} & \cellcolor{myblue!60}\textcolor{white}{0.7227} \\
{Silhouette Score} & \cellcolor{myblue!20}\textcolor{black}{-0.0796} & \cellcolor{myblue!30}\textcolor{black}{-0.0746} & \cellcolor{myblue!0}\textcolor{black}{-0.1080} \\
{Davies Bouldin Index} & \cellcolor{myblue!20}\textcolor{black}{371.6842} & \cellcolor{myblue!0}\textcolor{black}{1063.4143} & \cellcolor{myblue!70}\textcolor{white}{\textbf{0.0000}} \\
{Normalized Mutual Information} & \cellcolor{myblue!40}\textcolor{black}{0.4218} & \cellcolor{myblue!60}\textcolor{white}{0.4730} & \cellcolor{myblue!30}\textcolor{black}{0.4187} \\
{Adjusted Rand Index} & \cellcolor{myblue!35}\textcolor{black}{0.0380} & \cellcolor{myblue!46}\textcolor{white}{0.0730} & \cellcolor{myblue!11}\textcolor{black}{0.0170} \\
\bottomrule
\end{tabular}
\vspace{0.5em}
\begin{tabular}{lccc}
\toprule
\textbf{Metric} & \textbf{PaCMap} & \textbf{Random} & \textbf{TSNE} \\
\midrule
 & \includegraphics[width=0.18\textwidth]{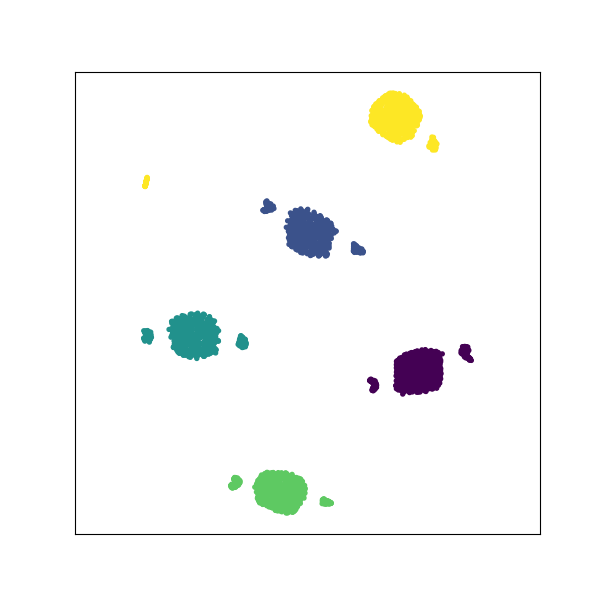} & \includegraphics[width=0.18\textwidth]{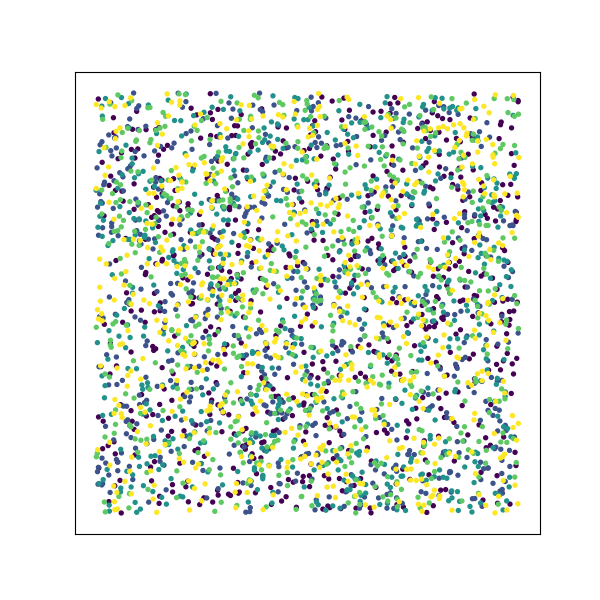} & \includegraphics[width=0.18\textwidth]{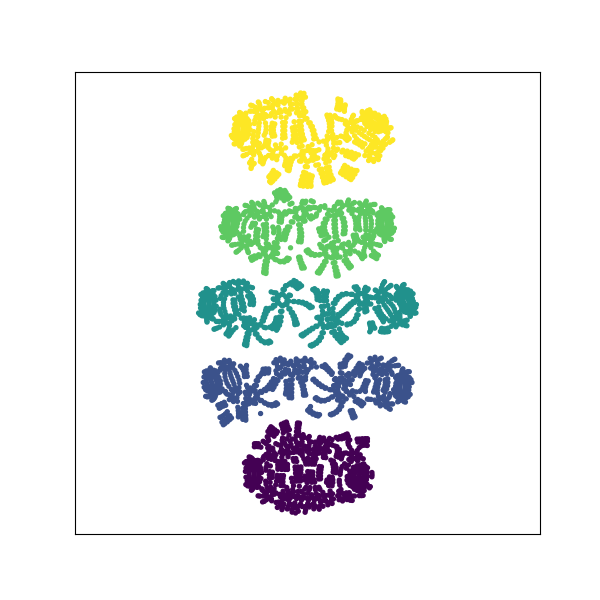} \\
\midrule
{\ourmetricFull} & \cellcolor{myblue!20}\textcolor{black}{0.1379} & \cellcolor{myblue!0}\textcolor{black}{0.1830} & \cellcolor{myblue!30}\textcolor{black}{0.1202} \\
{Cluster Distance Score} & \cellcolor{myblue!30}\textcolor{black}{0.5631} & \cellcolor{myblue!0}\textcolor{black}{0.7549} & \cellcolor{myblue!60}\textcolor{white}{0.3345} \\
{Label-T\&C} & \cellcolor{myblue!20}\textcolor{black}{0.6001} & \cellcolor{myblue!30}\textcolor{black}{0.9475} & \cellcolor{myblue!10}\textcolor{black}{0.5915} \\
{Steadiness \& Cohesiveness} & \cellcolor{myblue!10}\textcolor{black}{0.6510} & \cellcolor{myblue!0}\textcolor{black}{0.4049} & \cellcolor{myblue!70}\textcolor{white}{\textbf{0.7345}} \\
{Silhouette Score} & \cellcolor{myblue!70}\textcolor{white}{\textbf{0.7399}} & \cellcolor{myblue!40}\textcolor{black}{-0.0225} & \cellcolor{myblue!50}\textcolor{white}{0.3390} \\
{Davies Bouldin Index} & \cellcolor{myblue!60}\textcolor{white}{0.4346} & \cellcolor{myblue!30}\textcolor{black}{51.3994} & \cellcolor{myblue!40}\textcolor{black}{1.3893} \\
{Normalized Mutual Information} & \cellcolor{myblue!70}\textcolor{white}{\textbf{0.8048}} & \cellcolor{myblue!10}\textcolor{black}{0.0012} & \cellcolor{myblue!0}\textcolor{black}{0.0000} \\
{Adjusted Rand Index} & \cellcolor{myblue!70}\textcolor{white}{\textbf{0.7266}} & \cellcolor{myblue!0}\textcolor{black}{0.0000} & \cellcolor{myblue!0}\textcolor{black}{0.0000} \\
\bottomrule
\end{tabular}
\vspace{0.5em}
\begin{tabular}{lcc}
\toprule
\textbf{Metric} & \textbf{UMAP} & \textbf{UMATO} \\
\midrule
 & \includegraphics[width=0.18\textwidth]{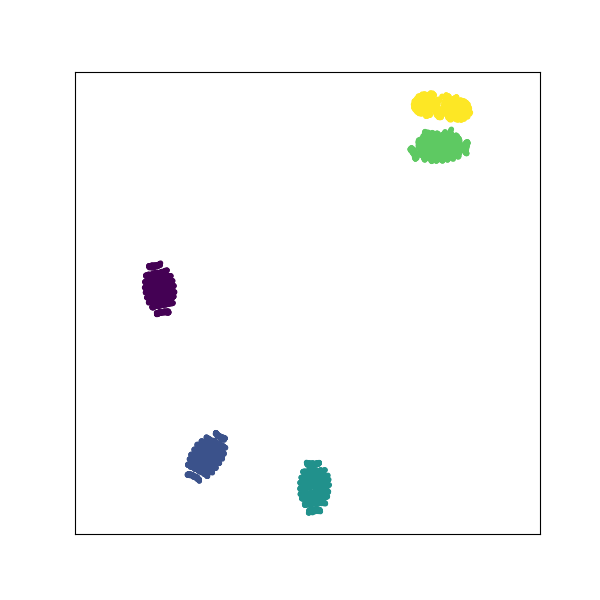} & \includegraphics[width=0.18\textwidth]{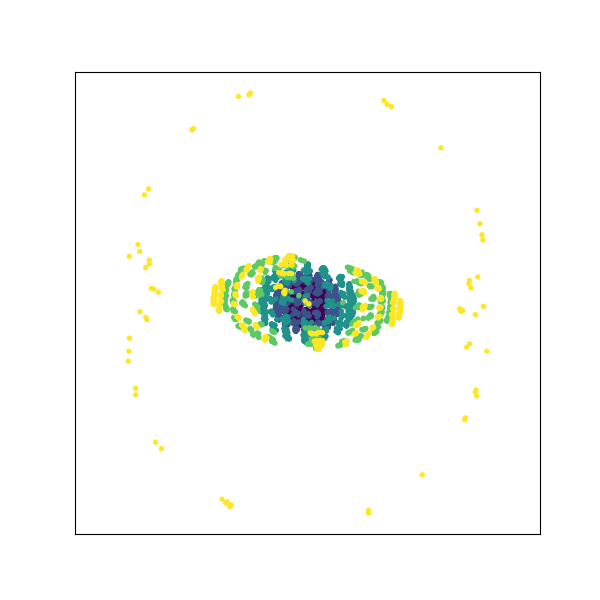} \\
\midrule
{\ourmetricFull} & \cellcolor{myblue!10}\textcolor{black}{0.1447} & \cellcolor{myblue!40}\textcolor{black}{0.0408} \\
{Cluster Distance Score} & \cellcolor{myblue!40}\textcolor{black}{0.4795} & \cellcolor{myblue!10}\textcolor{black}{0.7436} \\
{Label-T\&C} & \cellcolor{myblue!0}\textcolor{black}{0.5891} & \cellcolor{myblue!60}\textcolor{white}{0.9556} \\
{Steadiness \& Cohesiveness} & \cellcolor{myblue!30}\textcolor{black}{0.6974} & \cellcolor{myblue!20}\textcolor{black}{0.6641} \\
{Silhouette Score} & \cellcolor{myblue!60}\textcolor{white}{0.6988} & \cellcolor{myblue!10}\textcolor{black}{-0.0996} \\
{Davies Bouldin Index} & \cellcolor{myblue!50}\textcolor{white}{0.5025} & \cellcolor{myblue!10}\textcolor{black}{704.2616} \\
{Normalized Mutual Information} & \cellcolor{myblue!50}\textcolor{white}{0.4696} & \cellcolor{myblue!20}\textcolor{black}{0.3506} \\
{Adjusted Rand Index} & \cellcolor{myblue!58}\textcolor{white}{0.1214} & \cellcolor{myblue!23}\textcolor{black}{0.0219} \\
\bottomrule
\end{tabular}
\vspace{0.5em}
\end{table*}

\begin{table*}[ht!]
\centering
\caption{Embedding quality metrics for \textbf{concentric3} dataset.}
\vspace{0.5em}
\begin{tabular}{lccc}
\toprule
\textbf{Metric} & \textbf{AngleEmbedding} & \textbf{MDS} & \textbf{PCA} \\
\midrule
 & \includegraphics[width=0.18\textwidth]{images/emb_figs/concentric3/AngleEmbedding.png} & \includegraphics[width=0.18\textwidth]{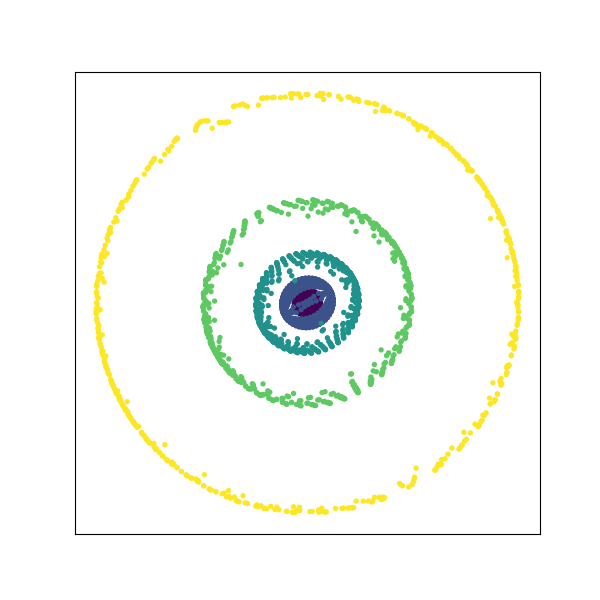} & \includegraphics[width=0.18\textwidth]{images/emb_figs/concentric3/PCA.png} \\
\midrule
{\ourmetricFull} & \cellcolor{myblue!70}\textcolor{white}{\textbf{0.0200}} & \cellcolor{myblue!60}\textcolor{white}{0.0230} & \cellcolor{myblue!50}\textcolor{white}{0.0236} \\
{Cluster Distance Score} & \cellcolor{myblue!70}\textcolor{white}{\textbf{0.0474}} & \cellcolor{myblue!40}\textcolor{black}{0.3508} & \cellcolor{myblue!60}\textcolor{white}{0.0588} \\
{Label-T\&C} & \cellcolor{myblue!13}\textcolor{black}{0.9627} & \cellcolor{myblue!42}\textcolor{black}{0.9660} & \cellcolor{myblue!28}\textcolor{black}{0.9646} \\
{Steadiness \& Cohesiveness} & \cellcolor{myblue!40}\textcolor{black}{0.6811} & \cellcolor{myblue!20}\textcolor{black}{0.6607} & \cellcolor{myblue!30}\textcolor{black}{0.6633} \\
{Silhouette Score} & \cellcolor{myblue!20}\textcolor{black}{-0.0796} & \cellcolor{myblue!30}\textcolor{black}{-0.0771} & \cellcolor{myblue!10}\textcolor{black}{-0.1045} \\
{Davies Bouldin Index} & \cellcolor{myblue!20}\textcolor{black}{150.8393} & \cellcolor{myblue!0}\textcolor{black}{4622.5487} & \cellcolor{myblue!70}\textcolor{white}{\textbf{0.0000}} \\
{Normalized Mutual Information} & \cellcolor{myblue!35}\textcolor{black}{0.6018} & \cellcolor{myblue!46}\textcolor{white}{0.6258} & \cellcolor{myblue!23}\textcolor{black}{0.5072} \\
{Adjusted Rand Index} & \cellcolor{myblue!46}\textcolor{white}{0.5158} & \cellcolor{myblue!35}\textcolor{black}{0.4743} & \cellcolor{myblue!23}\textcolor{black}{0.3391} \\
\bottomrule
\end{tabular}
\vspace{0.5em}
\begin{tabular}{lccc}
\toprule
\textbf{Metric} & \textbf{PaCMap} & \textbf{Random} & \textbf{TSNE} \\
\midrule
 & \includegraphics[width=0.18\textwidth]{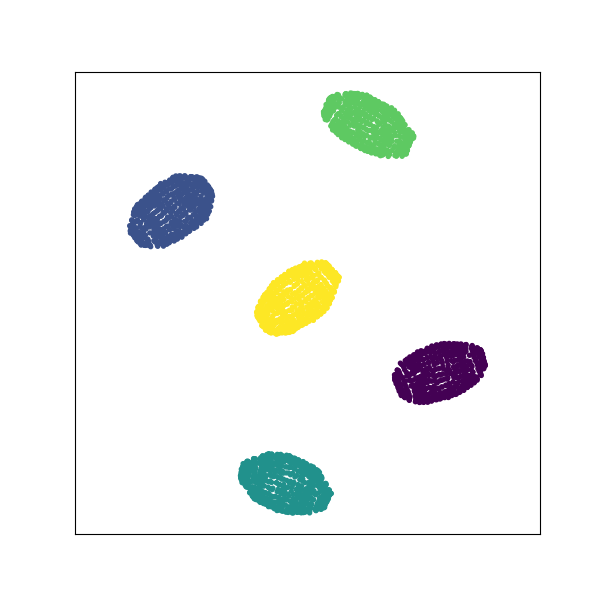} & \includegraphics[width=0.18\textwidth]{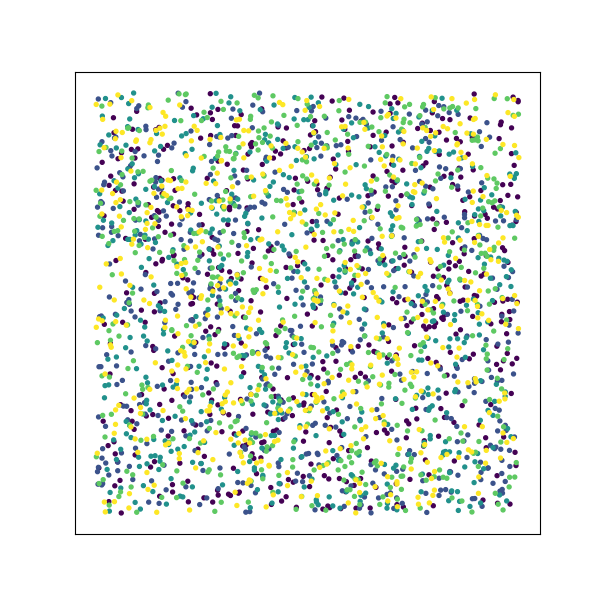} & \includegraphics[width=0.18\textwidth]{images/emb_figs/concentric3/TSNE.png} \\
\midrule
{\ourmetricFull} & \cellcolor{myblue!20}\textcolor{black}{0.1555} & \cellcolor{myblue!0}\textcolor{black}{0.1915} & \cellcolor{myblue!30}\textcolor{black}{0.1365} \\
{Cluster Distance Score} & \cellcolor{myblue!10}\textcolor{black}{0.7296} & \cellcolor{myblue!0}\textcolor{black}{0.7498} & \cellcolor{myblue!50}\textcolor{white}{0.3337} \\
{Label-T\&C} & \cellcolor{myblue!0}\textcolor{black}{0.6849} & \cellcolor{myblue!70}\textcolor{white}{\textbf{0.9674}} & \cellcolor{myblue!0}\textcolor{black}{0.6849} \\
{Steadiness \& Cohesiveness} & \cellcolor{myblue!60}\textcolor{white}{0.7174} & \cellcolor{myblue!0}\textcolor{black}{0.3513} & \cellcolor{myblue!70}\textcolor{white}{\textbf{0.7485}} \\
{Silhouette Score} & \cellcolor{myblue!60}\textcolor{white}{0.7624} & \cellcolor{myblue!40}\textcolor{black}{-0.0203} & \cellcolor{myblue!50}\textcolor{white}{0.4344} \\
{Davies Bouldin Index} & \cellcolor{myblue!50}\textcolor{white}{0.3501} & \cellcolor{myblue!30}\textcolor{black}{61.6170} & \cellcolor{myblue!40}\textcolor{black}{1.0719} \\
{Normalized Mutual Information} & \cellcolor{myblue!70}\textcolor{white}{\textbf{1.0000}} & \cellcolor{myblue!0}\textcolor{black}{0.0013} & \cellcolor{myblue!58}\textcolor{white}{0.9588} \\
{Adjusted Rand Index} & \cellcolor{myblue!70}\textcolor{white}{\textbf{1.0000}} & \cellcolor{myblue!0}\textcolor{black}{0.0000} & \cellcolor{myblue!58}\textcolor{white}{0.9353} \\
\bottomrule
\end{tabular}
\vspace{0.5em}
\begin{tabular}{lcc}
\toprule
\textbf{Metric} & \textbf{UMAP} & \textbf{UMATO} \\
\midrule
 & \includegraphics[width=0.18\textwidth]{images/emb_figs/concentric3/UMAP.png} & \includegraphics[width=0.18\textwidth]{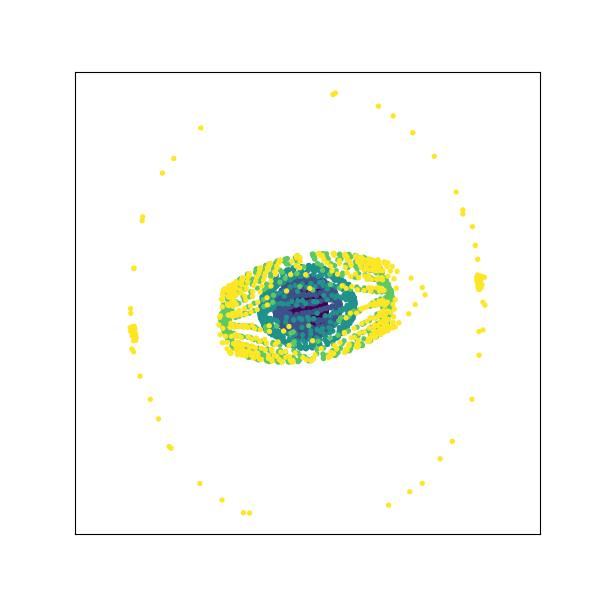} \\
\midrule
{\ourmetricFull} & \cellcolor{myblue!10}\textcolor{black}{0.1566} & \cellcolor{myblue!40}\textcolor{black}{0.0369} \\
{Cluster Distance Score} & \cellcolor{myblue!30}\textcolor{black}{0.4727} & \cellcolor{myblue!20}\textcolor{black}{0.6699} \\
{Label-T\&C} & \cellcolor{myblue!0}\textcolor{black}{0.6849} & \cellcolor{myblue!56}\textcolor{white}{0.9666} \\
{Steadiness \& Cohesiveness} & \cellcolor{myblue!50}\textcolor{white}{0.7097} & \cellcolor{myblue!10}\textcolor{black}{0.6256} \\
{Silhouette Score} & \cellcolor{myblue!70}\textcolor{white}{\textbf{0.8344}} & \cellcolor{myblue!0}\textcolor{black}{-0.1095} \\
{Davies Bouldin Index} & \cellcolor{myblue!60}\textcolor{white}{0.2451} & \cellcolor{myblue!10}\textcolor{black}{403.5866} \\
{Normalized Mutual Information} & \cellcolor{myblue!70}\textcolor{white}{\textbf{1.0000}} & \cellcolor{myblue!11}\textcolor{black}{0.0646} \\
{Adjusted Rand Index} & \cellcolor{myblue!70}\textcolor{white}{\textbf{1.0000}} & \cellcolor{myblue!11}\textcolor{black}{0.0034} \\
\bottomrule
\end{tabular}
\vspace{0.5em}
\end{table*}

\begin{table*}[ht!]
\centering
\caption{Embedding quality metrics for \textbf{usps} dataset.}
\vspace{0.5em}
\begin{tabular}{lccc}
\toprule
\textbf{Metric} & \textbf{AngleEmbedding} & \textbf{MDS} & \textbf{PCA} \\
\midrule
 & \includegraphics[width=0.18\textwidth]{images/emb_figs/usps/AngleEmbedding.png} & \includegraphics[width=0.18\textwidth]{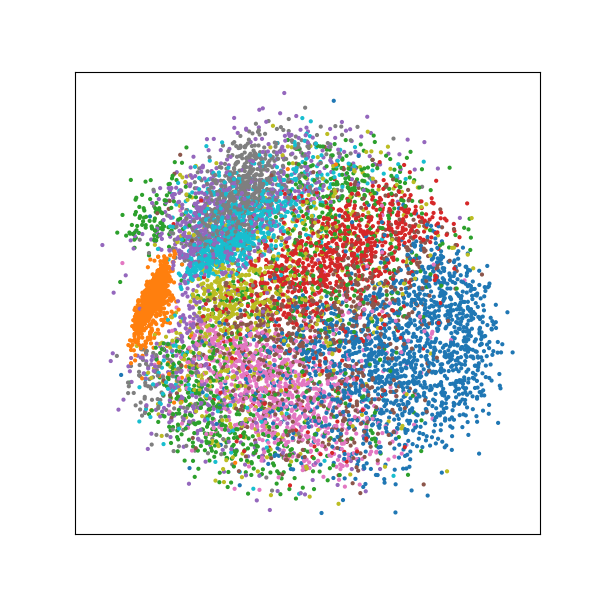} & \includegraphics[width=0.18\textwidth]{images/emb_figs/usps/PCA.png} \\
\midrule
{\ourmetricFull} & \cellcolor{myblue!70}\textcolor{white}{\textbf{0.0201}} & \cellcolor{myblue!10}\textcolor{black}{0.0505} & \cellcolor{myblue!20}\textcolor{black}{0.0367} \\
{Cluster Distance Score} & \cellcolor{myblue!10}\textcolor{black}{0.4257} & \cellcolor{myblue!40}\textcolor{black}{0.2479} & \cellcolor{myblue!30}\textcolor{black}{0.2640} \\
{Label-T\&C} & \cellcolor{myblue!70}\textcolor{white}{\textbf{0.9895}} & \cellcolor{myblue!10}\textcolor{black}{0.9308} & \cellcolor{myblue!20}\textcolor{black}{0.9525} \\
{Steadiness \& Cohesiveness} & \cellcolor{myblue!30}\textcolor{black}{0.7240} & \cellcolor{myblue!10}\textcolor{black}{0.5908} & \cellcolor{myblue!20}\textcolor{black}{0.5970} \\
{Silhouette Score} & \cellcolor{myblue!30}\textcolor{black}{0.2211} & \cellcolor{myblue!10}\textcolor{black}{0.0290} & \cellcolor{myblue!20}\textcolor{black}{0.0950} \\
{Davies Bouldin Index} & \cellcolor{myblue!40}\textcolor{black}{1.2654} & \cellcolor{myblue!10}\textcolor{black}{6.1510} & \cellcolor{myblue!20}\textcolor{black}{2.6720} \\
{Normalized Mutual Information} & \cellcolor{myblue!30}\textcolor{black}{0.4627} & \cellcolor{myblue!10}\textcolor{black}{0.0049} & \cellcolor{myblue!20}\textcolor{black}{0.0104} \\
{Adjusted Rand Index} & \cellcolor{myblue!30}\textcolor{black}{0.1372} & \cellcolor{myblue!20}\textcolor{black}{0.0000} & \cellcolor{myblue!0}\textcolor{black}{-0.0011} \\
\bottomrule
\end{tabular}
\vspace{0.5em}
\begin{tabular}{lccc}
\toprule
\textbf{Metric} & \textbf{PaCMap} & \textbf{Random} & \textbf{TSNE} \\
\midrule
 & \includegraphics[width=0.18\textwidth]{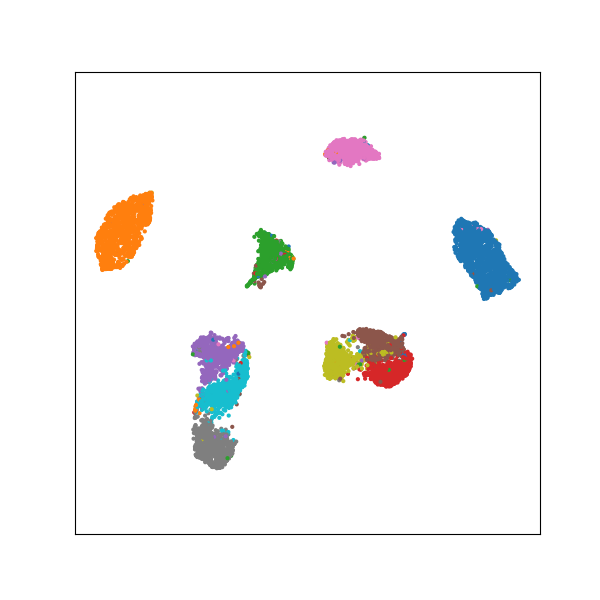} & \includegraphics[width=0.18\textwidth]{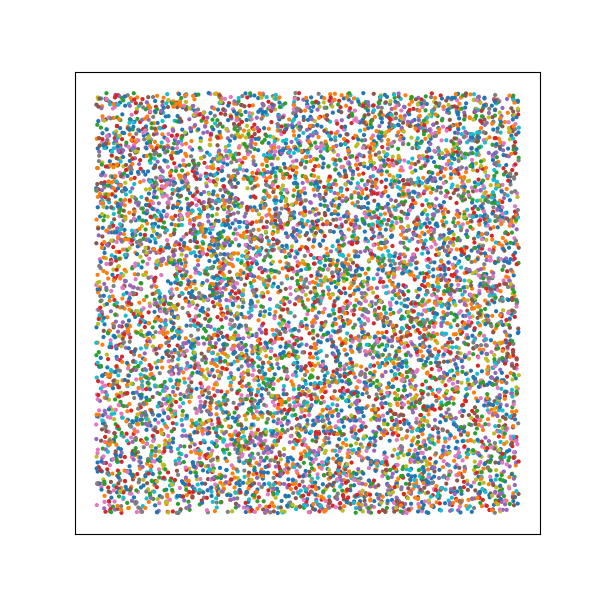} & \includegraphics[width=0.18\textwidth]{images/emb_figs/usps/TSNE.png} \\
\midrule
{\ourmetricFull} & \cellcolor{myblue!50}\textcolor{white}{0.0300} & \cellcolor{myblue!0}\textcolor{black}{0.1235} & \cellcolor{myblue!60}\textcolor{white}{0.0269} \\
{Cluster Distance Score} & \cellcolor{myblue!60}\textcolor{white}{0.2341} & \cellcolor{myblue!0}\textcolor{black}{0.5126} & \cellcolor{myblue!70}\textcolor{white}{\textbf{0.2272}} \\
{Label-T\&C} & \cellcolor{myblue!60}\textcolor{white}{0.9886} & \cellcolor{myblue!0}\textcolor{black}{0.7098} & \cellcolor{myblue!50}\textcolor{white}{0.9878} \\
{Steadiness \& Cohesiveness} & \cellcolor{myblue!70}\textcolor{white}{\textbf{0.8432}} & \cellcolor{myblue!0}\textcolor{black}{0.3804} & \cellcolor{myblue!60}\textcolor{white}{0.8413} \\
{Silhouette Score} & \cellcolor{myblue!70}\textcolor{white}{\textbf{0.5884}} & \cellcolor{myblue!0}\textcolor{black}{-0.0182} & \cellcolor{myblue!50}\textcolor{white}{0.4586} \\
{Davies Bouldin Index} & \cellcolor{myblue!60}\textcolor{white}{0.6649} & \cellcolor{myblue!0}\textcolor{black}{152.1682} & \cellcolor{myblue!50}\textcolor{white}{0.7907} \\
{Normalized Mutual Information} & \cellcolor{myblue!60}\textcolor{white}{0.8022} & \cellcolor{myblue!0}\textcolor{black}{0.0002} & \cellcolor{myblue!70}\textcolor{white}{\textbf{0.8908}} \\
{Adjusted Rand Index} & \cellcolor{myblue!50}\textcolor{white}{0.6373} & \cellcolor{myblue!10}\textcolor{black}{-0.0001} & \cellcolor{myblue!70}\textcolor{white}{\textbf{0.8753}} \\
\bottomrule
\end{tabular}
\vspace{0.5em}
\begin{tabular}{lcc}
\toprule
\textbf{Metric} & \textbf{UMAP} & \textbf{UMATO} \\
\midrule
 & \includegraphics[width=0.18\textwidth]{images/emb_figs/usps/UMAP.png} & \includegraphics[width=0.18\textwidth]{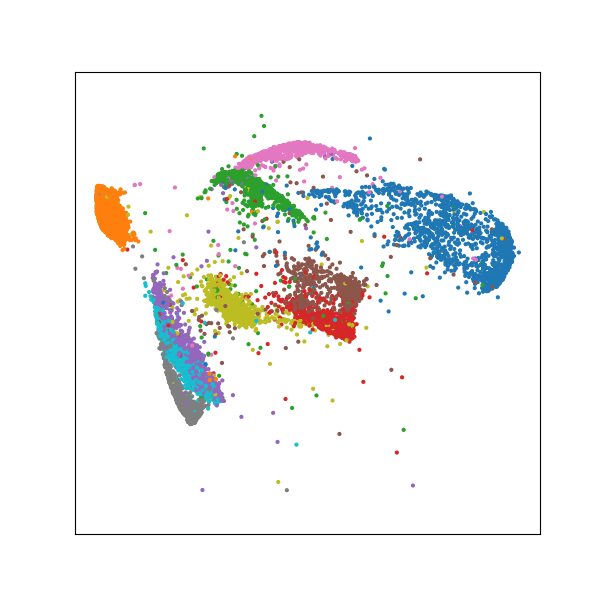} \\
\midrule
{\ourmetricFull} & \cellcolor{myblue!30}\textcolor{black}{0.0309} & \cellcolor{myblue!40}\textcolor{black}{0.0301} \\
{Cluster Distance Score} & \cellcolor{myblue!20}\textcolor{black}{0.3090} & \cellcolor{myblue!50}\textcolor{white}{0.2395} \\
{Label-T\&C} & \cellcolor{myblue!40}\textcolor{black}{0.9869} & \cellcolor{myblue!30}\textcolor{black}{0.9862} \\
{Steadiness \& Cohesiveness} & \cellcolor{myblue!50}\textcolor{white}{0.8411} & \cellcolor{myblue!40}\textcolor{black}{0.8043} \\
{Silhouette Score} & \cellcolor{myblue!60}\textcolor{white}{0.5798} & \cellcolor{myblue!40}\textcolor{black}{0.3215} \\
{Davies Bouldin Index} & \cellcolor{myblue!70}\textcolor{white}{\textbf{0.6543}} & \cellcolor{myblue!30}\textcolor{black}{1.9863} \\
{Normalized Mutual Information} & \cellcolor{myblue!50}\textcolor{white}{0.8014} & \cellcolor{myblue!40}\textcolor{black}{0.6496} \\
{Adjusted Rand Index} & \cellcolor{myblue!60}\textcolor{white}{0.6416} & \cellcolor{myblue!40}\textcolor{black}{0.4719} \\
\bottomrule
\end{tabular}
\vspace{0.5em}
\label{tab:usps-full}
\end{table*}

\begin{table*}[ht!]
\centering
\caption{Embedding quality metrics for \textbf{trec} dataset.}
\vspace{0.5em}
\begin{tabular}{lccc}
\toprule
\textbf{Metric} & \textbf{AngleEmbedding} & \textbf{MDS} & \textbf{PCA} \\
\midrule
 & \includegraphics[width=0.18\textwidth]{images/emb_figs/trec/AngleEmbedding.png} & \includegraphics[width=0.18\textwidth]{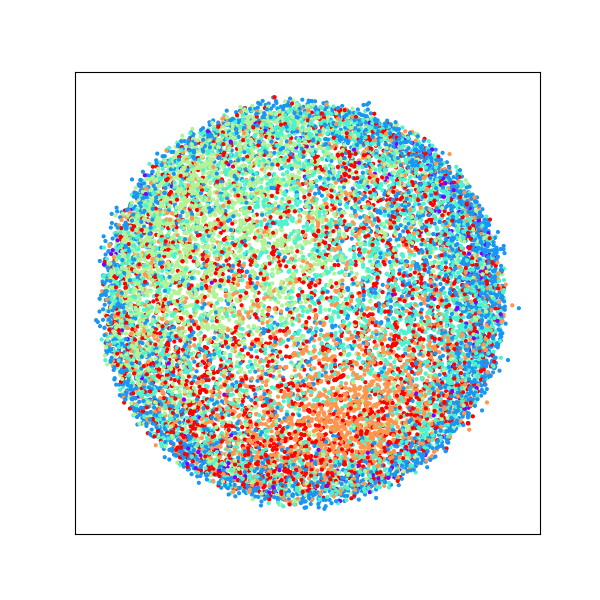} & \includegraphics[width=0.18\textwidth]{images/emb_figs/trec/PCA.png} \\
\midrule
{\ourmetricFull} & \cellcolor{myblue!70}\textcolor{white}{\textbf{0.0372}} & \cellcolor{myblue!20}\textcolor{black}{0.0884} & \cellcolor{myblue!60}\textcolor{white}{0.0654} \\
{Cluster Distance Score} & \cellcolor{myblue!70}\textcolor{white}{\textbf{0.2785}} & \cellcolor{myblue!0}\textcolor{black}{0.5340} & \cellcolor{myblue!50}\textcolor{white}{0.3160} \\
{Label-T\&C} & \cellcolor{myblue!70}\textcolor{white}{\textbf{0.9654}} & \cellcolor{myblue!20}\textcolor{black}{0.8217} & \cellcolor{myblue!60}\textcolor{white}{0.9247} \\
{Steadiness \& Cohesiveness} & \cellcolor{myblue!0}\textcolor{black}{0.4659} & \cellcolor{myblue!20}\textcolor{black}{0.5083} & \cellcolor{myblue!30}\textcolor{black}{0.5385} \\
{Silhouette Score} & \cellcolor{myblue!70}\textcolor{white}{\textbf{0.1905}} & \cellcolor{myblue!40}\textcolor{black}{-0.0739} & \cellcolor{myblue!30}\textcolor{black}{-0.1032} \\
{Davies Bouldin Index} & \cellcolor{myblue!70}\textcolor{white}{\textbf{1.3990}} & \cellcolor{myblue!20}\textcolor{black}{22.4497} & \cellcolor{myblue!60}\textcolor{white}{3.5748} \\
{Normalized Mutual Information} & \cellcolor{myblue!70}\textcolor{white}{\textbf{0.3395}} & \cellcolor{myblue!10}\textcolor{black}{0.0011} & \cellcolor{myblue!30}\textcolor{black}{0.2162} \\
{Adjusted Rand Index} & \cellcolor{myblue!56}\textcolor{white}{0.0037} & \cellcolor{myblue!0}\textcolor{black}{-0.0008} & \cellcolor{myblue!42}\textcolor{black}{0.0035} \\
\bottomrule
\end{tabular}
\vspace{0.5em}
\begin{tabular}{lccc}
\toprule
\textbf{Metric} & \textbf{PaCMap} & \textbf{Random} & \textbf{TSNE} \\
\midrule
 & \includegraphics[width=0.18\textwidth]{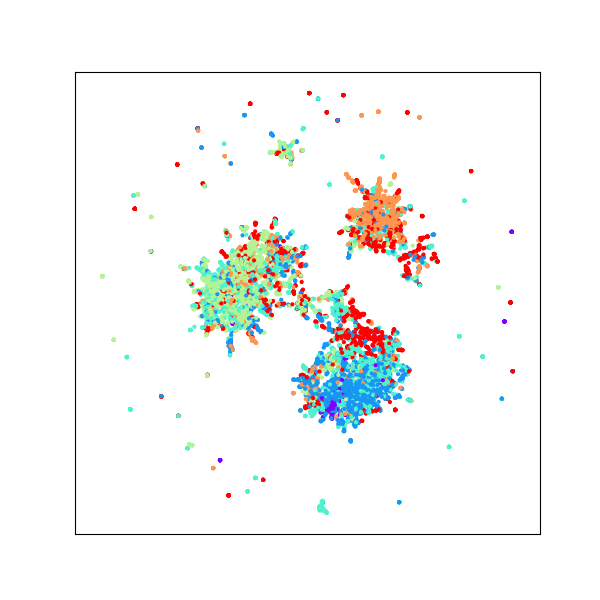} & \includegraphics[width=0.18\textwidth]{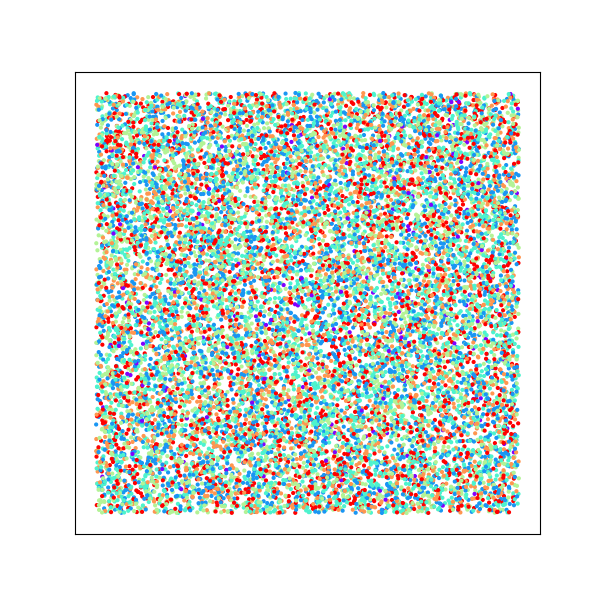} & \includegraphics[width=0.18\textwidth]{images/emb_figs/trec/TSNE.png} \\
\midrule
{\ourmetricFull} & \cellcolor{myblue!40}\textcolor{black}{0.0756} & \cellcolor{myblue!10}\textcolor{black}{0.1005} & \cellcolor{myblue!50}\textcolor{white}{0.0734} \\
{Cluster Distance Score} & \cellcolor{myblue!60}\textcolor{white}{0.2815} & \cellcolor{myblue!20}\textcolor{black}{0.3987} & \cellcolor{myblue!40}\textcolor{black}{0.3246} \\
{Label-T\&C} & \cellcolor{myblue!30}\textcolor{black}{0.8879} & \cellcolor{myblue!0}\textcolor{black}{0.7371} & \cellcolor{myblue!40}\textcolor{black}{0.8913} \\
{Steadiness \& Cohesiveness} & \cellcolor{myblue!60}\textcolor{white}{0.6254} & \cellcolor{myblue!10}\textcolor{black}{0.4942} & \cellcolor{myblue!70}\textcolor{white}{\textbf{0.7076}} \\
{Silhouette Score} & \cellcolor{myblue!20}\textcolor{black}{-0.1071} & \cellcolor{myblue!60}\textcolor{white}{-0.0294} & \cellcolor{myblue!10}\textcolor{black}{-0.1216} \\
{Davies Bouldin Index} & \cellcolor{myblue!50}\textcolor{white}{4.5360} & \cellcolor{myblue!0}\textcolor{black}{128.8955} & \cellcolor{myblue!40}\textcolor{black}{4.8631} \\
{Normalized Mutual Information} & \cellcolor{myblue!20}\textcolor{black}{0.2159} & \cellcolor{myblue!0}\textcolor{black}{0.0005} & \cellcolor{myblue!50}\textcolor{white}{0.2474} \\
{Adjusted Rand Index} & \cellcolor{myblue!70}\textcolor{white}{\textbf{0.0066}} & \cellcolor{myblue!13}\textcolor{black}{-0.0001} & \cellcolor{myblue!28}\textcolor{black}{0.0033} \\
\bottomrule
\end{tabular}
\vspace{0.5em}
\begin{tabular}{lcc}
\toprule
\textbf{Metric} & \textbf{UMAP} & \textbf{UMATO} \\
\midrule
 & \includegraphics[width=0.18\textwidth]{images/emb_figs/trec/UMAP.png} & \includegraphics[width=0.18\textwidth]{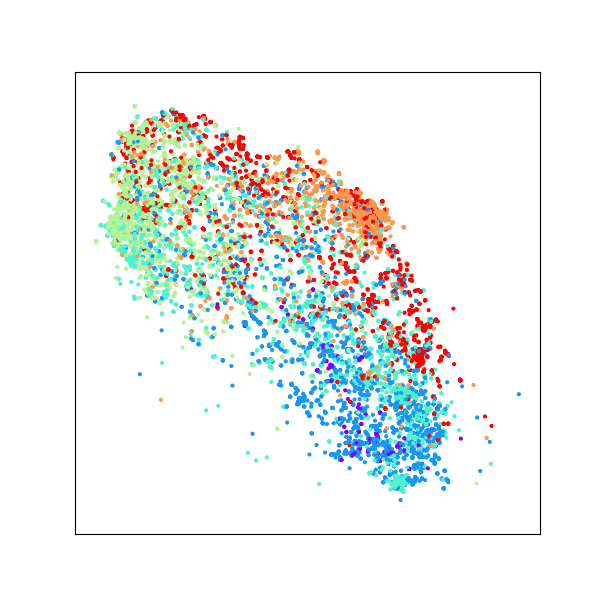} \\
\midrule
{\ourmetricFull} & \cellcolor{myblue!0}\textcolor{black}{0.1043} & \cellcolor{myblue!30}\textcolor{black}{0.0810} \\
{Cluster Distance Score} & \cellcolor{myblue!10}\textcolor{black}{0.4341} & \cellcolor{myblue!30}\textcolor{black}{0.3466} \\
{Label-T\&C} & \cellcolor{myblue!10}\textcolor{black}{0.7630} & \cellcolor{myblue!50}\textcolor{white}{0.8920} \\
{Steadiness \& Cohesiveness} & \cellcolor{myblue!50}\textcolor{white}{0.5723} & \cellcolor{myblue!40}\textcolor{black}{0.5692} \\
{Silhouette Score} & \cellcolor{myblue!50}\textcolor{white}{-0.0700} & \cellcolor{myblue!0}\textcolor{black}{-0.1438} \\
{Davies Bouldin Index} & \cellcolor{myblue!10}\textcolor{black}{30.2695} & \cellcolor{myblue!30}\textcolor{black}{5.2292} \\
{Normalized Mutual Information} & \cellcolor{myblue!60}\textcolor{white}{0.2500} & \cellcolor{myblue!40}\textcolor{black}{0.2332} \\
{Adjusted Rand Index} & \cellcolor{myblue!56}\textcolor{white}{0.0037} & \cellcolor{myblue!56}\textcolor{white}{0.0037} \\
\bottomrule
\end{tabular}
\vspace{0.5em}
\end{table*}

\begin{table*}[ht!]
\centering
\caption{Embedding quality metrics for \textbf{sentiment} dataset.}
\vspace{0.5em}
\begin{tabular}{lccc}
\toprule
\textbf{Metric} & \textbf{AngleEmbedding} & \textbf{MDS} & \textbf{PCA} \\
\midrule
 & \includegraphics[width=0.18\textwidth]{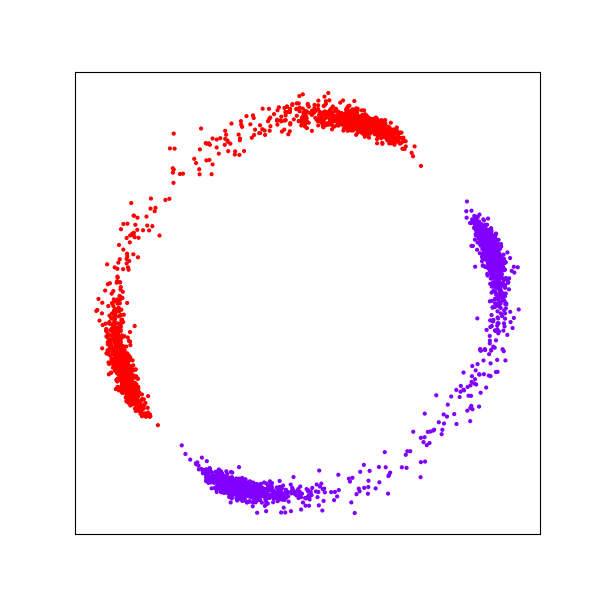} & \includegraphics[width=0.18\textwidth]{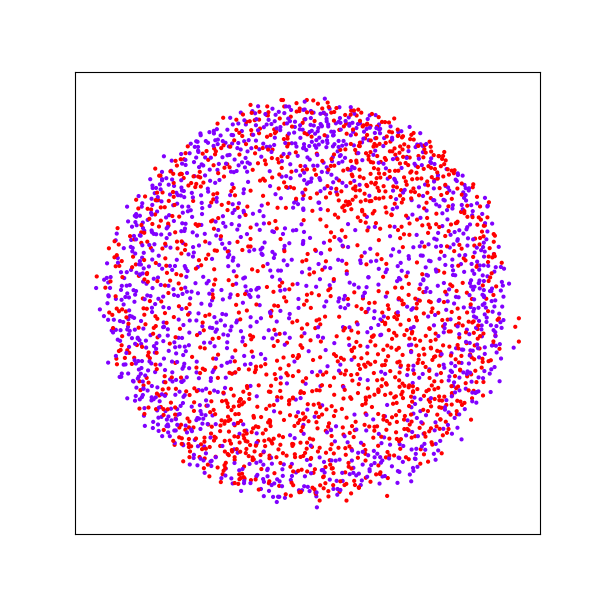} & \includegraphics[width=0.18\textwidth]{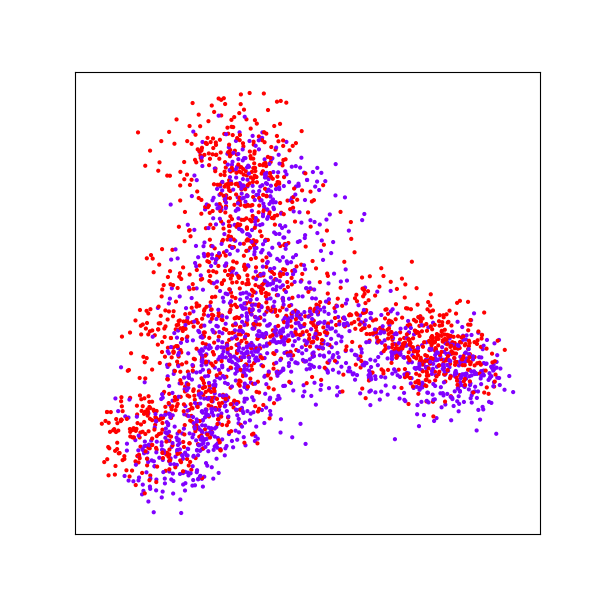} \\
\midrule
{\ourmetricFull} & \cellcolor{myblue!70}\textcolor{white}{\textbf{0.0278}} & \cellcolor{myblue!60}\textcolor{white}{0.0841} & \cellcolor{myblue!10}\textcolor{black}{0.1011} \\
{Cluster Distance Score} & \cellcolor{myblue!35}\textcolor{black}{\textbf{0.0000}} & \cellcolor{myblue!35}\textcolor{black}{\textbf{0.0000}} & \cellcolor{myblue!35}\textcolor{black}{\textbf{0.0000}} \\
{Label-T\&C} & \cellcolor{myblue!70}\textcolor{white}{\textbf{0.9367}} & \cellcolor{myblue!58}\textcolor{white}{0.8242} & \cellcolor{myblue!46}\textcolor{white}{0.8093} \\
{Steadiness \& Cohesiveness} & \cellcolor{myblue!10}\textcolor{black}{0.5103} & \cellcolor{myblue!20}\textcolor{black}{0.6255} & \cellcolor{myblue!30}\textcolor{black}{0.6518} \\
{Silhouette Score} & \cellcolor{myblue!70}\textcolor{white}{\textbf{0.3552}} & \cellcolor{myblue!60}\textcolor{white}{0.0167} & \cellcolor{myblue!50}\textcolor{white}{0.0155} \\
{Davies Bouldin Index} & \cellcolor{myblue!70}\textcolor{white}{\textbf{1.7090}} & \cellcolor{myblue!50}\textcolor{white}{9.4386} & \cellcolor{myblue!60}\textcolor{white}{7.8921} \\
{Normalized Mutual Information} & \cellcolor{myblue!70}\textcolor{white}{\textbf{0.4974}} & \cellcolor{myblue!0}\textcolor{black}{0.0001} & \cellcolor{myblue!13}\textcolor{black}{0.0008} \\
{Adjusted Rand Index} & \cellcolor{myblue!70}\textcolor{white}{\textbf{0.3650}} & \cellcolor{myblue!0}\textcolor{black}{-0.0000} & \cellcolor{myblue!17}\textcolor{black}{0.0001} \\
\bottomrule
\end{tabular}
\vspace{0.5em}
\begin{tabular}{lccc}
\toprule
\textbf{Metric} & \textbf{PaCMap} & \textbf{Random} & \textbf{TSNE} \\
\midrule
 & \includegraphics[width=0.18\textwidth]{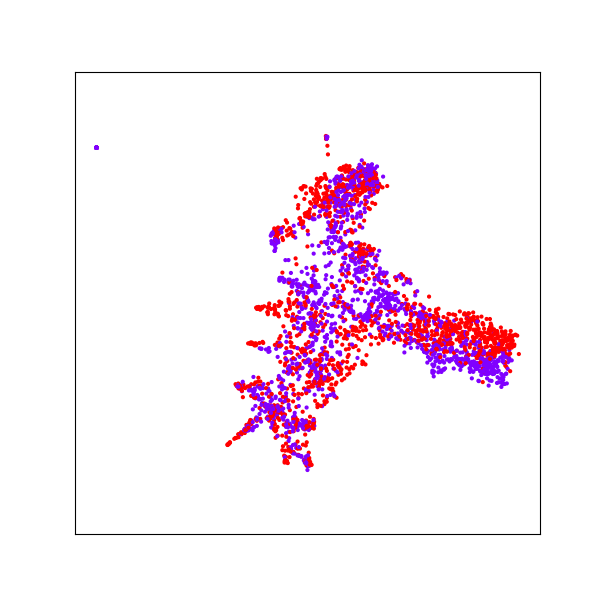} & \includegraphics[width=0.18\textwidth]{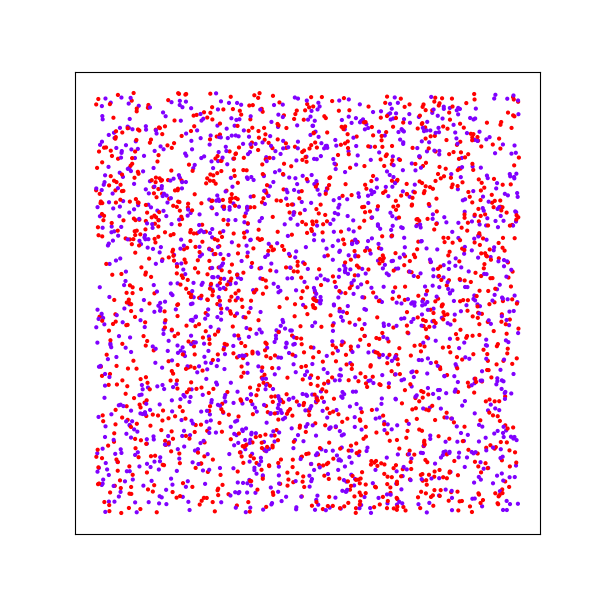} & \includegraphics[width=0.18\textwidth]{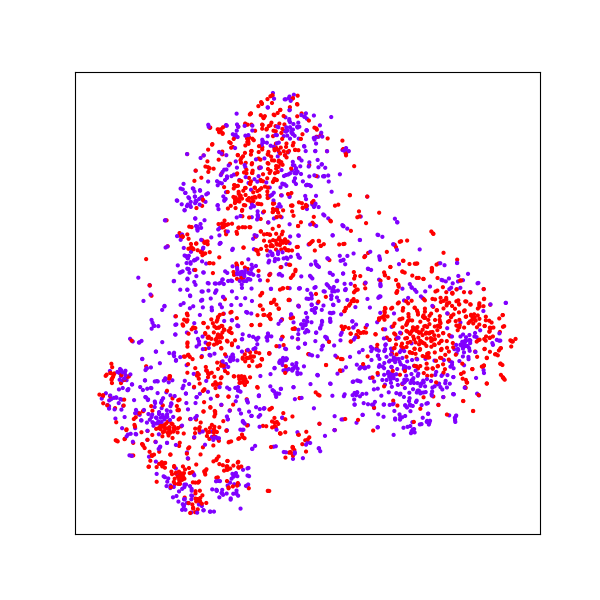} \\
\midrule
{\ourmetricFull} & \cellcolor{myblue!0}\textcolor{black}{0.1026} & \cellcolor{myblue!20}\textcolor{black}{0.0990} & \cellcolor{myblue!50}\textcolor{white}{0.0943} \\
{Cluster Distance Score} & \cellcolor{myblue!35}\textcolor{black}{\textbf{0.0000}} & \cellcolor{myblue!35}\textcolor{black}{\textbf{0.0000}} & \cellcolor{myblue!35}\textcolor{black}{\textbf{0.0000}} \\
{Label-T\&C} & \cellcolor{myblue!11}\textcolor{black}{0.7680} & \cellcolor{myblue!0}\textcolor{black}{0.7667} & \cellcolor{myblue!35}\textcolor{black}{0.7859} \\
{Steadiness \& Cohesiveness} & \cellcolor{myblue!40}\textcolor{black}{0.7184} & \cellcolor{myblue!0}\textcolor{black}{0.4276} & \cellcolor{myblue!60}\textcolor{white}{0.7570} \\
{Silhouette Score} & \cellcolor{myblue!10}\textcolor{black}{0.0049} & \cellcolor{myblue!0}\textcolor{black}{-0.0001} & \cellcolor{myblue!20}\textcolor{black}{0.0071} \\
{Davies Bouldin Index} & \cellcolor{myblue!10}\textcolor{black}{36.7681} & \cellcolor{myblue!0}\textcolor{black}{91.3173} & \cellcolor{myblue!20}\textcolor{black}{20.6965} \\
{Normalized Mutual Information} & \cellcolor{myblue!28}\textcolor{black}{0.0042} & \cellcolor{myblue!0}\textcolor{black}{0.0001} & \cellcolor{myblue!0}\textcolor{black}{0.0000} \\
{Adjusted Rand Index} & \cellcolor{myblue!17}\textcolor{black}{0.0001} & \cellcolor{myblue!0}\textcolor{black}{-0.0000} & \cellcolor{myblue!0}\textcolor{black}{-0.0000} \\
\bottomrule
\end{tabular}
\vspace{0.5em}
\begin{tabular}{lcc}
\toprule
\textbf{Metric} & \textbf{UMAP} & \textbf{UMATO} \\
\midrule
 & \includegraphics[width=0.18\textwidth]{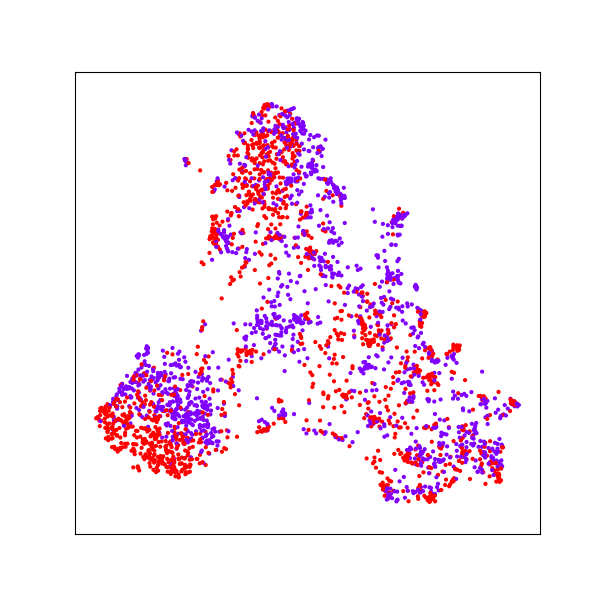} & \includegraphics[width=0.18\textwidth]{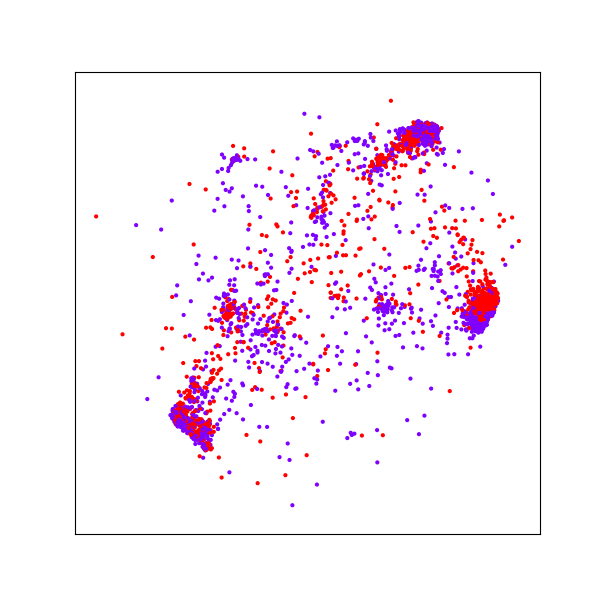} \\
\midrule
{\ourmetricFull} & \cellcolor{myblue!40}\textcolor{black}{0.0972} & \cellcolor{myblue!30}\textcolor{black}{0.0976} \\
{Cluster Distance Score} & \cellcolor{myblue!35}\textcolor{black}{\textbf{0.0000}} & \cellcolor{myblue!35}\textcolor{black}{\textbf{0.0000}} \\
{Label-T\&C} & \cellcolor{myblue!23}\textcolor{black}{0.7755} & \cellcolor{myblue!35}\textcolor{black}{0.7859} \\
{Steadiness \& Cohesiveness} & \cellcolor{myblue!70}\textcolor{white}{\textbf{0.7579}} & \cellcolor{myblue!50}\textcolor{white}{0.7188} \\
{Silhouette Score} & \cellcolor{myblue!40}\textcolor{black}{0.0100} & \cellcolor{myblue!30}\textcolor{black}{0.0081} \\
{Davies Bouldin Index} & \cellcolor{myblue!40}\textcolor{black}{13.0110} & \cellcolor{myblue!30}\textcolor{black}{16.2782} \\
{Normalized Mutual Information} & \cellcolor{myblue!56}\textcolor{white}{0.0436} & \cellcolor{myblue!42}\textcolor{black}{0.0223} \\
{Adjusted Rand Index} & \cellcolor{myblue!35}\textcolor{black}{0.0019} & \cellcolor{myblue!52}\textcolor{white}{0.0057} \\
\bottomrule
\end{tabular}
\vspace{0.5em}
\end{table*}

\begin{table*}[ht!]
\centering
\caption{Embedding quality metrics for \textbf{penguins} dataset.}
\vspace{0.5em}
\begin{tabular}{lccc}
\toprule
\textbf{Metric} & \textbf{AngleEmbedding} & \textbf{MDS} & \textbf{PCA} \\
\midrule
 & \includegraphics[width=0.18\textwidth]{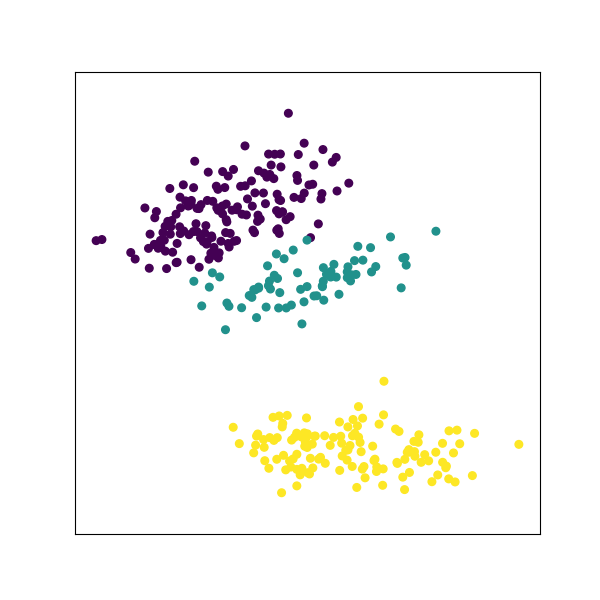} & \includegraphics[width=0.18\textwidth]{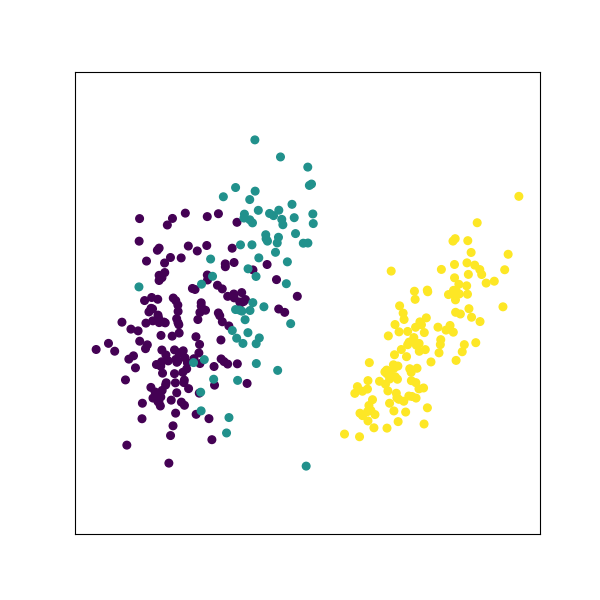} & \includegraphics[width=0.18\textwidth]{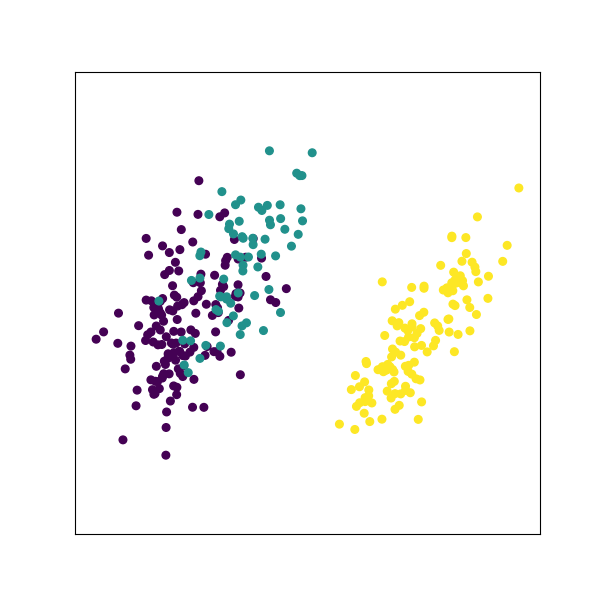} \\
\midrule
{\ourmetricFull} & \cellcolor{myblue!70}\textcolor{white}{\textbf{0.0014}} & \cellcolor{myblue!30}\textcolor{black}{0.0134} & \cellcolor{myblue!10}\textcolor{black}{0.0154} \\
{Cluster Distance Score} & \cellcolor{myblue!46}\textcolor{white}{0.1208} & \cellcolor{myblue!70}\textcolor{white}{\textbf{0.0729}} & \cellcolor{myblue!58}\textcolor{white}{0.0785} \\
{Label-T\&C} & \cellcolor{myblue!70}\textcolor{white}{\textbf{0.9992}} & \cellcolor{myblue!23}\textcolor{black}{0.9717} & \cellcolor{myblue!11}\textcolor{black}{0.9710} \\
{Steadiness \& Cohesiveness} & \cellcolor{myblue!60}\textcolor{white}{0.8150} & \cellcolor{myblue!50}\textcolor{white}{0.7956} & \cellcolor{myblue!20}\textcolor{black}{0.7586} \\
{Silhouette Score} & \cellcolor{myblue!30}\textcolor{black}{0.4671} & \cellcolor{myblue!20}\textcolor{black}{0.3915} & \cellcolor{myblue!10}\textcolor{black}{0.3781} \\
{Davies Bouldin Index} & \cellcolor{myblue!30}\textcolor{black}{0.8738} & \cellcolor{myblue!10}\textcolor{black}{1.1671} & \cellcolor{myblue!20}\textcolor{black}{1.0592} \\
{Normalized Mutual Information} & \cellcolor{myblue!70}\textcolor{white}{\textbf{0.7671}} & \cellcolor{myblue!35}\textcolor{black}{0.7504} & \cellcolor{myblue!70}\textcolor{white}{\textbf{0.7671}} \\
{Adjusted Rand Index} & \cellcolor{myblue!52}\textcolor{white}{0.6551} & \cellcolor{myblue!35}\textcolor{black}{0.6535} & \cellcolor{myblue!52}\textcolor{white}{0.6551} \\
\bottomrule
\end{tabular}
\vspace{0.5em}
\begin{tabular}{lccc}
\toprule
\textbf{Metric} & \textbf{PaCMap} & \textbf{Random} & \textbf{TSNE} \\
\midrule
 & \includegraphics[width=0.18\textwidth]{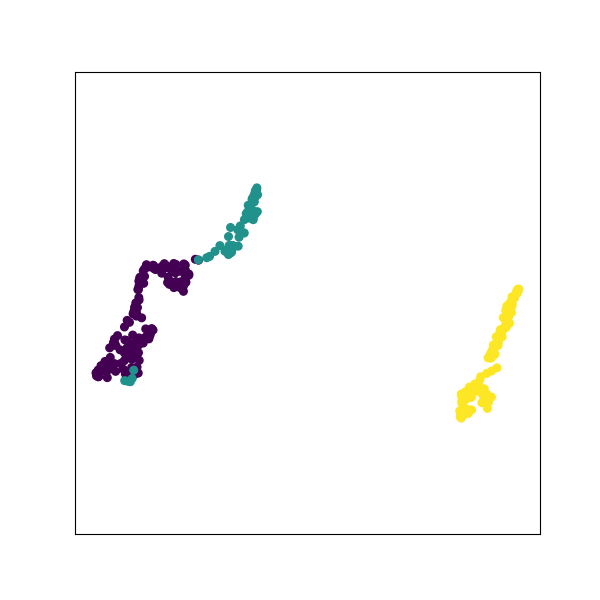} & \includegraphics[width=0.18\textwidth]{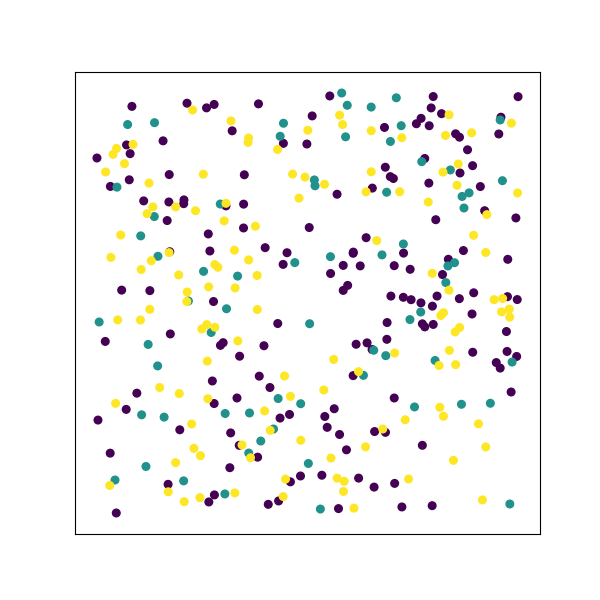} & \includegraphics[width=0.18\textwidth]{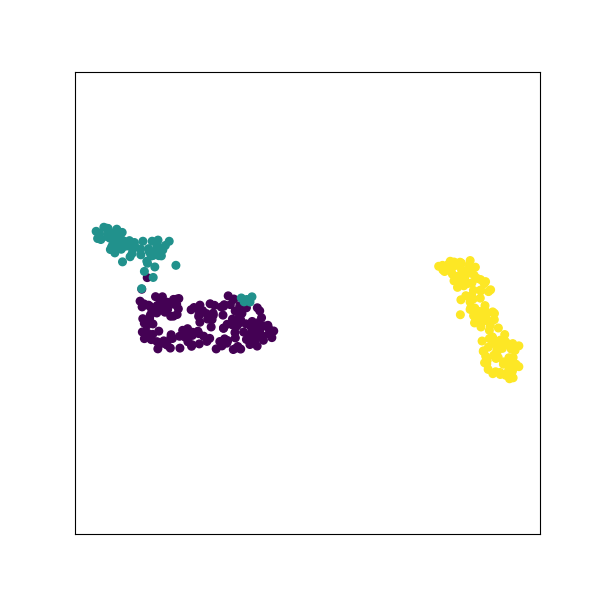} \\
\midrule
{\ourmetricFull} & \cellcolor{myblue!20}\textcolor{black}{0.0150} & \cellcolor{myblue!0}\textcolor{black}{0.1535} & \cellcolor{myblue!60}\textcolor{white}{0.0082} \\
{Cluster Distance Score} & \cellcolor{myblue!58}\textcolor{white}{0.0785} & \cellcolor{myblue!35}\textcolor{black}{0.1442} & \cellcolor{myblue!0}\textcolor{black}{0.2318} \\
{Label-T\&C} & \cellcolor{myblue!46}\textcolor{white}{0.9939} & \cellcolor{myblue!0}\textcolor{black}{0.7218} & \cellcolor{myblue!58}\textcolor{white}{0.9985} \\
{Steadiness \& Cohesiveness} & \cellcolor{myblue!10}\textcolor{black}{0.7206} & \cellcolor{myblue!0}\textcolor{black}{0.2828} & \cellcolor{myblue!70}\textcolor{white}{\textbf{0.8156}} \\
{Silhouette Score} & \cellcolor{myblue!60}\textcolor{white}{0.6387} & \cellcolor{myblue!0}\textcolor{black}{-0.0207} & \cellcolor{myblue!50}\textcolor{white}{0.6180} \\
{Davies Bouldin Index} & \cellcolor{myblue!60}\textcolor{white}{0.5248} & \cellcolor{myblue!0}\textcolor{black}{22.4132} & \cellcolor{myblue!50}\textcolor{white}{0.5666} \\
{Normalized Mutual Information} & \cellcolor{myblue!17}\textcolor{black}{0.4693} & \cellcolor{myblue!0}\textcolor{black}{0.0007} & \cellcolor{myblue!70}\textcolor{white}{\textbf{0.7671}} \\
{Adjusted Rand Index} & \cellcolor{myblue!17}\textcolor{black}{0.1944} & \cellcolor{myblue!0}\textcolor{black}{-0.0029} & \cellcolor{myblue!52}\textcolor{white}{0.6551} \\
\bottomrule
\end{tabular}
\vspace{0.5em}
\begin{tabular}{lcc}
\toprule
\textbf{Metric} & \textbf{UMAP} & \textbf{UMATO} \\
\midrule
 & \includegraphics[width=0.18\textwidth]{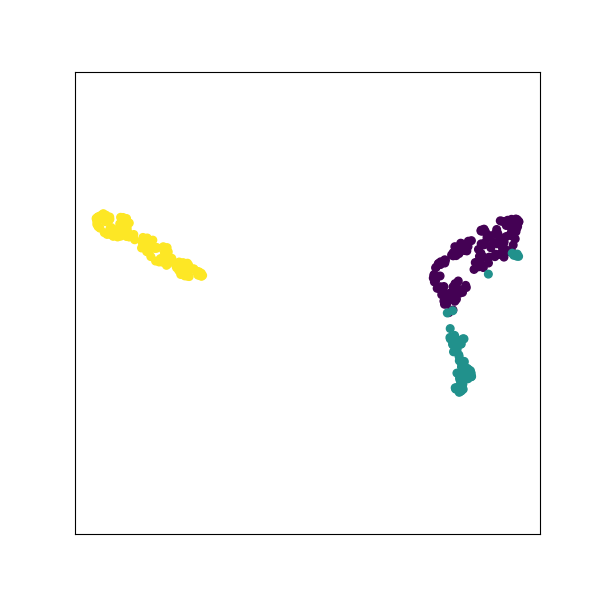} & \includegraphics[width=0.18\textwidth]{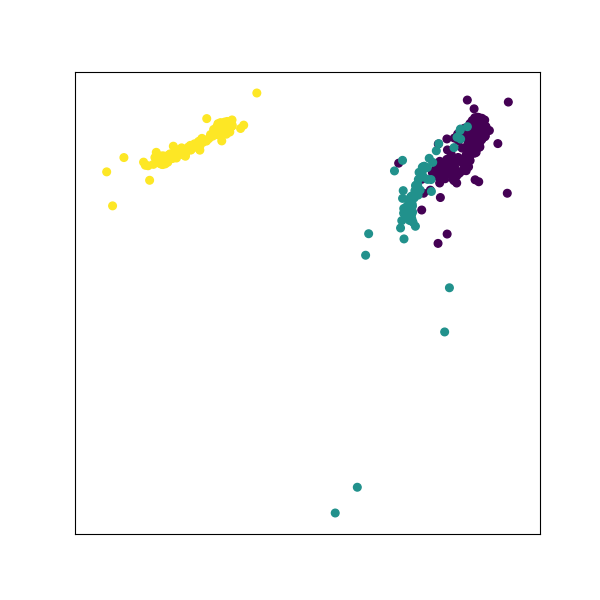} \\
\midrule
{\ourmetricFull} & \cellcolor{myblue!50}\textcolor{white}{0.0097} & \cellcolor{myblue!40}\textcolor{black}{0.0126} \\
{Cluster Distance Score} & \cellcolor{myblue!11}\textcolor{black}{0.1739} & \cellcolor{myblue!23}\textcolor{black}{0.1477} \\
{Label-T\&C} & \cellcolor{myblue!70}\textcolor{white}{\textbf{0.9992}} & \cellcolor{myblue!35}\textcolor{black}{0.9900} \\
{Steadiness \& Cohesiveness} & \cellcolor{myblue!30}\textcolor{black}{0.7680} & \cellcolor{myblue!40}\textcolor{black}{0.7689} \\
{Silhouette Score} & \cellcolor{myblue!70}\textcolor{white}{\textbf{0.6606}} & \cellcolor{myblue!40}\textcolor{black}{0.5789} \\
{Davies Bouldin Index} & \cellcolor{myblue!70}\textcolor{white}{\textbf{0.4986}} & \cellcolor{myblue!40}\textcolor{black}{0.6935} \\
{Normalized Mutual Information} & \cellcolor{myblue!70}\textcolor{white}{\textbf{0.7671}} & \cellcolor{myblue!52}\textcolor{white}{0.7603} \\
{Adjusted Rand Index} & \cellcolor{myblue!52}\textcolor{white}{0.6551} & \cellcolor{myblue!70}\textcolor{white}{\textbf{0.6594}} \\
\bottomrule
\end{tabular}
\vspace{0.5em}
\end{table*}

\begin{table*}[ht!]
\centering
\caption{Embedding quality metrics for \textbf{pendigits} dataset.}
\vspace{0.5em}
\begin{tabular}{lccc}
\toprule
\textbf{Metric} & \textbf{AngleEmbedding} & \textbf{MDS} & \textbf{PCA} \\
\midrule
 & \includegraphics[width=0.18\textwidth]{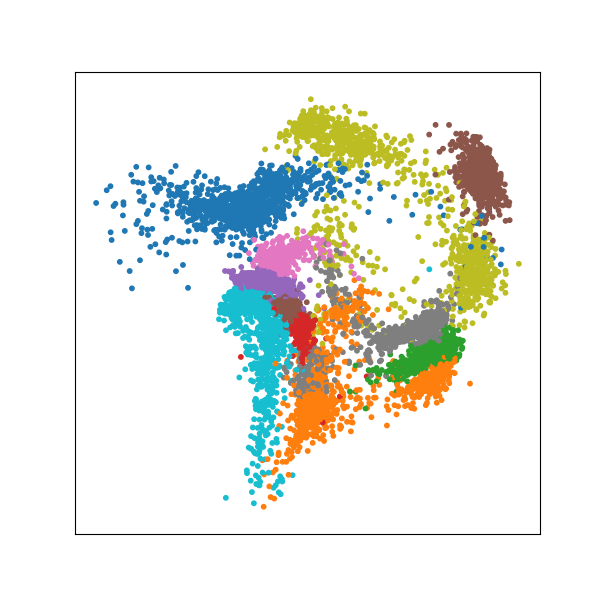} & \includegraphics[width=0.18\textwidth]{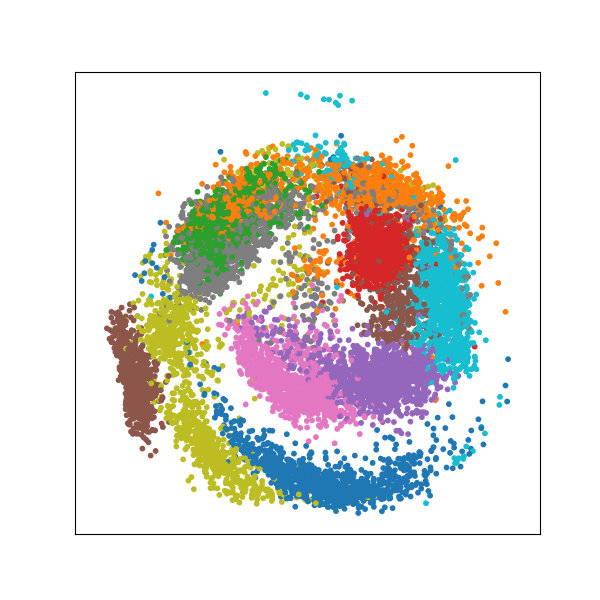} & \includegraphics[width=0.18\textwidth]{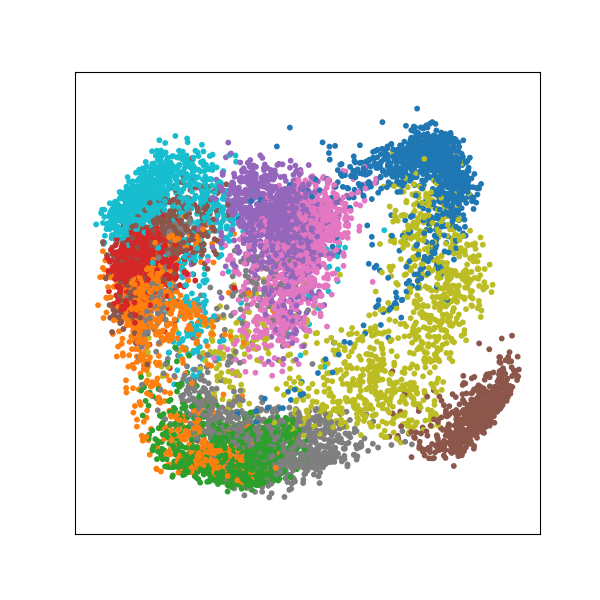} \\
\midrule
{\ourmetricFull} & \cellcolor{myblue!70}\textcolor{white}{\textbf{0.0101}} & \cellcolor{myblue!42}\textcolor{black}{0.0181} & \cellcolor{myblue!13}\textcolor{black}{0.0226} \\
{Cluster Distance Score} & \cellcolor{myblue!20}\textcolor{black}{0.2636} & \cellcolor{myblue!60}\textcolor{white}{0.2020} & \cellcolor{myblue!50}\textcolor{white}{0.2119} \\
{Label-T\&C} & \cellcolor{myblue!35}\textcolor{black}{0.9824} & \cellcolor{myblue!23}\textcolor{black}{0.9812} & \cellcolor{myblue!11}\textcolor{black}{0.9749} \\
{Steadiness \& Cohesiveness} & \cellcolor{myblue!30}\textcolor{black}{0.7375} & \cellcolor{myblue!20}\textcolor{black}{0.6270} & \cellcolor{myblue!10}\textcolor{black}{0.4786} \\
{Silhouette Score} & \cellcolor{myblue!30}\textcolor{black}{0.1815} & \cellcolor{myblue!20}\textcolor{black}{0.1215} & \cellcolor{myblue!10}\textcolor{black}{0.0565} \\
{Davies Bouldin Index} & \cellcolor{myblue!10}\textcolor{black}{3.0486} & \cellcolor{myblue!30}\textcolor{black}{2.4307} & \cellcolor{myblue!20}\textcolor{black}{2.9925} \\
{Normalized Mutual Information} & \cellcolor{myblue!30}\textcolor{black}{0.6495} & \cellcolor{myblue!10}\textcolor{black}{0.0020} & \cellcolor{myblue!20}\textcolor{black}{0.0043} \\
{Adjusted Rand Index} & \cellcolor{myblue!23}\textcolor{black}{0.4189} & \cellcolor{myblue!11}\textcolor{black}{0.0000} & \cellcolor{myblue!11}\textcolor{black}{0.0001} \\
\bottomrule
\end{tabular}
\vspace{0.5em}
\begin{tabular}{lccc}
\toprule
\textbf{Metric} & \textbf{PaCMap} & \textbf{Random} & \textbf{TSNE} \\
\midrule
 & \includegraphics[width=0.18\textwidth]{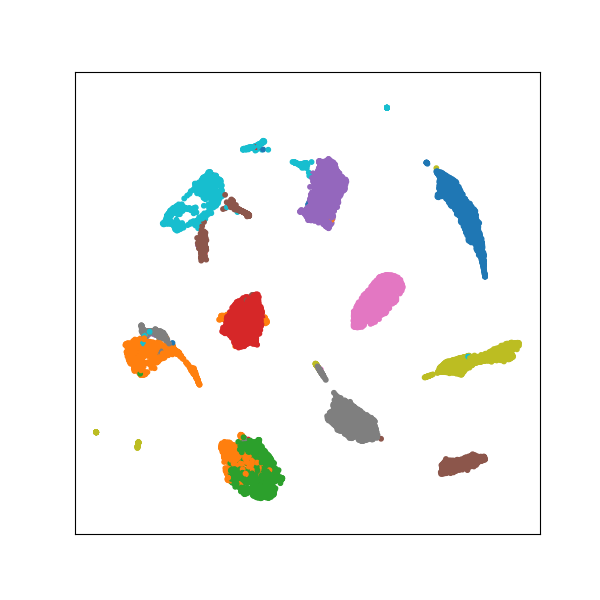} & \includegraphics[width=0.18\textwidth]{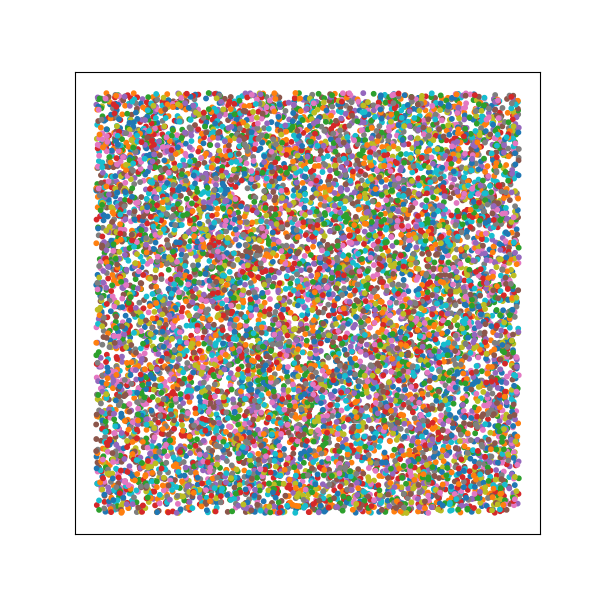} & \includegraphics[width=0.18\textwidth]{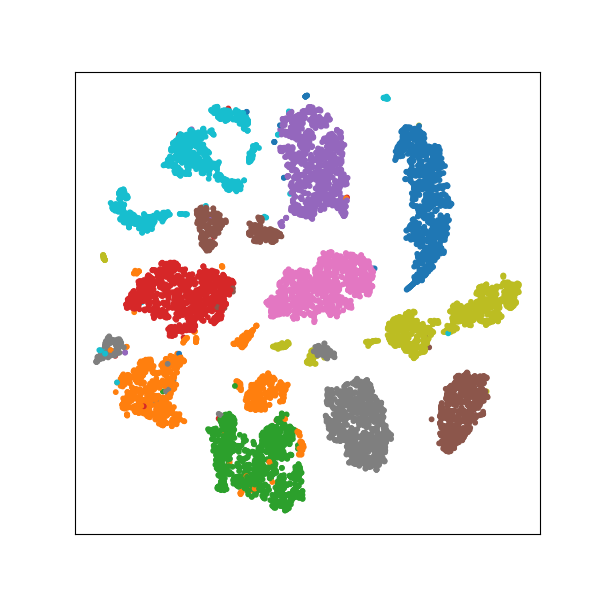} \\
\midrule
{\ourmetricFull} & \cellcolor{myblue!28}\textcolor{black}{0.0207} & \cellcolor{myblue!0}\textcolor{black}{0.1426} & \cellcolor{myblue!42}\textcolor{black}{0.0181} \\
{Cluster Distance Score} & \cellcolor{myblue!30}\textcolor{black}{0.2538} & \cellcolor{myblue!0}\textcolor{black}{0.5010} & \cellcolor{myblue!40}\textcolor{black}{0.2455} \\
{Label-T\&C} & \cellcolor{myblue!58}\textcolor{white}{0.9870} & \cellcolor{myblue!0}\textcolor{black}{0.7245} & \cellcolor{myblue!58}\textcolor{white}{0.9871} \\
{Steadiness \& Cohesiveness} & \cellcolor{myblue!50}\textcolor{white}{0.8495} & \cellcolor{myblue!0}\textcolor{black}{0.3271} & \cellcolor{myblue!40}\textcolor{black}{0.8421} \\
{Silhouette Score} & \cellcolor{myblue!60}\textcolor{white}{0.4681} & \cellcolor{myblue!0}\textcolor{black}{-0.0177} & \cellcolor{myblue!40}\textcolor{black}{0.3317} \\
{Davies Bouldin Index} & \cellcolor{myblue!40}\textcolor{black}{2.0347} & \cellcolor{myblue!0}\textcolor{black}{158.4236} & \cellcolor{myblue!60}\textcolor{white}{1.7945} \\
{Normalized Mutual Information} & \cellcolor{myblue!70}\textcolor{white}{\textbf{0.8477}} & \cellcolor{myblue!0}\textcolor{black}{0.0003} & \cellcolor{myblue!40}\textcolor{black}{0.7617} \\
{Adjusted Rand Index} & \cellcolor{myblue!70}\textcolor{white}{\textbf{0.7903}} & \cellcolor{myblue!0}\textcolor{black}{-0.0000} & \cellcolor{myblue!35}\textcolor{black}{0.6498} \\
\bottomrule
\end{tabular}
\vspace{0.5em}
\begin{tabular}{lcc}
\toprule
\textbf{Metric} & \textbf{UMAP} & \textbf{UMATO} \\
\midrule
 & \includegraphics[width=0.18\textwidth]{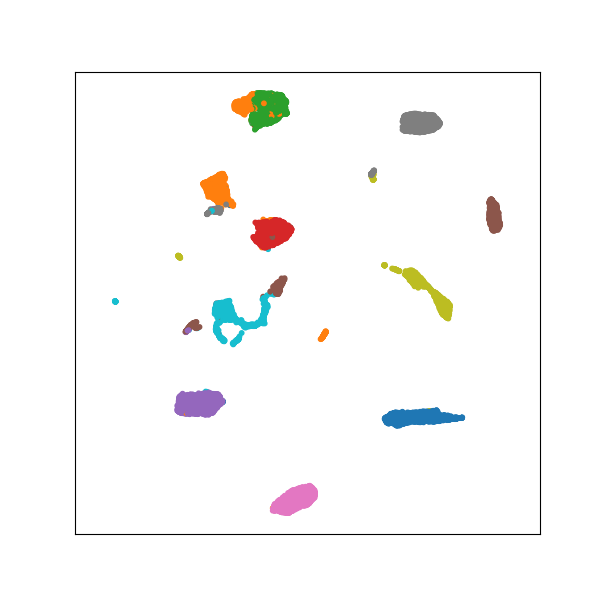} & \includegraphics[width=0.18\textwidth]{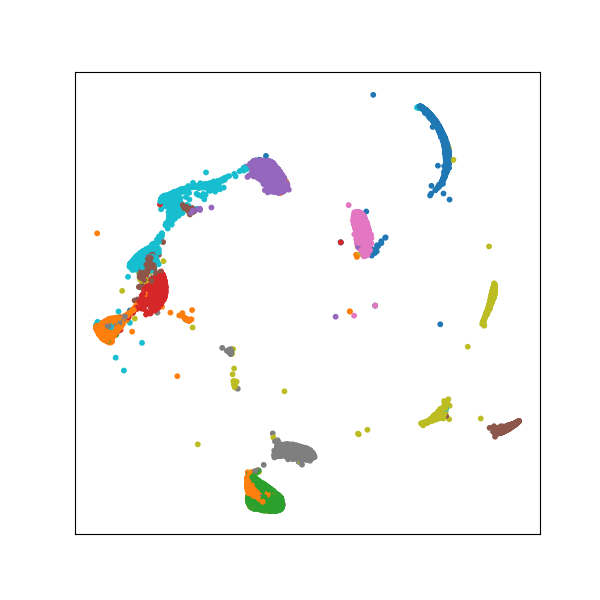} \\
\midrule
{\ourmetricFull} & \cellcolor{myblue!13}\textcolor{black}{0.0225} & \cellcolor{myblue!56}\textcolor{white}{0.0167} \\
{Cluster Distance Score} & \cellcolor{myblue!10}\textcolor{black}{0.2765} & \cellcolor{myblue!70}\textcolor{white}{\textbf{0.1904}} \\
{Label-T\&C} & \cellcolor{myblue!70}\textcolor{white}{\textbf{0.9876}} & \cellcolor{myblue!46}\textcolor{white}{0.9840} \\
{Steadiness \& Cohesiveness} & \cellcolor{myblue!60}\textcolor{white}{0.8639} & \cellcolor{myblue!70}\textcolor{white}{\textbf{0.8797}} \\
{Silhouette Score} & \cellcolor{myblue!70}\textcolor{white}{\textbf{0.5297}} & \cellcolor{myblue!50}\textcolor{white}{0.3367} \\
{Davies Bouldin Index} & \cellcolor{myblue!70}\textcolor{white}{\textbf{1.6005}} & \cellcolor{myblue!50}\textcolor{white}{1.9495} \\
{Normalized Mutual Information} & \cellcolor{myblue!60}\textcolor{white}{0.8303} & \cellcolor{myblue!50}\textcolor{white}{0.8152} \\
{Adjusted Rand Index} & \cellcolor{myblue!58}\textcolor{white}{0.7843} & \cellcolor{myblue!46}\textcolor{white}{0.7437} \\
\bottomrule
\end{tabular}
\vspace{0.5em}
\end{table*}

\begin{table*}[ht!]
\centering
\caption{Embedding quality metrics for \textbf{pbmc3k} dataset.}
\vspace{0.5em}
%
\vspace{0.5em}
\end{table*}

\begin{table*}[ht!]
\centering
\caption{Embedding quality metrics for \textbf{olivetti} dataset.}
\vspace{0.5em}
%
\vspace{0.5em}
\end{table*}

\begin{table*}[ht!]
\centering
\caption{Embedding quality metrics for \textbf{liver} dataset.}
\vspace{0.5em}
%
\vspace{0.5em}
\end{table*}

\begin{table*}[ht!]
\centering
\caption{Embedding quality metrics for \textbf{F-MNIST} dataset.}
\vspace{0.5em}
%
\vspace{0.5em}
\end{table*}

\begin{table*}[ht!]
\centering
\caption{Embedding quality metrics for \textbf{emotion} dataset.}
\vspace{0.5em}
%
\vspace{0.5em}
\end{table*}

\begin{table*}[ht!]
\centering
\caption{Embedding quality metrics for \textbf{coil20} dataset.}
\vspace{0.5em}
%
\vspace{0.5em}
\end{table*}

\begin{table*}[ht!]
\centering
\caption{Embedding quality metrics for \textbf{coil100} dataset.}
\vspace{0.5em}
%
\vspace{0.5em}
\end{table*}

\begin{table*}[ht!]
\centering
\caption{Embedding quality metrics for \textbf{acl\_imdb} dataset.}
\vspace{0.5em}
%
\vspace{0.5em}
\end{table*}

\begin{table*}[ht!]
\centering
\caption{Embedding quality metrics for \textbf{MNIST} dataset.}
\vspace{0.5em}
%
\vspace{0.5em}
\end{table*}

\end{document}